\documentclass[journal]{IEEEtran}
\usepackage{amsmath}
\usepackage{amsfonts}       
\usepackage{multirow}
\usepackage{makecell}
\usepackage[capitalize]{cleveref}

\usepackage{ifpdf}

\usepackage{cite}

\ifCLASSINFOpdf
   \usepackage[pdftex]{graphicx}
\else
   \usepackage[dvips]{graphicx}
\fi
\usepackage{amsmath}
\usepackage{algorithmic,algorithm}

\usepackage{array}
\usepackage{adjustbox}
\usepackage{tikz}
\usepackage{algorithmic,algorithm}

\ifCLASSOPTIONcompsoc
  \usepackage[caption=false,font=normalsize,labelfont=sf,textfont=sf]{subfig}
\else
  \usepackage[caption=false,font=footnotesize]{subfig}
\fi

\usepackage{xspace}
\makeatletter
\newcommand\onedot{\futurelet\@let@token\@onedot}
\def\@onedot{\ifx\@let@token.\else.\null\fi\xspace}
\def\eg{\emph{e.g}\onedot} 
\def\ie{\emph{i.e}\onedot} 
 
\def\etal{\emph{et al}\onedot}

\makeatother

\newcommand*\annotatedFigureBoxCustom[3]{\draw[#1,thin] (#2) rectangle (#3);}
\newcommand*\annotatedFigureBox[3]{\annotatedFigureBoxCustom{#1}{#2}{#3}}
\newenvironment {annotatedFigure}[1]
{\centering
	\begin{tikzpicture}
		\node[anchor=south west,inner sep=0] (image) at (0,0) { #1};\begin{scope}[x={(image.south east)},y={(image.north west)}]}{\end{scope}
	\end{tikzpicture}}

\begin{document}
%
\title{Unsupervised Pansharpening Based on Self-Attention Mechanism}
%
%
%

\author{Ying~Qu, \IEEEmembership{Member,~IEEE,}
	Razieh Kaviani Baghbaderani, \IEEEmembership{Student Member,~IEEE,}
	Hairong~Qi, \IEEEmembership{Fellow,~IEEE}
	and Chiman~Kwan, \IEEEmembership{Senior Member,~IEEE,}
\thanks{Manuscript received by TGRS on January 2, 2020; revised April 24, 2020 and June 9, 2020; accepted July 8, 2020.} 
\thanks{Ying~Qu, Razieh Kaviani Baghbaderani and Hairong~Qi are with the Advanced Imaging and Collaborative Information Processing Group, Department of Electrical Engineering and Computer Science, University of Tennessee, Knoxville, TN 37996 USA (e-mail: yqu3@vols.utk.edu; rkavian1@vols.utk.edu; hqi@utk.edu).}
\thanks{Chiman~Kwan is with Applied Research LLC, Rockville, MD,20850 USA (chiman.kwan@arllc.net)}}

%
%

\markboth{Accepted by IEEE TRANSACTIONS ON GEOSCIENCE AND REMOTE SENSING}%
{Shell \MakeLowercase{\textit{et al.}}: Bare Demo of IEEEtran.cls for IEEE Journals}
%



\maketitle

\begin{abstract}
\textcolor{blue}{(Please find the final version from IEEE Transactions on Geoscience and Remote Sensing on IEEE Xplore.)} Pansharpening is to fuse a multispectral image (MSI) of low-spatial-resolution (LR) but rich spectral characteristics with a panchromatic image (PAN) of high-spatial-resolution (HR) but poor spectral characteristics. Traditional methods usually inject the extracted high-frequency details from PAN into the up-sampled MSI. Recent deep learning endeavors are mostly supervised assuming the HR MSI is available, which is unrealistic especially for satellite images. Nonetheless, these methods could not fully exploit the rich spectral characteristics in the MSI. Due to the wide existence of mixed pixels in satellite images where each pixel tends to cover more than one constituent material, pansharpening at the subpixel level becomes essential. In this paper, we propose an unsupervised pansharpening (UP) method in a deep-learning framework to address the above challenges based on the self-attention mechanism (SAM), referred to as UP-SAM. The contribution of this paper is three-fold. First, the self-attention mechanism is proposed where the spatial varying detail extraction and injection functions are estimated according to the attention representations indicating spectral characteristics of the MSI with sub-pixel accuracy. Second, such attention representations are derived from mixed pixels with the proposed stacked attention network powered with a stick-breaking structure to meet the physical constraints of mixed pixel formulations. Third, the detail extraction and injection functions are spatial varying based on the attention representations, which largely improves the reconstruction accuracy. Extensive experimental results demonstrate that the proposed approach is able to reconstruct sharper MSI of different types, with more details and less spectral distortion as compared to the state-of-the-art.

\end{abstract}

\begin{IEEEkeywords}
Unsupervised deep learning, Attention Mechanism, Multispectral Image, Pansharpening 
\end{IEEEkeywords}

%

\section{Introduction}
\label{sec:intro}
\IEEEPARstart{M}ultispectral image (MSI) is one of the most widely utilized satellite optical image. It usually covers the electromagnetic spectrum at the visible and near-infrared wavelengths with up to eight spectral bands. Due to the rich spectral characteristics, MSI has been deeply involved in various human activities, including, but not limited to, environmental change detection~\cite{bovolo2009analysis}, agriculture monitoring~\cite{souza2003mapping,gilbertson2017effect}, weather forecasting~\cite{song2015adaptive}, and so on. Very often, the analysis results of these applications rely heavily on both the spatial and spectral resolution of MSI. However, due to the physical limitations of satellite optical sensors, the high-spectral resolution of MSI can only be achieved by sacrificing its spatial resolution. On the other hand, satellite sensors are also able to acquire Panchromatic images (PAN) with high-spatial resolution although the spectral resolution is poor with only one spectral band. 

The technique of ``Pansharpening'' overcomes the limitations of both MSI and PAN by fusing the high-spatial resolution (HR) PAN and low-spatial resolution (LR) MSI (but with rich spectral characteristics)~\cite{aiazzi2012twenty, vivone2014critical,restaino2017context,8435988}. The fused image, with high resolution in both the spatial and spectral domain, would largely facilitate more effective problem solving. 

Classical pansharpening methods generally include two sequential steps,~\ie, the extraction of high-frequency spatial details from HR PAN and the injection of such details into the upsampled LR MSI~\cite{aiazzi2012twenty,vivone2014critical,restaino2017context}. According to the way how spatial details are extracted, classical methods can be roughly categorized into two groups, namely, the component substitution (CS) ~\cite{thomas2008synthesis,chavez1991comparison,aiazzi2007improving,garzelli2008,choi2010new,vivone2019robust} and the multi-resolution analysis (MRA) based approaches~\cite{aiazzi2006mtf,alparone2007comparison,vivone2018full}. The injection procedure also varies according to how the injection coefficients are defined. Global-based methods inject the details in a global way~\cite{aiazzi2007improving,garzelli2008,choi2010new,aiazzi2006mtf,alparone2007comparison},~\ie, for each band, the injection coefficient is the same for the entire image. 

Context-adaptive based methods, on the other hand, inject the details in a local manner,~\ie, for each band, the coefficients vary according to the context of the image~\cite{wang2013robust, garzelli2014pansharpening, restaino2017context}. 
Restaino~\etal~\cite{restaino2017context} illustrated that the optimal injection coefficients should be spatially variant corresponding to the spectral characteristics of the pixels.  
They proposed a context-adaptive approach where the injection coefficients are estimated according to the image context segmented through a binary partition tree.
However, the potential problem of this method is that the injection coefficients rely heavily on the context segments defined according to the raw pixels of LR MSI. Since the raw pixels of LR MSI are usually mixed,~\ie, each pixel tends to cover more than one constituent material within the instantaneous field of view of the sensor~\cite{bioucas2012hyperspectral,li2016robust, qu2018udas}, the same segment may cover the pixels possessing different spectral characteristics.

Recently, deep learning, as the emerging frontier of machine learning, has been adopted to address the problem of pansharpening and achieved state-of-the-art performance~\cite{scarpa2018target,he2019hyperpnn,he2019pansharpening}. The major principle of deep-learning based methods is to learn a mapping function between a pair of LR MSI and HR PAN images with an HR MSI image in a supervised fashion.  Such mapping function, defined by the weights of the network, reflects the combined functionality of detail extraction and detail injection as in traditional approaches. In 2015, a modified sparse tied-weights denoising autoencoder was proposed by Huang \textit{et al.}~\cite{huang2015new} to enhance the resolution of MSI. The method assumes that the mapping function between LR and HR PAN are the same as the one between LR and HR MSI. Giuseppe \etal proposed a supervised three-layer SRCNN~\cite{masi2016pansharpening} to learn the mapping function between LR MSI and HR MSI. Similar to~\cite{masi2016pansharpening}, Wei \etal~\cite{wei2017boosting} learned the mapping function with deep residual networks~\cite{he2016deep}. Yang~\etal~\cite{yang2017pannet} preserved the spectral information of the reconstructed HR MSI by training a mapping function between the high-frequency information extracted from multi-modalities and the residual of LR and HR MSI. Scarpa~\etal~\cite{scarpa2018target} further improved the quality of pan-sharpened MSI through target-adaptive tuning phase, which shows better results across different sensors. 

All the above deep-learning based methods are trained in a supervised fashion, which requires a large dataset, and the availability of ground truth HR MSI. However, there are three potential issues preventing the supervised approaches from being deployed in a real-world scenario: 1) The ground truth HR MSI is, in general, nonexistent. Thus the supervised approaches are inherently impractical. 2) The mapping function may vary from sensor to sensor due to the different properties/configurations of  optical sensing. However, the supervised learning process only yields a universal network model which may very well fail when applied to a sensing system the training set has not seen. 
3) The mapping function extracts and injects details in a global fashion which may not reflect the distinctive spectral characteristics in a local region. These are the three main reasons why existing supervised pansharpening approaches tend to introduce artifacts and blur in the resulting HR MSI.

In this paper, we propose an unsupervised deep learning approach for pansharpening to address the above challenges. The method is motivated based on three major principles, including the \textit{unsupervised} principle, the \textit{pixel-variant} principle, and the \textit{sub-pixel} principle. For each LR MSI, the detail extraction and injection functions should be specifically defined according to its own  characteristics. That is, no training process or training set is needed. Given a pair of LR MSI and HR PAN as the input, the network learns the mapping function and outputs the HR MSI solely based on this pair of inputs, hence the ``unsupervised'' principle. In addition, due to the unsupervised nature of the learning process with very limited inputs, more details need to be exploited. Compared to context-adaptive based approaches where mapping functions are defined based on segments of raw LR MSI, the unsupervised approach estimates the functions based on the spectral characteristics at the pixel level, hence the ``pixel-variant'' principle.  The raw pixels of LR MSI are usually mixed. To improve the reconstruction accuracy and enhance the representative power of the network, the mapping functions are defined at the sub-pixel level, hence the ``sub-pixel'' principle.  The enabling technique to these three principles is the so-called \textit{self-attention} mechanism, where the mapping functions learned in the unsupervised way look beyond what is immediately perceivable at the context level (i.e., image segment level) and pays special attention to the spectral characteristics of the pixel itself. 
We refer to the proposed method as unsupervised pansharpening with self-attention mechanism, abbreviated as UP-SAM.

The contribution of this paper is three-fold. First, we propose a self-attention mechanism where the detail extraction and injection functions are predicted based on the spectral characteristics of the pixels themselves. Second, a stacked self-attention network is proposed, where the attention representation layer is based on the stick-breaking structure that naturally identifies the spectral characteristics of mixed pixels with sub-pixel accuracy. Third, the detail extraction functions are estimated based on the attention representations to extract high-frequency details, and such details are injected on the representations of LR MSI with sub-pixel accuracy. The injection coefficients are spatially variant according to the attention representations. UP-SAM is an unsupervised approach designed in a deep-learning framework without the need of a large training dataset, such that it  works effectively on different types of images and reconstructs the HR MSI with more details and less spectral distortion.

The rest of the paper is organized as follows. Sec.~\ref{sec:formulate} provides the general formulation of the pansharpening problem and discusses the rationale of the proposed approach. Sec.~\ref{sec:proposed} elaborates on the proposed UP-SAM method. Sec.~\ref{sec:exp} performs comprehensive evaluations with both traditional and deep-learning based approaches. Conclusions are drawn in Sec.~\ref{sec:conclusion}.

\begin{figure*}[htp]
	\begin{minipage}{0.5\linewidth}
		\includegraphics[width=1\linewidth]{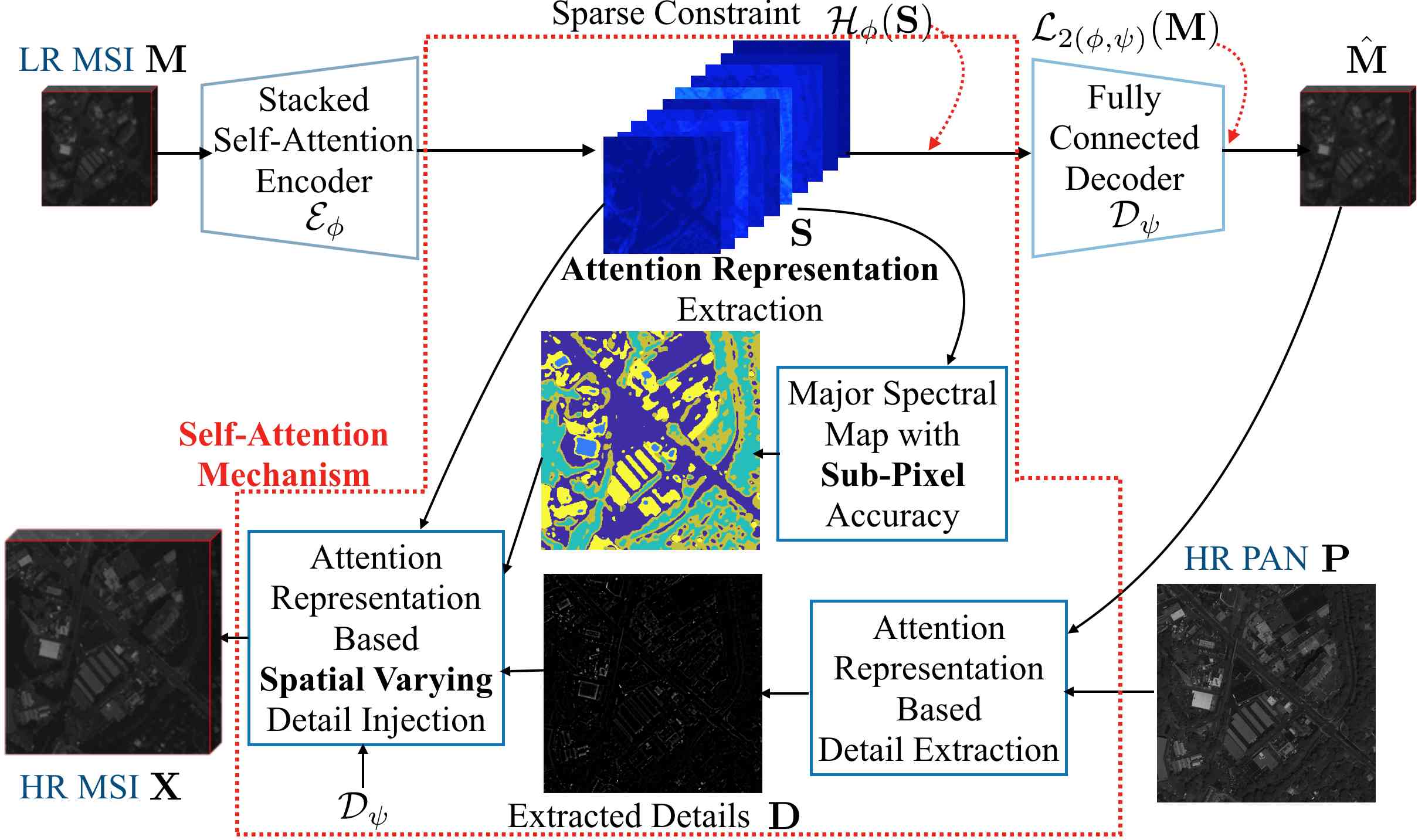}
		\caption{Flowchart of the proposed unsupervised pansharpening based on self-attention mechanism (UP-SAM).}
		\label{fig:flow}
	\end{minipage}
	\begin{minipage}{0.5\linewidth}
		\includegraphics[width=1\linewidth]{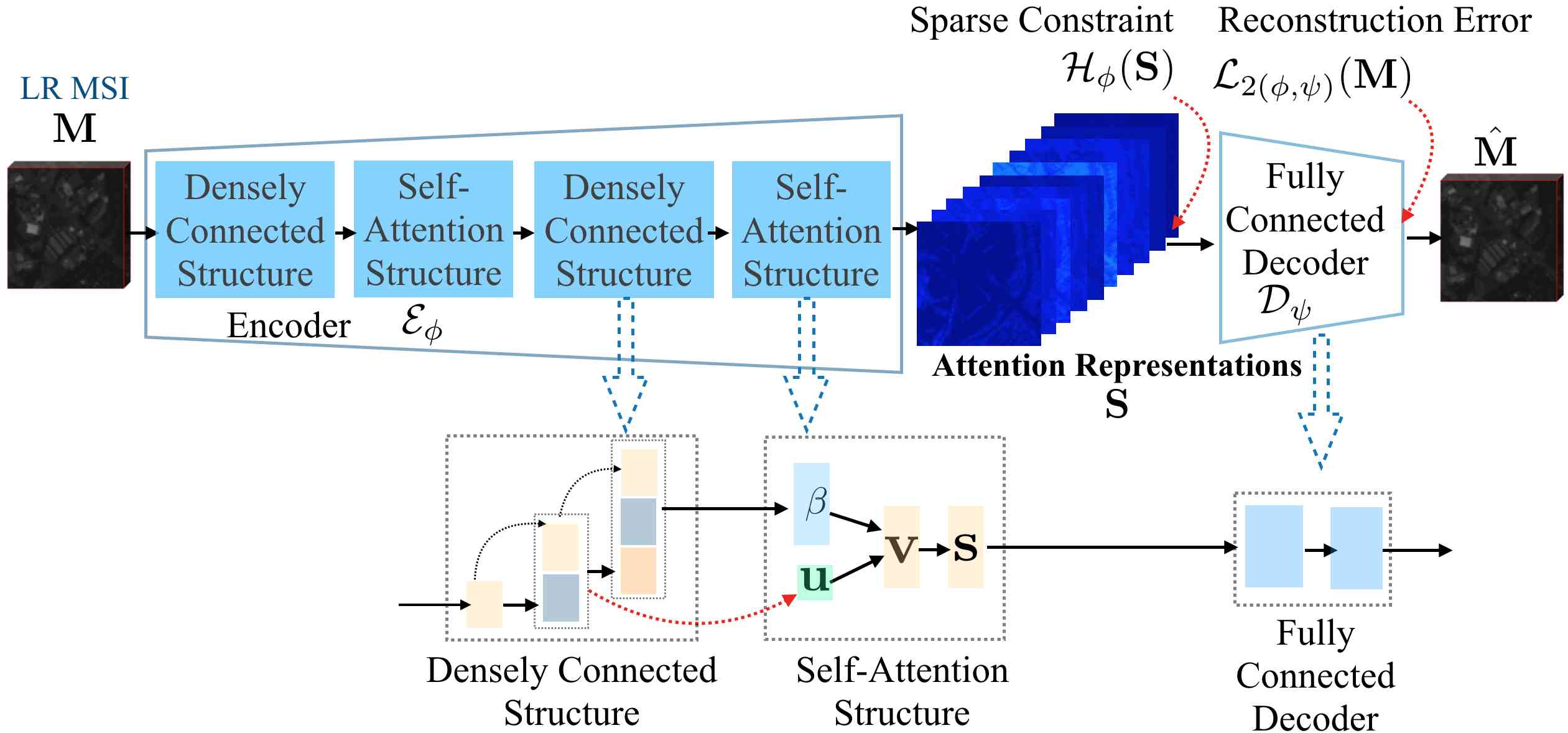}
		\caption{Illustration of the proposed stacked self-attention network.}
		\label{fig:flow_extract}
	\end{minipage}
\end{figure*}
\section{Problem Formulation and Motivations}
\label{sec:formulate}
Given the low-spatial resolution multispectral image (LR MSI), $\mathbf{M} \in \mathbb{R}^{m\times n\times L}$, where $m$, $n$ and $L$ denote the width, height, and number of spectral bands of the MSI, respectively, and the corresponding high-spatial resolution panchromatic image (HR PAN), $\mathbf{P} \in \mathbb{R}^{M \times N}$, where $M$ and $N$ denote the width and height of the PAN, respectively, the goal of pansharpening is to reconstruct the HR MSI, $\mathbf{X} \in \mathbb{R}^{M\times N \times L}$ with both high spectral and spatial resolution. The super-resolution resize factor is defined as $r=M/m=N/n$.

To facilitate the subsequent formulation, we use $\mathbf{M}_k$ to indicate the $k$th band of the MSI, where $k\in{1,\cdots,L}$. All the traditional methods as well as the state-of-the-art deep learning based approaches can be formulated as 
\begin{equation}
\mathbf{X}_k = \uparrow{\mathbf{M}}_k + g_k({d}_k(\uparrow{\mathbf{M}}_k, \mathbf{P})),
\end{equation}
where $\uparrow{\mathbf{M}}_k\in \mathbb{R}^{M\times N}$ is the upscaling version of $\mathbf{M}_k$ with the super-resolution factor $r$. $d_k$ is the detail extraction function and $g_k$ denotes the injection function.

For traditional methods, most detail extraction functions can be summarized as 
\begin{equation}
d_k(\uparrow{\mathbf{M}}_k, \mathbf{P}) = \mathbf{P}-p_{low}(\uparrow{\mathbf{M}}_k),
\label{equ:detail}
\end{equation}
where $p_{low}(\uparrow{\mathbf{M}}_k)$ denotes the function that generates the low-spatial resolution version of PAN from $\uparrow{\mathbf{M}}_k$. 
And the injection function can be summarized as
\begin{equation}
g_k(\uparrow{\mathbf{M}}_k, \mathbf{P}) = G_k\circ d_k(\uparrow{\mathbf{M}}_k, \mathbf{P}),
\end{equation}
where $G_k$ represents the injection coefficient for the $k$th band and $\circ$ denotes the element-wise multiplication. Previous researchers have shown that the optimal $G_k$ should be spatial varying according to different spectral characteristics.However, with the wide existence of mixed pixels in satellite images, it is difficult to achieve optimal performance based on the spatial segments derived from raw pixel spectra of the LR MSI.

For deep learning based methods, the feature extraction function $d_k$ and the injection function $g_k$ are defined by the weights of the deep network. And the weights are updated according to the carefully designed objective functions given the ground truth HR MSI. However, the ground truth HR MSI is generally unavailable. Furthermore, different satellite sensors usually possess different mapping characteristics, resulting in the learned detail extraction and injection functions not performing optimally on images taken from a sensor never seen before, thus introducing blur and artifacts to the HR MSI.

In summary, in order to reconstruct accurate HR MSI, the key is to predict spatial varying detail extraction and injection functions with attention given to the pixels' own spectral characteristics at the sub-pixel accuracy.

\section{Proposed Approach}
\label{sec:proposed}
We propose an unsupervised pansharpening approach based on the self-attention mechanism, as shown in Fig.~\ref{fig:flow}. The architecture mainly consists of three components that handle the tasks of 1) attention representation extraction, 2) detail extraction, and  3) detail injection. The first component is constructed with stacked self-attention network powered with the stick-breaking structure, which extracts the attention-based representations from LR MSI with sub-pixel accuracy. To reconstruct accurate details, the second component extracts the high-frequency details according to the extracted attention representations. And the third component injects the details with spatial varying coefficients estimated from the attention representations to reconstruct the HR MSI. Since both the detail extraction and injection components rely on the attention-based representation, in the following section, we first introduce the proposed stacked self-attention network.

\subsection{Extraction of Attention Representation based on Stacked Self-Attention Network}

As discussed above, most existing works on detail injection achieve spatial-varying mapping based on image  segmentation,~\ie, different segments have different injection functions. Popularly used segmentation algorithms include k-means clustering~\cite{pelleg2000x}, binary partition tree~\cite{salembier2000binary}, and mean-shift clustering~\cite{comaniciu2002mean}. However, due to the fact that each pixel in LR MSI tends to cover more than one constituent material, what we need is for the injection function to perform at sub-pixel accuracy.

One solution to the above challenge is to introduce unmixing based methods. State-of-the-art unmixing methods~\cite{li2016robust, qu2018udas,tang2017integrating,luo2018bilinear} could successfully estimate the distributions of different materials (or abundance) in making up the mixed pixel at a specific location, thus representing spectral distribution details at the sub-pixel level~\cite{xu2016using,qu2018hyperspectral}. However, unmixing approaches work the best, mostly, on hyperspectral images. It is nontrivial to adopt the unmixing methods on multispectral images acquired by remote sensors, because the number of spectral bands (usually 4 or 8) of MSI is usually much fewer than that of the constituent materials. 

To address these challenges of pansharpening, we propose an unsupervised stacked self-attention network, which could extract representations (\ie, abundance) from LR MSI with limited number of spectral bands. 

In computer vision, the concept of attention model was originally designed in the context of neural machine translation~\cite{bahdanau2014neural}. It has since been widely adopted in deep learning networks  for various applications~\cite{xu2015show,gregor2015draw, vaswani2017attention, chen2017pixelsnail, zhang2018self}. Attention model allows the network to pay attention to only part of the image or sentence. This is usually realized by applying a weight vector, with which the network pays more attention on the positions carrying higher weights and ignores positions with  lower weights. The weights are usually learned during training in a supervised way. 

Different from other attention models, in our network design, we do not need to introduce additional weights as attention weights --- the representation layer itself naturally serves as the attention weights of the spectral characteristics of each pixel. The designed structure is illustrated in Fig.~\ref{fig:flow_extract}.  In this work, we follow the majority of the previous works and 
assume a linear mixing model. The observed spectrum at a single pixel $\mathbf{m}\in\mathbb{R}^{1\times L}$ can be constructed as a linear combination of $c$ spectral signatures~\cite{bioucas2012hyperspectral}, as expressed in Eq. \eqref{equ:linear}

\begin{equation}
\mathbf{m} = \mathbf{s}{\Psi},
\label{equ:linear}
\end{equation}
where ${\Psi}\in\mathbb{R}^{c\times L}$ denotes the $c$ spectral signatures, each of which has $L$ spectral bands, and $\mathbf{s}\in\mathbb{R}^{1\times c}$ denotes the proportional coefficients of the spectral signatures in the given pixel. $\mathbf{s}$ should satisfy two physical constraints, the non-negative and sum-to-one constraints~\cite{bioucas2012hyperspectral,qu2018udas}. 

As shown in Fig.~\ref{fig:flow_extract}, the representation $\mathbf{s}$ in the network acts as the mixing coefficients extracted by the encoder network $\mathcal{E}_{\phi}$, and the decoder network $\mathcal{D}_{\psi}$ acts as the spectral signatures. Following the work of~\cite{sethuraman1994constructive, nalisnick2017stick, qu2018unsupervised}, we adopt the stick-breaking structure in the representation layer to encourage the representations to meet the two physical constraints, \ie, non-negative and sum-to-one. The stick-breaking process can be illustrated as breaking a unit-length stick into $c$ pieces, and the length of a single piece, $s_j$, can be expressed by
\begin{equation}
s_j =\left\{
\begin{array}{ll}
v_1 \quad & \text{for} \quad j = 1\\
v_j\prod_{o<j}(1-v_o) \quad &\text{for} \quad j>1 , 
\end{array}\right.
\label{equ:stick}
\end{equation}
where $v_j$ is sequentially broken from the remaining length $\prod_{o<j}(1-v_o)$. $v_j$ is drawn from a Kumaraswamy distribution, \ie, $v_j\sim \text{Kuma}(u, 1,\beta)$ as shown in Eq.~\eqref{equ:draw}
\begin{equation}
v_j\sim (1-(1-u^\frac{1}{\beta})).
\label{equ:draw}
\end{equation}
Then there are two parameters used to extract representations $\mathbf{s}$,~\ie, $u$ and $\beta$, both of which are hidden layers in the encoder of the network. 
A softplus activation function is adopted on the layer $\beta$ due to its non-negative property, and a sigmoid is used to map $u$ into the $(0,1)$ range at the layer $\mathbf{u}$. More details of the stick-breaking structure can be found in~\cite{nalisnick2017stick} and~\cite{qu2018unsupervised}. 

Since $\mathbf{s}$ indicates how the source spectra are combined in a given pixel, it acts as the attention weights, which imply the importance of each spectral signature in a given pixel.  The representation layer is referred to as the \textit{self-attention structure} in the network, and the representations are referred to as the \textit{attention representations}. Note that if we extract $s_j$, $j=1,\cdots,c$ from all the pixels in the image, it would indicate the distribution of the $j$th spectral signature over the whole image. Therefore, the $s_j$'s of the entire image form the $j$th representation map $\mathbf{S}_j$ in the network, providing sub-pixel representation accuracy from the mixture LR MSI. 

One potential issue in extracting the attention representation is that the number of spectral signatures $c$ in MSI is much larger than that of the spectral bands $L$,~\ie, $c\gg L$, making it prone to extracting poor representations.To solve this problem, we first stack two self-attention structures together, as shown in Fig.~\ref{fig:flow_extract}, where the first structure has larger number of nodes than that of the second one. The stacked structure effectively increases the representative power of the network. In addition, the entropy function~\cite{huang2018sparse} is adopted to reinforce the sparsity of the representation layer. Let $\hat{s}_j = \frac{\vert s_j \vert}{\Vert \mathbf{s} \Vert}$, for each pixel, the entropy function is defined as
\begin{equation}
\mathcal{H}_{\phi}(\mathbf{s}) = -\sum_{j=1}^{c}\hat{s}_j 
\log\hat{s}_j .
\label{equ:entropyfun}
\end{equation}

We did not adopt the popular convolution layer with shared kernels in our network, because it focuses more on extracting features from local neighborhood. In our problem, non-local regions may share similar spectral characteristics. Therefore, we could not use shared kernels for the entire image. Instead, we adopt the more general fully connected layers in the network, with densely connected structure~\cite{huang2017densely} in the encoder.

The objective functions of the proposed network architecture can be expressed as
\begin{equation}
\begin{split}
\mathcal{L}(\phi, \psi) &= \sum_{i=1}^{m}\sum_{j=1}^{n}(\Vert\mathcal{D}_\psi(\mathcal{E}_{\phi}(\mathbf{m}_{i,j}))-\mathbf{m}_{i,j}\Vert_2\\
&+\lambda \mathcal{H}_{\phi}(\mathcal{E}_{\phi}(\mathbf{m}_{i,j}))),
\end{split}\label{equ:objhsi}
\end{equation}
where $\mathbf{m}_{i,j}$, $1\leq i\leq m$, $1\leq j\leq n$ denotes a single pixel in MSI. 
The first component of Eq.~\ref{equ:objhsi} is the reconstruction error $\mathcal{L}_{2(\phi, \psi)}$ and the second one denotes the sparse constraint $\mathcal{H}_{\phi}$. $\Vert \cdot \Vert_2$ denotes the $l_2$ norm. The network is optimized with back-propagation illustrated with red dashed lines in Fig.~\ref{fig:flow_extract}.

\subsection{Attention Representation based Detail Extraction}
\begin{figure*}[htbp]
	\begin{center}
		\begin{minipage}{0.9\linewidth}
			\subfloat[Spectrals]{\includegraphics[width=0.11\linewidth]{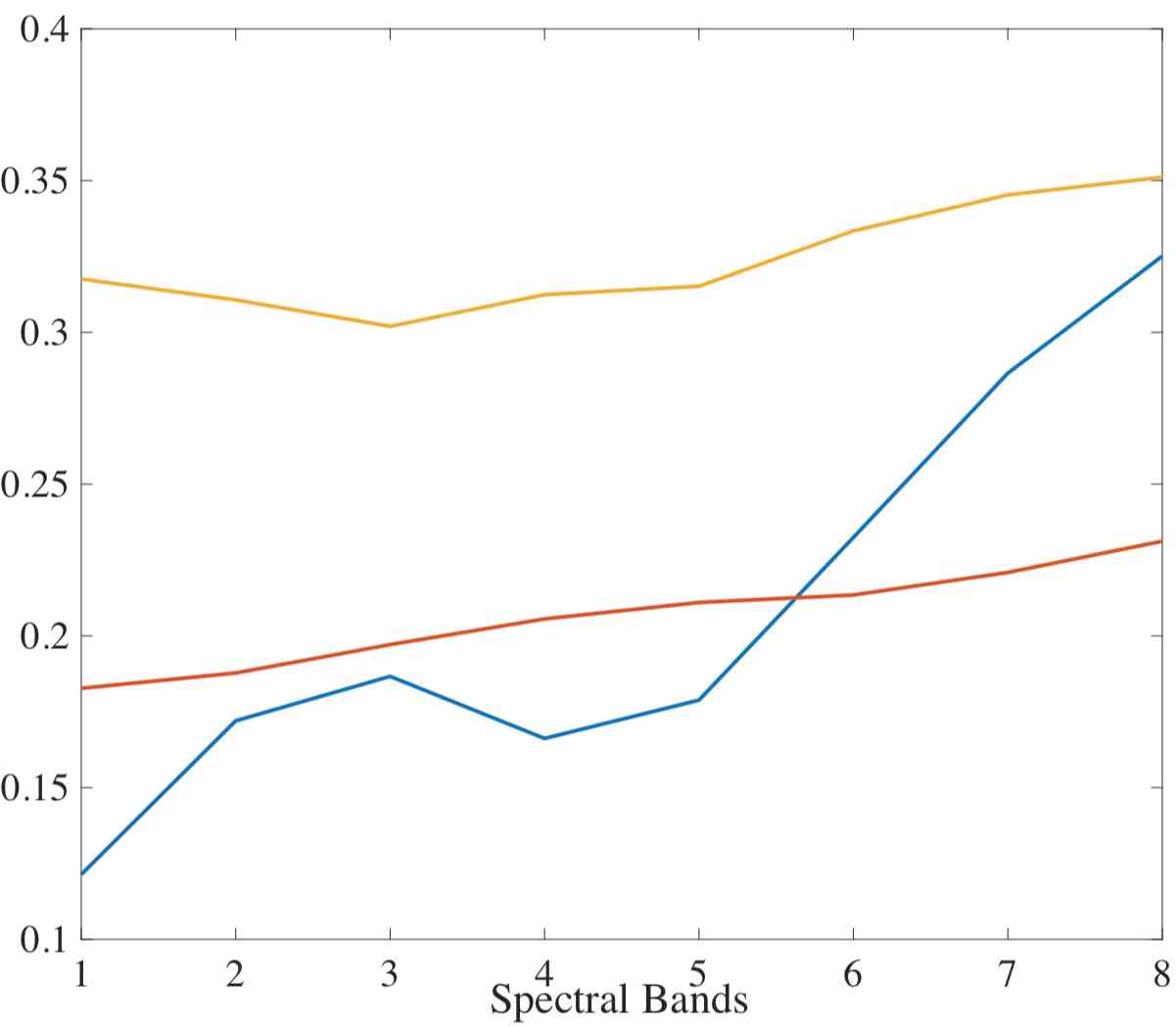}\label{fig:toy:a}}\hfill
			\subfloat[MSI]{\includegraphics[width=0.10\linewidth]{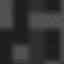}\label{fig:toy:b}}\hfill
			\subfloat[$\mathbf{S}_1$]{\includegraphics[width=0.12\linewidth]{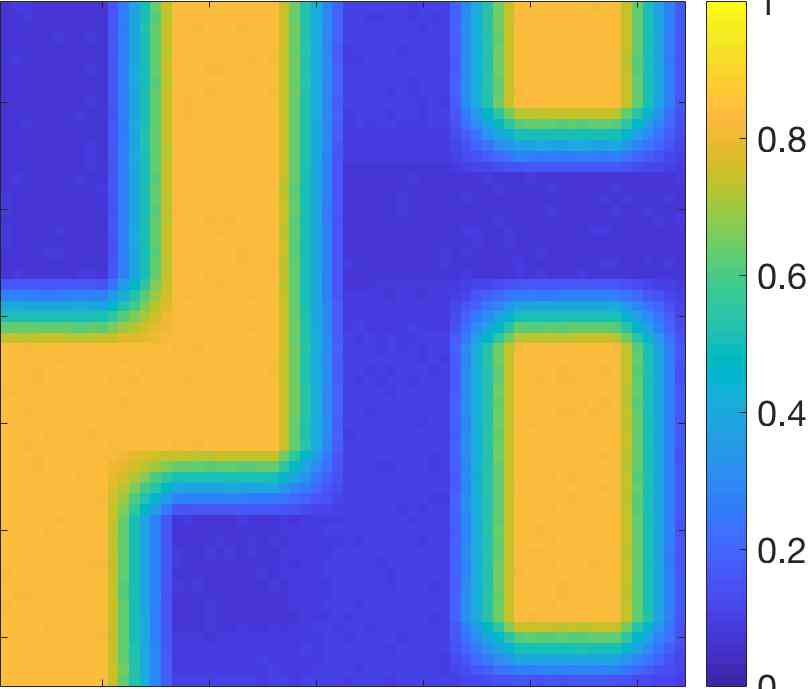}\label{fig:toy:c}}\hfill
			\subfloat[$\mathbf{S}_2$]{\includegraphics[width=0.12\linewidth]{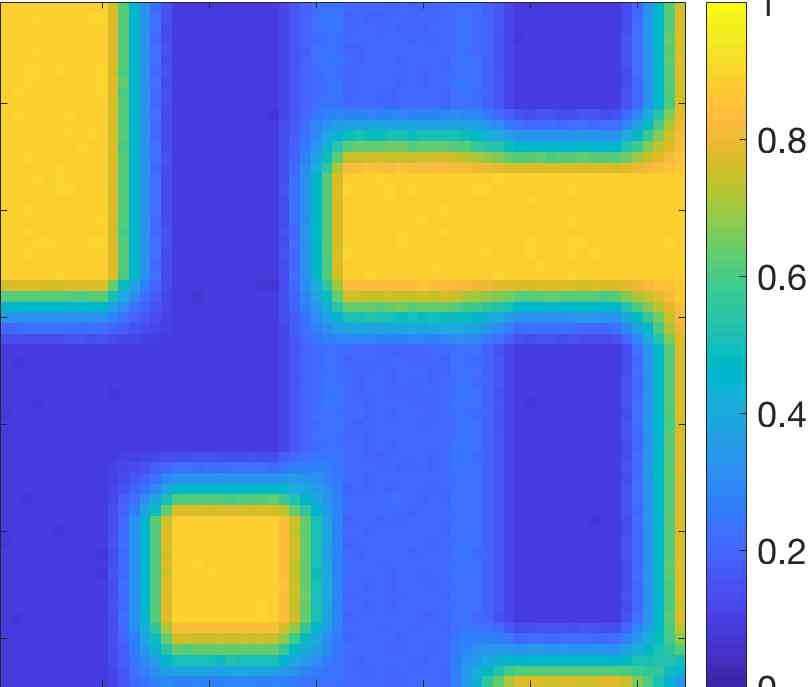}\label{fig:toy:d}}\hfill
			\subfloat[$\mathbf{S}_3$]{\includegraphics[width=0.12\linewidth]{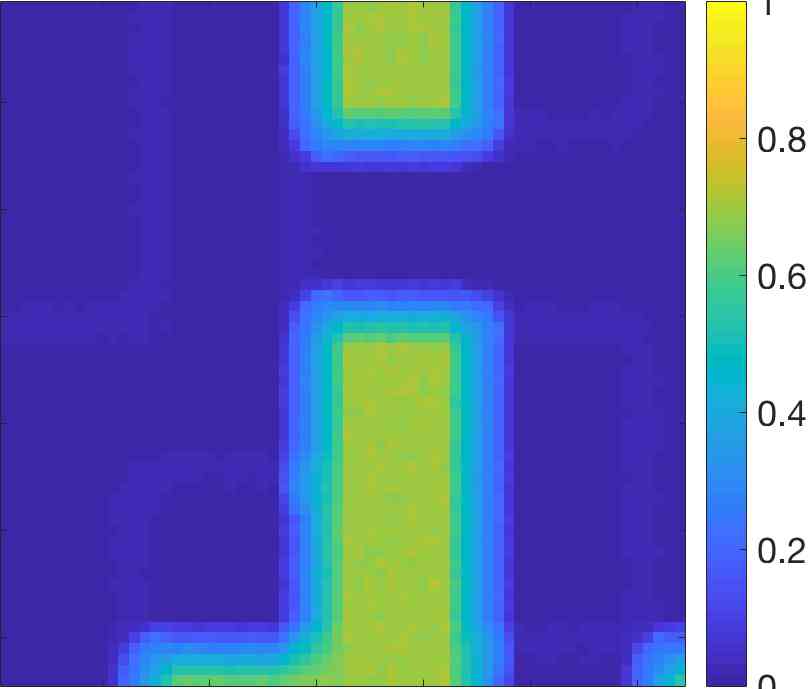}\label{fig:toy:e}}\hfill
			\subfloat[$\mathbf{S}_4$]{\includegraphics[width=0.12\linewidth]{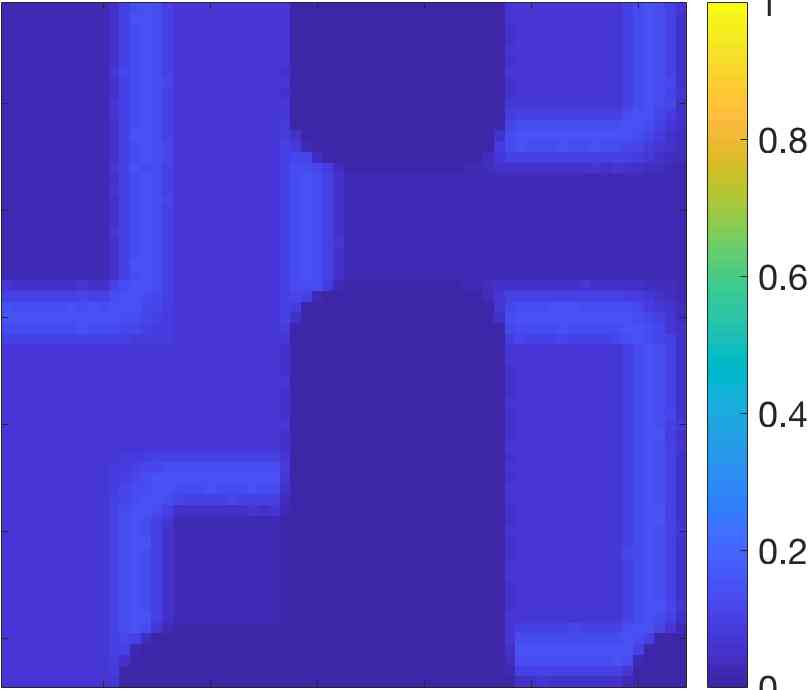}\label{fig:toy:f}}\hfill
			\subfloat[Ours]{\includegraphics[width=0.10\linewidth]{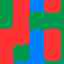}\label{fig:toy:g}}\hfill
			\subfloat[K-means]{\includegraphics[width=0.10\linewidth]{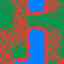}\label{fig:toy:h}}\hfill
			\caption{Toy Example. (a) Spectral signatures. (b) MSI with mixed pixels. (c)-(f) The four attention representation maps 1. (g) Major Spectral Map. (h) Cluster results with K-means.}
			\label{fig:toy}
		\end{minipage}
	\end{center}
\vspace{-3mm}
\end{figure*}
With the stacked self-attention model and the extracted attention representations (from MSI) with sub-pixel accuracy, both detail extraction and injection functions can be developed based on such representations to increase the reconstruction accuracy. Therefore, to better capture the high-frequency details with respect to the attention representations, we extract the details based on the reconstructed MSI, which is generated with the representation and the decoder of the network.
To extract such details, we need to first generate the synthetic low-resolution PAN, as indicated in Eq.~\eqref{equ:remsi}.

Following previous works~\cite{aiazzi2007improving,thomas2008synthesis}, we assume PAN is a linear combination of different bands of MSI at the same corresponding resolution. The original PAN $\mathbf{P}$ is first down-sampled through nearest neighborhood to an image at reduced-resolution with the same spatial dimension as that of the MSI,  
$\mathbf{P}_{low}\in\mathbb{R}^{m\times n}$,
using low-pass filtering and decimation~\cite{aiazzi2007improving}. Based on the assumption, we have
\begin{equation}
\mathbf{P}_{low} = \sum_{k=1}^{L} \alpha_k \mathbf{M}_k + b,
\label{equ:regress}
\end{equation}
where $\mathbf{M}_k\in\mathbb{R}^{m\times n}$ is the $k$th band of the given LR MSI. Then the coefficients $\mathbf{\alpha} = [\alpha_1,\cdots, \alpha_L,b]$ can be estimated by linear regression~\cite{aiazzi2007improving}. With the coefficient $\mathbf{\alpha}$, we can generate the low-spatial resolution PAN $\hat{\mathbf{P}}_{low}\in\mathbb{R}^{M\times N}$ with
\begin{equation}
\hat{\mathbf{P}}_{low} = \sum_{k=1}^{L} \alpha_k \uparrow\hat{\mathbf{M}}_k + b, 
\label{equ:regress2}
\end{equation}
where $\uparrow\hat{\mathbf{M}}_k\in\mathbb{R}^{M\times N}$ is the $k$th band of the up-sampled $\hat{\mathbf{M}}$, and $\hat{\mathbf{M}}$ is the reconstructed LR MSI from the network generated by the attention representations and decoder weights as expressed in Eq.~\eqref{equ:remsi},
\begin{equation}
\hat{\mathbf{M}} = \mathcal{D}_{\psi}(\mathbf{S}).
\label{equ:remsi}
\end{equation}
where $\mathbf{S} = \mathcal{E}_{\phi}(\mathbf{M})$, $\mathcal{E}_{\phi}$ and $\mathcal{D}_{\psi}$ denote the encoder and decoder of the network, respectively.

Then the high-frequency detail matrix is extracted with
\begin{equation}
\mathbf{D} = \mathbf{P} - \hat{\mathbf{P}}_{low}. 
\label{equ:detail2}
\end{equation}

\subsection{Attention Representation based Spatial Varying Detail Injection}
Different from the common practice in detail injection, either spatial-invariant or spatial-variant, since the proposed UP-SAM provides sub-pixel representation accuracy using the attention mechanism, correspondingly, the detail injection process in UP-SAM is conducted on each representation map, $S_i$, $i=1,\cdots,c$ with $c$ being the number of spectral signatures, instead of on each spectral band, $M_k$, $k=1,\cdots,L$ with $L$ being the number of spectral bands in MSI.

We start the design by investigating the simplest case where the injection coefficients are spatially invariant. We adopt the projective injection model~\cite{aiazzi2007improving,restaino2017context,vivone2018full} to estimate the injection coefficients for each representation map, which can be expressed as
\begin{equation}
\tilde{G_i} = \frac{cov(\uparrow\mathbf{S}_i,{\hat{\mathbf{P}}}_{low})}{var({\hat{\mathbf{P}}}_{low})},
\end{equation}
where $\tilde{G_i}$ is the $i$th coefficient for the $i$th representation map, $i=1,\cdots,c$. The operation $cov(\cdot)$ denotes the covariance of two images and $var(\cdot)$ denotes the variance of an image. $\mathbf{S}_i$ denotes the $i$th representation map derived from $\mathbf{S}=\mathcal{E}_{\phi}(\mathbf{M})$ and $\uparrow$ is the up-sampling operation. 

To further improve the reconstruction accuracy, instead of adopting global injection coefficients, we estimate the injection coefficients for each mixed pixel based on the attention weights that indicating its major constituent spectral signatures. Let $\mathbf{s}\in\mathbb{R}^{1\times c}$ denote the attention representation of a given pixel,~\ie, $\mathbf{s} = \{s_1,\cdots,s_j,\cdots,s_c\}$, where $s_j$ denotes the proportion of the $j$th spectral signature in making up the mixed pixel. For each pixel, we find the ``index'' of largest proportion, which indicates the most important constituent spectral signature for a given pixel. Mathematically, it can be expressed as
\begin{equation}
t =\mathop{\arg\max}_{j}\{s_1,\cdots,s_j,\cdots,s_c\}
\label{equ:max}
\end{equation}

With the above equation, we are able to identify the major spectral signature for each pixel. The index number of the major spectral signature in the entire image is used to construct a map for further spatial-variant injection coefficients estimation. And such map is referred to as the major spectral index map (MSIM), as shown in Fig.~\ref{fig:toy:g}. 

In Fig.~\ref{fig:toy}, a toy example is provided to explain the process. We use the linear mixing of three spectral signatures of 8 bands ($L= 8$), as shown in Fig.~\ref{fig:toy:a}, to generate the synthetic MSI in Fig.~\ref{fig:toy:b} with Gaussian white noise of a signal-to-noise ratio (SNR) of 30dB. 
Given this MSI as the input, four attention representation maps (i.e., $c=4$) are extracted by the stacked self-attention network, as shown in \crefrange{fig:toy:c}{fig:toy:f}. Each representation map demonstrates the distribution (or proportion) of the corresponding spectral signature in making up the entire image, \eg, $\mathbf{S}_1$ shows the proportion of the first spectral signature. Then we can identify the indices of the most important spectrum for each pixel and show the MSIM in Fig.~\ref{fig:toy:g}. For example, the locations marked with the red color indicate the most important spectrum of that pixel is the first spectrum. Similarly, the locations marked with the green and the blue colors indicate that the most important spectrum at the corresponding pixels are the second and the third spectra, respectively. Note that since the MSI consists of three spectral signatures and we extract four attention representation maps, the fourth map, $\mathbf{S}_4$, mainly consists of outliers (or noise) with small proportions as shown in Fig.~\ref{fig:toy:f}. Therefore, those smaller proportions tend to be filtered out with the selection process in Eq.~\eqref{equ:max}. Thus, the MSIM contains only three index numbers which are marked with three different colors. For comparison, we also show the clustering results on the raw LR MSI with K-means in Fig.~\ref{fig:toy:h}. We can observe that compared to the MSIM in Fig.~\ref{fig:toy:g}, the result of K-means cannot identify the spectra well due to the existence of noise and mixed pixels. However, the proposed method is able to identify the spectral signatures with sub-pixel accuracy.

Given the MSIM, the spatial-variant injection coefficients can then be calculated using Eq. (15), where the coefficients vary for both the representation maps and pixels with different spectral characteristics. For the $i$th representation map, the pixels carry the same index number (or same color as shown in the toy example in Fig. 3) share the same coefficient determined by the corresponding image subsets. Let $\uparrow\mathbf{M}^{\{t\}}$ denote the pixels in a subset of image $\uparrow\mathbf{M}$, whose indexes of the most important constituent spectral signature are $t$, the injection coefficient for their $i$th representation map is then estimated with
			\begin{equation}
			G_i^{\{\uparrow\mathbf{M}^{\{t\}}\}} = 
			\frac{
				cov(
				\uparrow{\mathbf{S}_i^{\{\uparrow\mathbf{M}^{\{t\}}\}}},
				{\hat{\mathbf{P}}}_{low}^{\{\uparrow\mathbf{M}^{\{t\}}\}}
				)
			}
			{
				var(
				{\hat{\mathbf{P}}}_{low}^{\{\uparrow\mathbf{M}^{\{t\}}\}}
				)
			},
			\label{equ:variance}
			\end{equation}
			where $t$ denotes the index of the most important constituent spectral signature, and $\uparrow{\mathbf{S}_i^{\{\uparrow\mathbf{M}^{\{t\}}\}}}$ and ${\hat{\mathbf{P}}}_{low}^{\{\uparrow\mathbf{M}^{\{t\}}\}}$ denote the subset of the representation map and the low-spatial resolution PAN (generated by Eq.~(10)) that corresponding to $\uparrow\mathbf{M}^{\{t\}}$, respectively.

The final reconstructed $\mathbf{X}$ can then be estimated by projecting the detail injected attention representations back to the image domain with
\begin{equation}
\mathbf{X} = \mathcal{D}_{\psi}(\uparrow{\mathcal{E}_{\phi}}(\mathbf{M}) + G\circ \mathbf{D})
\label{equ:reconstruct}
\end{equation}
where $\mathbf{D}$ is the extracted high-frequency details in Eq.~\eqref{equ:detail2}. $G$ is the spatially-variant injection coefficient set estimated with Eq.~\eqref{equ:variance}. $\circ$ denotes the element by element multiplication. $\mathcal{E}_{\phi}$ and $\mathcal{D}_{\psi}$ are the encoder and decoder of the network, respectively.

\subsection{Implementation Details}
\begin{algorithm}[htbp]
	\caption{Unsupervised Pansharpening based on Self-Attention Mechanism (UP-SAM)}
	\begin{algorithmic}
		\renewcommand{\algorithmicrequire}{\textbf{Input:}}
		\renewcommand{\algorithmicensure}{\textbf{Output:}}
		\REQUIRE LR MSI $\mathbf{M}\in\mathbb{R}^{m\times n \times L}$, HR PAN $\mathbf{P}\in \mathbb{R}^{M\times N}$
		\ENSURE The Pansharpened HR MSI $\mathbf{X}\in\mathbb{R}^{M\times N \times L}$
		\\\textbf{Begin}:\\
		1) Attention representation extraction: Given $\mathbf{M}$, the attention representations $\mathbf{S} = \mathcal{E}(\mathbf{M})\in\mathbb{R}^{m\times n\times c}$ is extracted with the stacked self-attention network using the objective function as in Eq.~\ref{equ:objhsi}.\\

		2) Low-resolution PAN synthesis: Given $\mathbf{P}$, first downsample it to  $\mathbf{P}_{low} \in\mathbb{R}^{m\times n}$, then estimate the weights $\mathbf{\alpha} = {\alpha_1,\cdots,\alpha_L,b}$ in Eq.~\eqref{equ:regress} according to the linear regression model~\cite{aiazzi2007improving}, finally generate the low-spatial resolution PAN $\hat{\mathbf{P}}_{low}\in\mathbb{R}^{M\times N}$ using Eq.~\ref{equ:regress2}.\\
		3) Detail extraction: Extract the high-frequency detail, $\mathbf{D}$, based on the attention representations according to Eqs.~\eqref{equ:remsi} and ~\eqref{equ:detail2}.\\

		4) Space-variant injection estimation: Estimate the injection coefficients, $G_i^{\{t\}}$ according to the subset of attention representations with Eqs.~\eqref{equ:max} and~\eqref{equ:variance}.\\

		5) HR MSI reconstruction: Reconstruct the HR MSI, $\mathbf{X}$, according to Eq.~\eqref{equ:reconstruct}.
		\begin{equation}
        \mathbf{X} = \mathcal{D}_{\psi}(\uparrow{\mathcal{E}_{\phi}}(\mathbf{M}) + G\circ \mathbf{D})\nonumber
		\end{equation}
		\\\textbf{End}\\
	\end{algorithmic}
	\label{alg}
\end{algorithm}

The whole procedure of the proposed UP-SAM is summarized in Algorithm~\ref{alg}.
The detail structure of the stacked self-attention encoder and fully connected decoder is shown at the lower part of Fig.~\ref{fig:flow_extract}.
The network consists of fully-connected layers. Each block denotes a single fully-connected layer, and the stacked block denotes the densely connected structure, where each layer is connected with its subsequent layers. Mathematically, a single fully-connected layer is defined by~\eqref{equ:fully}.
\begin{equation}
\mathbf{y} = \mathbf{W}^T\mathbf{x}+\mathbf{b}
\label{equ:fully}
\end{equation}
where $\mathbf{x}$ denotes the input and $\mathbf{y}$ denotes the output of the fully-connected layer, respectively. $\mathbf{W}$ and $\mathbf{b}$ denote the network weights. 

As elaborated in Fig.~\ref{fig:flow_extract}, in order to extract the attention representations, $\mathbf{S}$, 
given the LR MSI, $\mathbf{M}$, two stacked self-attention structures are used as the encoder, where $\mathbf{s}$ for each pixel is constructed with $v$, and $v$ is drawn based on $u$ and $\beta$ learned from the densely connected structures, as in Eqs.~\eqref{equ:stick} and \eqref{equ:draw}. There are 3 hidden layers in the first densely connected network and each layer contains 3 nodes. The outputs of this network serve as the input to the first self-attention structure where the two parameters, $u$ and $\beta$, as used in Eq.~\eqref{equ:draw}, are estimated. Since each mixed pixel in $\mathbf{M}$ has its own weights estimation, we denote them using vector format $\mathbf{u}^1$ and $\mathbf{\beta}^1$ with the superscript indicating the first set of the self-attention network. We use a 20-node network to learn $\mathbf{u}^1$ and a 1-node network to learn $\mathbf{\beta}^1$. Then  $\mathbf{v}^1$ can be drawn from $\mathbf{u}^1$ and $\mathbf{\beta}^1$ according to Eq.~\eqref{equ:draw}, which will yield the self-attention map, as in Eq.~\eqref{equ:stick}.  
The second densely connected network has similar settings with 3 hidden layers and each layer contains 3 nodes. To extract more representative features, the second self-attention structure has less nodes, with 
$\mathbf{u}^2$ having 10 nodes and $\mathbf{\beta}^2$ having 1 node in the network. The decoder of the network has 2 layers without bias and each layer has 10 nodes. 

The network is updated by back-propagation illustrated with the red dashed lines as in Fig.~\ref{fig:flow_extract}. There is only one free parameter in the network,~\ie, the sparse regularization $\lambda$. We set $\lambda = 0.001$, such that it can encourage the representation layer to be sparse, while at the same time reconstructing the image with less artifact and blur. Note that, the entropy function defined in Eq.~\eqref{equ:entropyfun} is a theoretical function that cannot be implemented in real systems since the function is not differentiable at 0. Hence, when we implement the function, we added a small perturbation to $s_j$ such that the perturbed $s_j$ will never be exactly 0.

\section{Experiments and Results} 
\label{sec:exp}
\subsection{Datesets and Experimental Setup}
To evaluate the performance of the proposed UP-SAM, we apply the method on four sets of real multispectral and panchromatic image pairs with different spectral characteristics that collected by four different sensors, including GeoEye-1 (GE), Ikonos (IK), WorldView-2 (WV2)~\cite{scarpa2018target}, and WorldView-3 (WV3). The number of spectral bands is 4 (blue, green, red, near-infrared) for the MSI taken from GeoEye-1 and Ikonos sensors, and 8 (red, red edge, coastal, blue, green, yellow, near-IR1 and near-IR2) for the MSI taken from WV2 and WV3 sensors.  More details of the datasets are shown in Table~\ref{tab:data}. 

\begin{table}[htbp]
	\caption{Dataset pairs from different sensors used in the experiments.}
	\label{tab:data}
	\centering
	\begin{tabular}{c| c c c c }
		\hline
		& GE& IK & WV2 & WV3 \\
		\hline
		Resolution of PAN & 0.46m & 0.82m & 0.46m & 0.31m \\
		Resolution of MSI & 1.84m & 3.28m & 1.84m & 1.24m \\
		\# of bands&4&4&8&8\\
		\hline
		{\# of pixels of PAN}&\multicolumn{4}{c}{$1280\times 1280 $} \\
	    {\# of pixels of MSI}&\multicolumn{4}{c}{$320\times 320$}\\
		\hline
	\end{tabular}%
\end{table}
 
The results of the proposed approach is compared with eight state-of-the-art approaches including GSA~\cite{aiazzi2007improving},  PRACS~\cite{choi2010new}, BDSD-PC~\cite{vivone2019robust}, which are component substitution (CS) based;  MTF-CBD~\cite{aiazzi2006mtf} and GLP-Reg-FS~\cite{vivone2018full}, which are multi-resolution analysis (MRA) based; GSA-Segm and GLP-Segm~\cite{restaino2017context}, which are context based;  and Target-CNN~\cite{scarpa2018target} which is deep-learning based. All these methods are reported the best performance in recent literature with available codes\footnote{http://openremotesensing.net/}~\cite{thomas2008synthesis, aiazzi2012twenty,vivone2014critical, restaino2017context, scarpa2018target, vivone2018full}. The code of BDSD-PC is offered by the author of~\cite{vivone2019robust}. Note that Target-CNN~\cite{scarpa2018target} is a supervised method, which is trained with the ground truth HR MSI of the first three datasets at reduced resolution. Therefore, it is unfair to compare the results from Target-CNN with those from the unsupervised methods without the ground truth MSI made available. But since it is a deep learning based method with the best performance, we list it here as a reference. We show that although supervised Target-CNN could achieve good quantitative values when the ground truth images are available during training, it tends to introduce artifacts when applied to unseen images or the test images at full resolution. 

Since the ground truth HR MSI is not available for the real dataset, we first evaluate the methods at reduced resolution,~\ie, the original MSI and PAN are further down-sampled to a reduced resolution with super-resolution resize factor $r=4$. The degradation procedure follows the Wald’s protocol~\cite{wald1997fusion,vivone2014critical}, where the resolution of MSI is reduced by applying a degradation filter matching the modulation transfer function (MTF) of the sensor and a decimation operator characterized by the resize factor~\cite{aiazzi2006mtf}. Similarly,  the PAN is downsampled with an ideal filter to preserve the details at the reduced resolution~\cite{vivone2014critical}. In this way, the original MSI can be adopted as ground truth MSI to evaluate the reconstruction capability of different algorithms. For quantitative comparison, four popularly used measurement metrics, including the peak signal-to-noise ratio (PSNR), the spectral angle mapper (SAM)~\cite{yuhas1992discrimination}, the erreur relative globale adimensionnelle de synthese (ERGAS)~\cite{wald1997fusion}, and $Q2^n$~\cite{garzelli2009hypercomplex}, are adopted for evaluation.

For the full-resolution validation, the quantitative results estimated with the  no-reference index~\cite{alparone2008multispectral} (QNR), its spectral component $\mathbf{D}_\lambda$ and spatial component $\mathbf{D}_S$ are adopted for evaluation. However, the assessment is challenging since the ground truth HR MSI is not available and the quantitative results rely heavily on the upsampled MSI. This tends to introduce bias in the metric calculation. For example, in the extreme case, if the reconstructed HR MSI is the same as the LR MSI, we can obtain $\mathbf{D}_\lambda=0$, which leads to a high QNR value. This explains why some methods can generate images with high QNR values but poor image quality. Therefore, both qualitative and quantitative comparisons should be considered for performance evaluation.
 
We conduct three sets of experiments to perform comprehensive evaluations on the proposed UP-SAM, including both quantitative and qualitative comparisons at reduced resolution (Sec.~\ref{sec:reduced}), at full resolution (Sec.~\ref{sec:full}), as well as a more in-depth study of the self-attention network (Secs.~\ref{sec:attention} and~\ref{sec:injection}). The computational efficiency is analyzed in Sec.~\ref{sec:time}.

\subsection{Comparison at Reduced Resolution}
\label{sec:reduced}
The first set of experiments is conducted on the four datasets at reduced resolution. In this case, the original MSI is available and can be used as the ground truth MSI for quantitative evaluations.

\begin{figure*}[htbp]
	\subfloat[\hspace{-0.5mm}GT,$\mathbf{M}$,$\mathbf{P}$]{\begin{minipage}[t]{0.095\linewidth}
		\begin{annotatedFigure}
			{\adjincludegraphics[width=\linewidth, trim={0 {.5\width} {.5\width} {.0\width}},clip]	{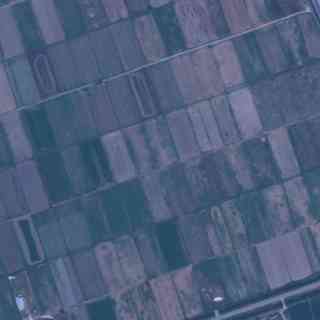}};
		\end{annotatedFigure}\vspace{0.5mm}
		\begin{annotatedFigure}
			{\adjincludegraphics[width=\linewidth, trim={0 {.5\width} {.5\width} {.0\width}},clip]	{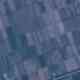}};
		\end{annotatedFigure}\vspace{0.5mm}
		\begin{annotatedFigure}
			{\adjincludegraphics[width=\linewidth, trim={0 {.5\width} {.5\width} {.0\width}},clip]	{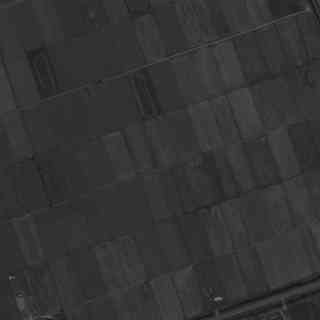}};
		\end{annotatedFigure}\vspace{0.5mm}
		\begin{annotatedFigure}
			{\adjincludegraphics[width=\linewidth, trim={0 {.0\width} {.65\width} {.65\width}},clip]	{fig/imgGE/imgGE_rd_msi_sc.jpg}};
		\end{annotatedFigure}\vspace{0.5mm}
		\begin{annotatedFigure}
			{\adjincludegraphics[width=\linewidth, trim={0 {.0\width} {.65\width} {.65\width}},clip]	{fig/imgGE/imgGE_rd_msi_lr_sc.jpg}};
		\end{annotatedFigure}\vspace{0.5mm}
		\begin{annotatedFigure}
			{\adjincludegraphics[width=\linewidth, trim={0 {.0\width} {.65\width} {.65\width}},clip]	{fig/imgGE/imgGE_rd_pan.jpg}};
		\end{annotatedFigure}	
	\end{minipage}\label{fig:geoeye_rd:a}}\hspace{0.001mm}
	\subfloat[GSA]{\begin{minipage}[t]{0.095\linewidth}
		\begin{annotatedFigure}
			{\adjincludegraphics[width=\linewidth, trim={0 {.5\width} {.5\width} {.0\width}},clip]	{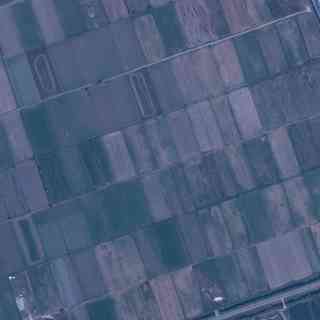}};
		\end{annotatedFigure}\vspace{0.5mm}
		\begin{annotatedFigure}
			{\adjincludegraphics[width=\linewidth, trim={0 {.5\width} {.5\width} {.0\width}},clip]	{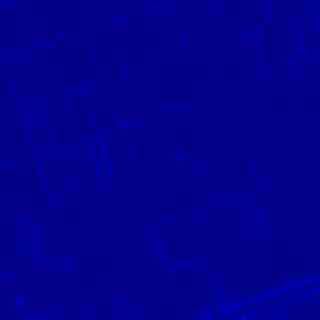}};
		\end{annotatedFigure}\vspace{0.5mm}
		\begin{annotatedFigure}
			{\adjincludegraphics[width=\linewidth, trim={0 {.5\width} {.5\width} {.0\width}},clip]	{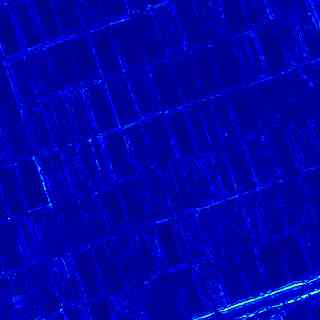}};
		\end{annotatedFigure}\vspace{0.5mm}
		\begin{annotatedFigure}
			{\adjincludegraphics[width=\linewidth, trim={0 {.0\width} {.65\width} {.65\width}},clip]	{fig/imgGE/imgGE_rd_gsa_sc.jpg}};
		\end{annotatedFigure}\vspace{0.5mm}
		\begin{annotatedFigure}
			{\adjincludegraphics[width=\linewidth, trim={0 {.0\width} {.65\width} {.65\width}},clip]	{fig/imgGE/imgGE_rmse_gsa.jpg}};
		\end{annotatedFigure}\vspace{0.5mm}
		\begin{annotatedFigure}
			{\adjincludegraphics[width=\linewidth, trim={0 {.0\width} {.65\width} {.65\width}},clip]	{fig/imgGE/imgGE_sam_gsa.jpg}};
		\end{annotatedFigure}	
	\end{minipage}\label{fig:geoeye_rd:b}}\hspace{0.001mm}
	\subfloat[PRACS]{\begin{minipage}[t]{0.095\linewidth}
		\begin{annotatedFigure}
			{\adjincludegraphics[width=\linewidth, trim={0 {.5\width} {.5\width} {.0\width}},clip]	{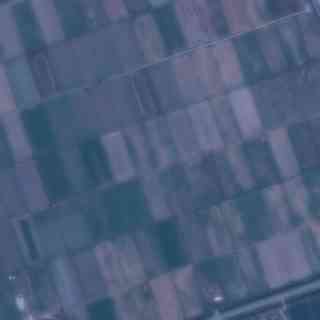}};
		\end{annotatedFigure}\vspace{0.5mm}
		\begin{annotatedFigure}
			{\adjincludegraphics[width=\linewidth, trim={0 {.5\width} {.5\width} {.0\width}},clip]	{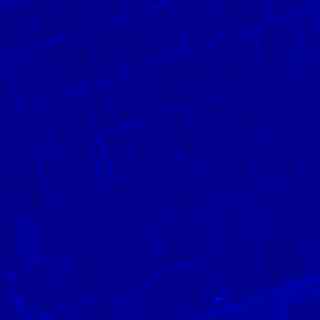}};
		\end{annotatedFigure}\vspace{0.5mm}
		\begin{annotatedFigure}
			{\adjincludegraphics[width=\linewidth, trim={0 {.5\width} {.5\width} {.0\width}},clip]	{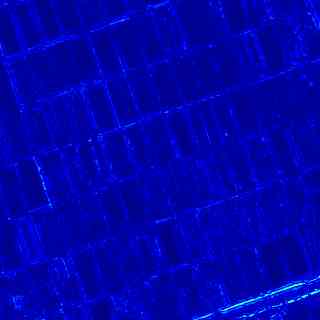}};
		\end{annotatedFigure}\vspace{0.5mm}
		\begin{annotatedFigure}
			{\adjincludegraphics[width=\linewidth, trim={0 {.0\width} {.65\width} {.65\width}},clip]	{fig/imgGE/imgGE_rd_pracs_sc.jpg}};
		\end{annotatedFigure}\vspace{0.5mm}
		\begin{annotatedFigure}
			{\adjincludegraphics[width=\linewidth, trim={0 {.0\width} {.65\width} {.65\width}},clip]	{fig/imgGE/imgGE_rmse_pracs.jpg}};
		\end{annotatedFigure}\vspace{0.5mm}
		\begin{annotatedFigure}
			{\adjincludegraphics[width=\linewidth, trim={0 {.0\width} {.65\width} {.65\width}},clip]	{fig/imgGE/imgGE_sam_pracs.jpg}};
		\end{annotatedFigure}	
	\end{minipage}\label{fig:geoeye_rd:c}}\hspace{0.01mm}
	\subfloat[BDSD-PC]{\begin{minipage}[t]{0.095\linewidth}
		\begin{annotatedFigure}
			{\adjincludegraphics[width=\linewidth, trim={0 {.5\width} {.5\width} {.0\width}},clip]	{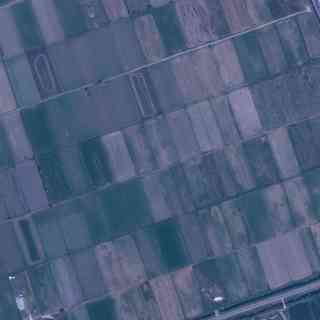}};
		\end{annotatedFigure}\vspace{0.5mm}
		\begin{annotatedFigure}
			{\adjincludegraphics[width=\linewidth, trim={0 {.5\width} {.5\width} {.0\width}},clip]	{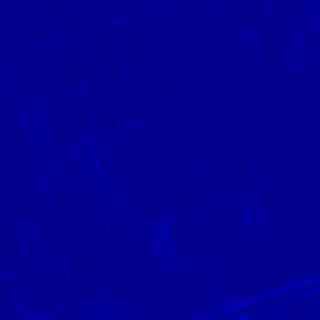}};
		\end{annotatedFigure}\vspace{0.5mm}
		\begin{annotatedFigure}
			{\adjincludegraphics[width=\linewidth, trim={0 {.5\width} {.5\width} {.0\width}},clip]	{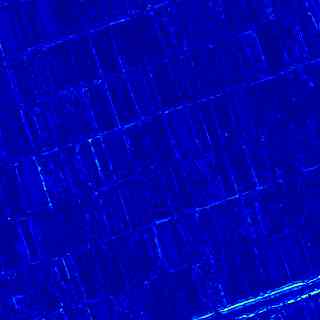}};
		\end{annotatedFigure}\vspace{0.5mm}
		\begin{annotatedFigure}
			{\adjincludegraphics[width=\linewidth, trim={0 {.0\width} {.65\width} {.65\width}},clip]	{fig/imgGE/imgGE_rd_bdsd_pc_sc.jpg}};
		\end{annotatedFigure}\vspace{0.5mm}
		\begin{annotatedFigure}
			{\adjincludegraphics[width=\linewidth, trim={0 {.0\width} {.65\width} {.65\width}},clip]	{fig/imgGE/imgGE_rmse_bdsd_pc.jpg}};
		\end{annotatedFigure}\vspace{0.5mm}
		\begin{annotatedFigure}
			{\adjincludegraphics[width=\linewidth, trim={0 {.0\width} {.65\width} {.65\width}},clip]	{fig/imgGE/imgGE_sam_bdsd_pc.jpg}};
		\end{annotatedFigure}	
	\end{minipage}\label{fig:geoeye_rd:d}}\hspace{0.01mm}
	\subfloat[MTF-CBD]{\begin{minipage}[t]{0.095\linewidth}
		\begin{annotatedFigure}
			{\adjincludegraphics[width=\linewidth, trim={0 {.5\width} {.5\width} {.0\width}},clip]	{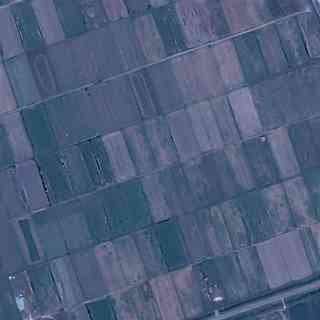}};
		\end{annotatedFigure}\vspace{0.5mm}
		\begin{annotatedFigure}
			{\adjincludegraphics[width=\linewidth, trim={0 {.5\width} {.5\width} {.0\width}},clip]	{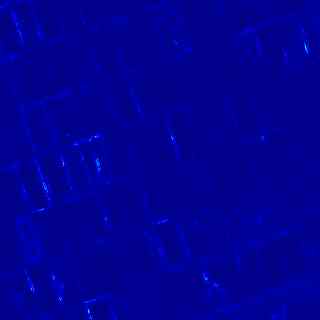}};
		\end{annotatedFigure}\vspace{0.5mm}
		\begin{annotatedFigure}
			{\adjincludegraphics[width=\linewidth, trim={0 {.5\width} {.5\width} {.0\width}},clip]	{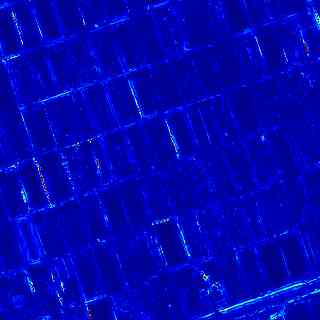}};
		\end{annotatedFigure}\vspace{0.5mm}
		\begin{annotatedFigure}
			{\adjincludegraphics[width=\linewidth, trim={0 {.0\width} {.65\width} {.65\width}},clip]	{fig/imgGE/imgGE_rd_cbd_sc.jpg}};
		\end{annotatedFigure}\vspace{0.5mm}
		\begin{annotatedFigure}
			{\adjincludegraphics[width=\linewidth, trim={0 {.0\width} {.65\width} {.65\width}},clip]	{fig/imgGE/imgGE_rmse_cbd.jpg}};
		\end{annotatedFigure}\vspace{0.5mm}
		\begin{annotatedFigure}
			{\adjincludegraphics[width=\linewidth, trim={0 {.0\width} {.65\width} {.65\width}},clip]	{fig/imgGE/imgGE_sam_cbd.jpg}};
		\end{annotatedFigure}	
	\end{minipage}\label{fig:geoeye_rd:e}}\hspace{0.01mm}
	\subfloat[\hspace{-1.1mm}GLP-Reg-FS]{\begin{minipage}[t]{0.095\linewidth}
		\begin{annotatedFigure}
			{\adjincludegraphics[width=\linewidth, trim={0 {.5\width} {.5\width} {.0\width}},clip]	{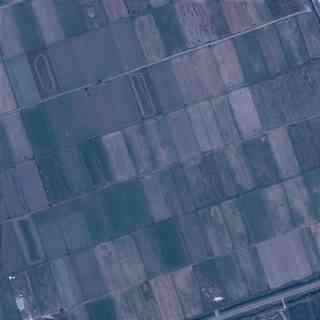}};
		\end{annotatedFigure}\vspace{0.5mm}
		\begin{annotatedFigure}
			{\adjincludegraphics[width=\linewidth, trim={0 {.5\width} {.5\width} {.0\width}},clip]	{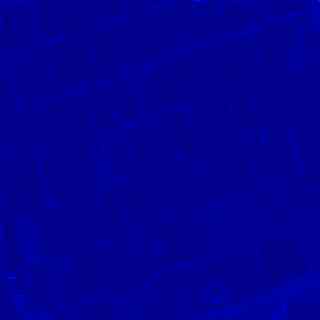}};
		\end{annotatedFigure}\vspace{0.5mm}
		\begin{annotatedFigure}
			{\adjincludegraphics[width=\linewidth, trim={0 {.5\width} {.5\width} {.0\width}},clip]	{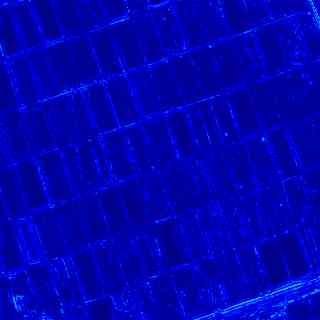}};
		\end{annotatedFigure}\vspace{0.5mm}
		\begin{annotatedFigure}
			{\adjincludegraphics[width=\linewidth, trim={0 {.0\width} {.65\width} {.65\width}},clip]	{fig/imgGE/imgGE_rd_glp_reg_fs_sc.jpg}};
		\end{annotatedFigure}\vspace{0.5mm}
		\begin{annotatedFigure}
			{\adjincludegraphics[width=\linewidth, trim={0 {.0\width} {.65\width} {.65\width}},clip]	{fig/imgGE/imgGE_rmse_glp_reg_fs.jpg}};
		\end{annotatedFigure}\vspace{0.5mm}
		\begin{annotatedFigure}
			{\adjincludegraphics[width=\linewidth, trim={0 {.0\width} {.65\width} {.65\width}},clip]	{fig/imgGE/imgGE_sam_glp_reg_fs.jpg}};
		\end{annotatedFigure}	
	\end{minipage}\label{fig:geoeye_rd:f}}\hspace{0.01mm}
	\subfloat[GSA-Segm]{\begin{minipage}[t]{0.095\linewidth}
		\begin{annotatedFigure}
			{\adjincludegraphics[width=\linewidth, trim={0 {.5\width} {.5\width} {.0\width}},clip]	{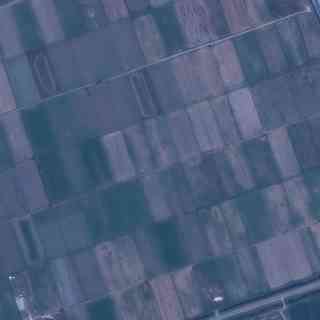}};
		\end{annotatedFigure}\vspace{0.5mm}
		\begin{annotatedFigure}
			{\adjincludegraphics[width=\linewidth, trim={0 {.5\width} {.5\width} {.0\width}},clip]	{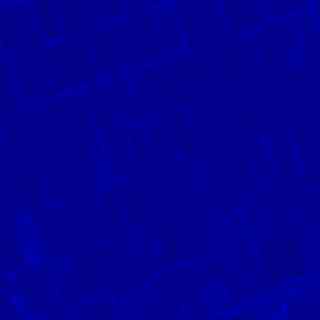}};
		\end{annotatedFigure}\vspace{0.5mm}
		\begin{annotatedFigure}
			{\adjincludegraphics[width=\linewidth, trim={0 {.5\width} {.5\width} {.0\width}},clip]	{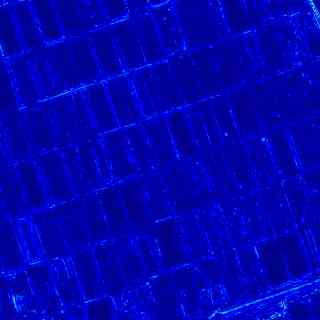}};
		\end{annotatedFigure}\vspace{0.5mm}
		\begin{annotatedFigure}
			{\adjincludegraphics[width=\linewidth, trim={0 {.0\width} {.65\width} {.65\width}},clip]	{fig/imgGE/imgGE_rd_seg_gsa_sc.jpg}};
		\end{annotatedFigure}\vspace{0.5mm}
		\begin{annotatedFigure}
			{\adjincludegraphics[width=\linewidth, trim={0 {.0\width} {.65\width} {.65\width}},clip]	{fig/imgGE/imgGE_rmse_seg_gsa.jpg}};
		\end{annotatedFigure}\vspace{0.5mm}
		\begin{annotatedFigure}
			{\adjincludegraphics[width=\linewidth, trim={0 {.0\width} {.65\width} {.65\width}},clip]	{fig/imgGE/imgGE_sam_seg_gsa.jpg}};
		\end{annotatedFigure}	
	\end{minipage}\label{fig:geoeye_rd:g}}\hspace{0.01mm}
	\subfloat[GLP-Segm]{\begin{minipage}[t]{0.095\linewidth}
		\begin{annotatedFigure}
			{\adjincludegraphics[width=\linewidth, trim={0 {.5\width} {.5\width} {.0\width}},clip]	{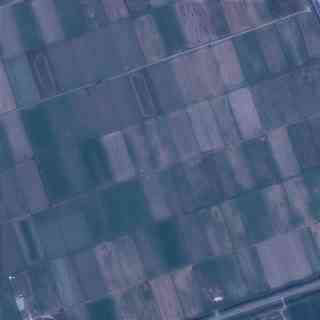}};
		\end{annotatedFigure}\vspace{0.5mm}
		\begin{annotatedFigure}
			{\adjincludegraphics[width=\linewidth, trim={0 {.5\width} {.5\width} {.0\width}},clip]	{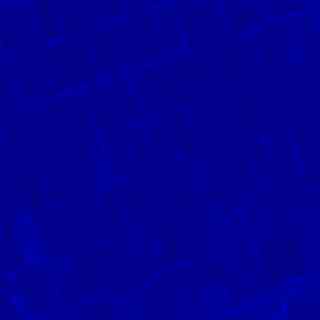}};
		\end{annotatedFigure}\vspace{0.5mm}
		\begin{annotatedFigure}
			{\adjincludegraphics[width=\linewidth, trim={0 {.5\width} {.5\width} {.0\width}},clip]	{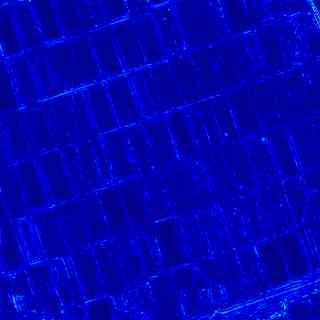}};
		\end{annotatedFigure}\vspace{0.5mm}
		\begin{annotatedFigure}
			{\adjincludegraphics[width=\linewidth, trim={0 {.0\width} {.65\width} {.65\width}},clip]	{fig/imgGE/imgGE_rd_seg_glp_sc.jpg}};
		\end{annotatedFigure}\vspace{0.5mm}
		\begin{annotatedFigure}
			{\adjincludegraphics[width=\linewidth, trim={0 {.0\width} {.65\width} {.65\width}},clip]	{fig/imgGE/imgGE_rmse_seg_glp.jpg}};
		\end{annotatedFigure}\vspace{0.5mm}
		\begin{annotatedFigure}
			{\adjincludegraphics[width=\linewidth, trim={0 {.0\width} {.65\width} {.65\width}},clip]	{fig/imgGE/imgGE_sam_seg_glp.jpg}};
		\end{annotatedFigure}	
	\end{minipage}\label{fig:geoeye_rd:h}}\hspace{0.01mm}
	\subfloat[\hspace{-0.5mm}Target-CNN]{\begin{minipage}[t]{0.095\linewidth}
		\begin{annotatedFigure}
			{\adjincludegraphics[width=\linewidth, trim={0 {.5\width} {.5\width} {.0\width}},clip]	{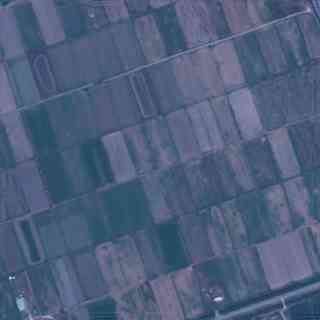}};
		\end{annotatedFigure}\vspace{0.5mm}
		\begin{annotatedFigure}
			{\adjincludegraphics[width=\linewidth, trim={0 {.5\width} {.5\width} {.0\width}},clip]	{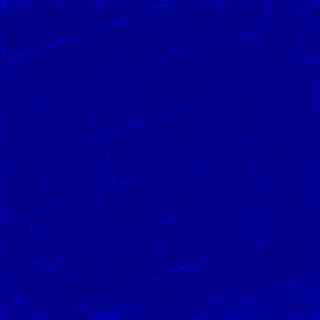}};
		\end{annotatedFigure}\vspace{0.5mm}
		\begin{annotatedFigure}
			{\adjincludegraphics[width=\linewidth, trim={0 {.5\width} {.5\width} {.0\width}},clip]	{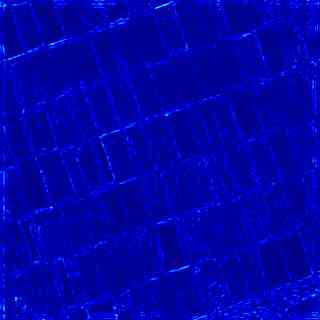}};
		\end{annotatedFigure}\vspace{0.5mm}
		\begin{annotatedFigure}
			{\adjincludegraphics[width=\linewidth, trim={0 {.0\width} {.65\width} {.65\width}},clip]	{fig/imgGE/imgGE_rd_pancnn_sc.jpg}};
		\end{annotatedFigure}\vspace{0.5mm}
		\begin{annotatedFigure}
			{\adjincludegraphics[width=\linewidth, trim={0 {.0\width} {.65\width} {.65\width}},clip]	{fig/imgGE/imgGE_rmse_pancnn.jpg}};
		\end{annotatedFigure}\vspace{0.5mm}
		\begin{annotatedFigure}
			{\adjincludegraphics[width=\linewidth, trim={0 {.0\width} {.65\width} {.65\width}},clip]	{fig/imgGE/imgGE_sam_pancnn.jpg}};
		\end{annotatedFigure}	
	\end{minipage}\label{fig:geoeye_rd:i}}\hspace{0.01mm}
	\subfloat[Ours]{\begin{minipage}[t]{0.095\linewidth}
		\begin{annotatedFigure}
			{\adjincludegraphics[width=\linewidth, trim={0 {.5\width} {.5\width} {.0\width}},clip]	{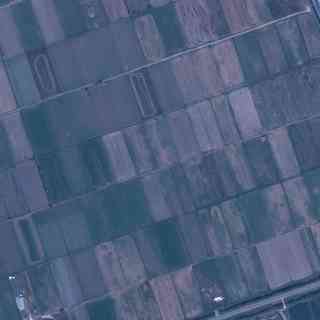}};
		\end{annotatedFigure}\vspace{0.5mm}
		\begin{annotatedFigure}
			{\adjincludegraphics[width=\linewidth, trim={0 {.5\width} {.5\width} {.0\width}},clip]	{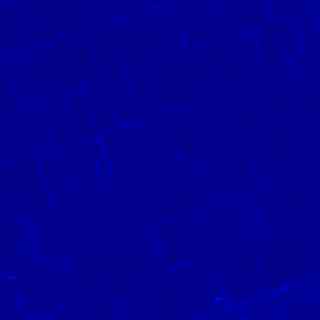}};
		\end{annotatedFigure}\vspace{0.5mm}
		\begin{annotatedFigure}
			{\adjincludegraphics[width=\linewidth, trim={0 {.5\width} {.5\width} {.0\width}},clip]	{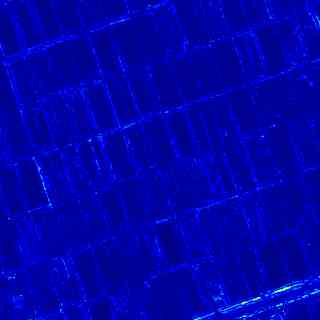}};
		\end{annotatedFigure}\vspace{0.5mm}
		\begin{annotatedFigure}
			{\adjincludegraphics[width=\linewidth, trim={0 {.0\width} {.65\width} {.65\width}},clip]	{fig/imgGE/imgGE_rd_ours_sc.jpg}};
		\end{annotatedFigure}\vspace{0.5mm}
		\begin{annotatedFigure}
			{\adjincludegraphics[width=\linewidth, trim={0 {.0\width} {.65\width} {.65\width}},clip]	{fig/imgGE/imgGE_rmse_ours.jpg}};
		\end{annotatedFigure}\vspace{0.5mm}
		\begin{annotatedFigure}
			{\adjincludegraphics[width=\linewidth, trim={0 {.0\width} {.65\width} {.65\width}},clip]	{fig/imgGE/imgGE_sam_ours.jpg}};
		\end{annotatedFigure}
	\end{minipage}\label{fig:geoeye_rd:j}}\hspace{0.01mm}
	\begin{minipage}{1\linewidth}
		\centering
		\includegraphics[width=0.5\linewidth]{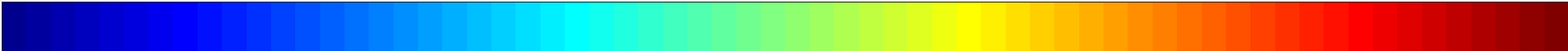}
	\end{minipage}
	\caption{Cropped patches of the pansharpened results from different methods on the GeoEye-1 dataset at reduced resolution. The results of the two patches are shown at the top three rows and the bottom three rows, respectively. For each patch, the top row shows the RGB channels of the pansharpened results. The middle row shows the average intensity difference maps between the reconstructed and the ground truth MSI. And the bottom row shows the average spectral difference maps between the reconstructed and the ground truth MSI. The first column shows the ground truth MSI (GT), LR MSI (M), and panchromatic (P) images for the two patches, respectively.}
	\label{fig:geoeye_rd}
\end{figure*}

\begin{table}[htbp]
	\caption{Performance comparison on the GeoEye-1 dataset at reduced resolution.}
	\label{tab:geoeye_rd}
	\centering
	\begin{tabular}{c| c c c c}
		\hline
		& PSNR & SAM & ERGAS & $Q2^n$ \\
		\hline
		GSA & 33.8281 & 2.2921 & 1.4189 & 0.9031 \\
		PRACS & 32.5368 & 2.3153 & 1.5839 & 0.8649\\
		BDSD-PC &33.168 & 2.4999 & 1.5091 & 0.9105\\
		MTF-CBD & 29.7486 & 2.9065 & 2.4275 & 0.8186\\
		GLP-Reg-FS& 33.4291 & 2.5428 & 1.5352& 0.89162\\
		GSA-Segm & 33.8058 & 2.565 & 1.5261 & 0.8909 \\
		GLP-Segm & 33.494& 2.304 & 1.4552 & 0.90365 \\
		Ours &\textbf{33.9028} & \textbf{2.2108}& \textbf{1.3908}& \textbf{0.9109}\\
		\hline
		Target-CNN & 33.8231 & 2.1566 & 1.3941 & 0.9153 \\
		HR MSI & $+\infty$ & 0 & 0 & 1 \\
		\hline  
	\end{tabular}%
\end{table}

\begin{figure*}[htbp]
	\subfloat[\hspace{-0.5mm}GT,$\mathbf{M}$,$\mathbf{P}$]{\begin{minipage}[t]{0.095\linewidth}
			\begin{annotatedFigure}
				{\adjincludegraphics[width=\linewidth,  trim={{.3\width} {.0\width} {.2\width} {.5\width}},clip]	{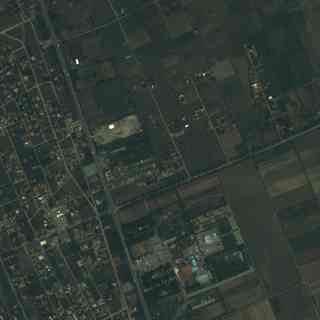}};
			\end{annotatedFigure}\vspace{0.5mm}
			\begin{annotatedFigure}
				{\adjincludegraphics[width=\linewidth,  trim={{.3\width} {.0\width} {.2\width} {.5\width}},clip]	{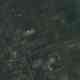}};
			\end{annotatedFigure}\vspace{0.5mm}
			\begin{annotatedFigure}
				{\adjincludegraphics[width=\linewidth,  trim={{.3\width} {.0\width} {.2\width} {.5\width}},clip]	{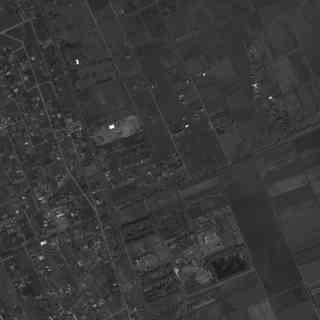}};
			\end{annotatedFigure}\vspace{0.5mm}
			\begin{annotatedFigure}
				{\adjincludegraphics[width=\linewidth,  trim={{.7\width} {.33\width} {.0\width} {.3\width}},clip]	{fig/imgIK/imgIK_rd_msi.jpg}};
			\end{annotatedFigure}\vspace{0.5mm}
			\begin{annotatedFigure}
				{\adjincludegraphics[width=\linewidth,  trim={{.7\width} {.33\width} {.0\width} {.3\width}},clip]	{fig/imgIK/imgIK_rd_msi_lr.jpg}};
			\end{annotatedFigure}\vspace{0.5mm}
			\begin{annotatedFigure}
				{\adjincludegraphics[width=\linewidth,  trim={{.7\width} {.33\width} {.0\width} {.3\width}},clip]	{fig/imgIK/imgIK_rd_pan.jpg}};
			\end{annotatedFigure}\vspace{1mm}
	\end{minipage}\label{fig:imgIK_rd:a}}\hspace{0.001mm}
	\subfloat[GSA]{\begin{minipage}[t]{0.095\linewidth}
			\begin{annotatedFigure}
				{\adjincludegraphics[width=\linewidth,  trim={{.3\width} {.0\width} {.2\width} {.5\width}},clip]	{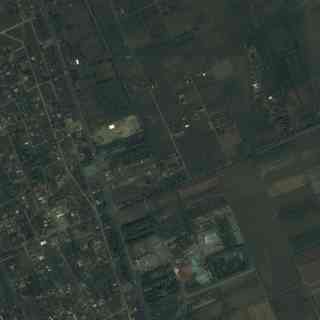}};
			\end{annotatedFigure}\vspace{0.5mm}
			\begin{annotatedFigure}
				{\adjincludegraphics[width=\linewidth,  trim={{.3\width} {.0\width} {.2\width} {.5\width}},clip]	{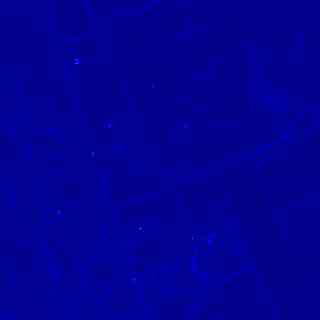}};
			\end{annotatedFigure}\vspace{0.5mm}
			\begin{annotatedFigure}
				{\adjincludegraphics[width=\linewidth,  trim={{.3\width} {.0\width} {.2\width} {.5\width}},clip]	{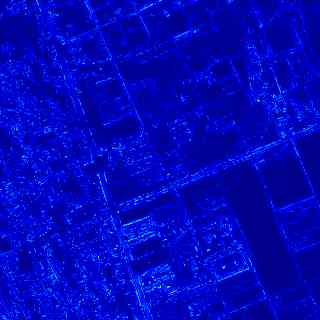}};
			\end{annotatedFigure}\vspace{0.5mm}
			\begin{annotatedFigure}
				{\adjincludegraphics[width=\linewidth,  trim={{.7\width} {.33\width} {.0\width} {.3\width}},clip]	{fig/imgIK/imgIK_rd_gsa.jpg}};
			\end{annotatedFigure}\vspace{0.5mm}
			\begin{annotatedFigure}
				{\adjincludegraphics[width=\linewidth,  trim={{.7\width} {.33\width} {.0\width} {.3\width}},clip]	{fig/imgIK/imgIK_rmse_gsa.jpg}};
			\end{annotatedFigure}\vspace{0.5mm}
			\begin{annotatedFigure}
				{\adjincludegraphics[width=\linewidth,  trim={{.7\width} {.33\width} {.0\width} {.3\width}},clip]	{fig/imgIK/imgIK_sam_gsa.jpg}};
			\end{annotatedFigure}\vspace{1mm}
	\end{minipage}\label{fig:imgIK_rd:b}}\hspace{0.001mm}
	\subfloat[PRACS]{\begin{minipage}[t]{0.095\linewidth}
		\begin{annotatedFigure}
			{\adjincludegraphics[width=\linewidth,  trim={{.3\width} {.0\width} {.2\width} {.5\width}},clip]	{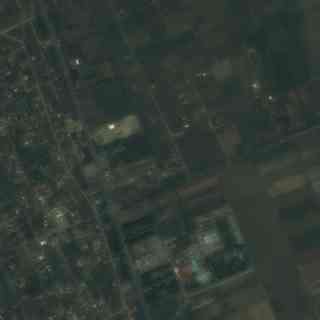}};
		\end{annotatedFigure}\vspace{0.5mm}
		\begin{annotatedFigure}
			{\adjincludegraphics[width=\linewidth,  trim={{.3\width} {.0\width} {.2\width} {.5\width}},clip]	{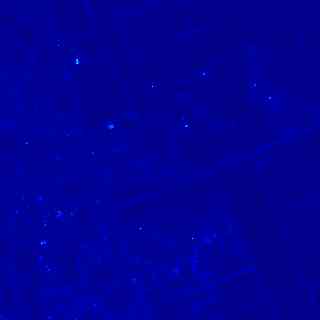}};
		\end{annotatedFigure}\vspace{0.5mm}
		\begin{annotatedFigure}
			{\adjincludegraphics[width=\linewidth,  trim={{.3\width} {.0\width} {.2\width} {.5\width}},clip]	{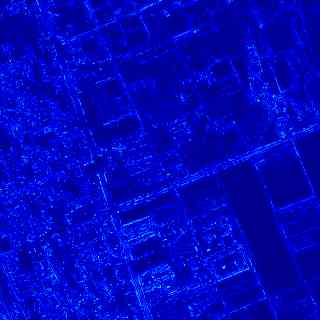}};
		\end{annotatedFigure}\vspace{0.5mm}
		\begin{annotatedFigure}
			{\adjincludegraphics[width=\linewidth,  trim={{.7\width} {.33\width} {.0\width} {.3\width}},clip]	{fig/imgIK/imgIK_rd_pracs.jpg}};
		\end{annotatedFigure}\vspace{0.5mm}
		\begin{annotatedFigure}
			{\adjincludegraphics[width=\linewidth,  trim={{.7\width} {.33\width} {.0\width} {.3\width}},clip]	{fig/imgIK/imgIK_rmse_pracs.jpg}};
		\end{annotatedFigure}\vspace{0.5mm}
		\begin{annotatedFigure}
			{\adjincludegraphics[width=\linewidth,  trim={{.7\width} {.33\width} {.0\width} {.3\width}},clip]	{fig/imgIK/imgIK_sam_pracs.jpg}};
		\end{annotatedFigure}\vspace{1mm}
	\end{minipage}\label{fig:imgIK_rd:c}}\hspace{0.001mm}
	\subfloat[BDSD-PC]{\begin{minipage}[t]{0.095\linewidth}
		\begin{annotatedFigure}
			{\adjincludegraphics[width=\linewidth,  trim={{.3\width} {.0\width} {.2\width} {.5\width}},clip]	{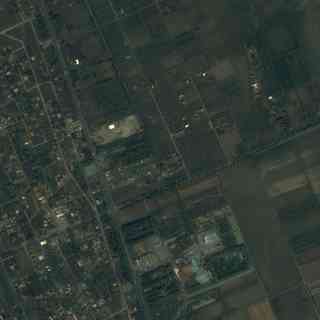}};
		\end{annotatedFigure}\vspace{0.5mm}
		\begin{annotatedFigure}
			{\adjincludegraphics[width=\linewidth,  trim={{.3\width} {.0\width} {.2\width} {.5\width}},clip]	{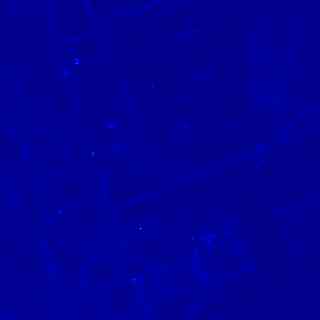}};
		\end{annotatedFigure}\vspace{0.5mm}
		\begin{annotatedFigure}
			{\adjincludegraphics[width=\linewidth,  trim={{.3\width} {.0\width} {.2\width} {.5\width}},clip]	{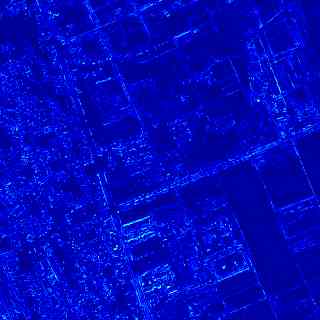}};
		\end{annotatedFigure}\vspace{0.5mm}
		\begin{annotatedFigure}
			{\adjincludegraphics[width=\linewidth,  trim={{.7\width} {.33\width} {.0\width} {.3\width}},clip]	{fig/imgIK/imgIK_rd_bdsd_pc.jpg}};
		\end{annotatedFigure}\vspace{0.5mm}
		\begin{annotatedFigure}
			{\adjincludegraphics[width=\linewidth,  trim={{.7\width} {.33\width} {.0\width} {.3\width}},clip]	{fig/imgIK/imgIK_rmse_bdsd_pc.jpg}};
		\end{annotatedFigure}\vspace{0.5mm}
		\begin{annotatedFigure}
			{\adjincludegraphics[width=\linewidth,  trim={{.7\width} {.33\width} {.0\width} {.3\width}},clip]	{fig/imgIK/imgIK_sam_bdsd_pc.jpg}};
		\end{annotatedFigure}\vspace{1mm}
	\end{minipage}\label{fig:imgIK_rd:d}}\hspace{0.001mm}
	\subfloat[MTF-CBD]{\begin{minipage}[t]{0.095\linewidth}
		\begin{annotatedFigure}
			{\adjincludegraphics[width=\linewidth,  trim={{.3\width} {.0\width} {.2\width} {.5\width}},clip]	{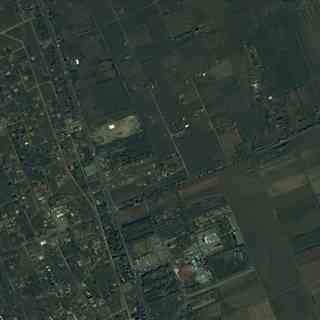}};
		\end{annotatedFigure}\vspace{0.5mm}
		\begin{annotatedFigure}
			{\adjincludegraphics[width=\linewidth,  trim={{.3\width} {.0\width} {.2\width} {.5\width}},clip]	{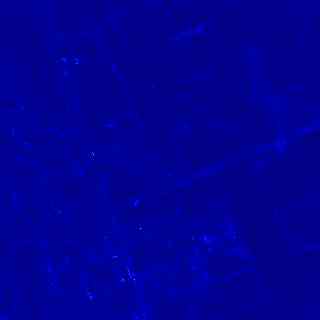}};
		\end{annotatedFigure}\vspace{0.5mm}
		\begin{annotatedFigure}
			{\adjincludegraphics[width=\linewidth,  trim={{.3\width} {.0\width} {.2\width} {.5\width}},clip]	{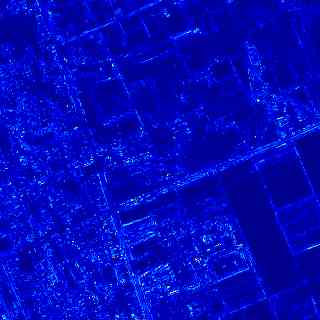}};
		\end{annotatedFigure}\vspace{0.5mm}
		\begin{annotatedFigure}
			{\adjincludegraphics[width=\linewidth,  trim={{.7\width} {.33\width} {.0\width} {.3\width}},clip]	{fig/imgIK/imgIK_rd_cbd.jpg}};
		\end{annotatedFigure}\vspace{0.5mm}
		\begin{annotatedFigure}
			{\adjincludegraphics[width=\linewidth,  trim={{.7\width} {.33\width} {.0\width} {.3\width}},clip]	{fig/imgIK/imgIK_rmse_cbd.jpg}};
		\end{annotatedFigure}\vspace{0.5mm}
		\begin{annotatedFigure}
			{\adjincludegraphics[width=\linewidth,  trim={{.7\width} {.33\width} {.0\width} {.3\width}},clip]	{fig/imgIK/imgIK_sam_cbd.jpg}};
		\end{annotatedFigure}\vspace{1mm}
	\end{minipage}\label{fig:imgIK_rd:e}}\hspace{0.001mm}
	\subfloat[\hspace{-1.1mm}GLP-Reg-FS]{\begin{minipage}[t]{0.095\linewidth}
		\begin{annotatedFigure}
			{\adjincludegraphics[width=\linewidth,  trim={{.3\width} {.0\width} {.2\width} {.5\width}},clip]	{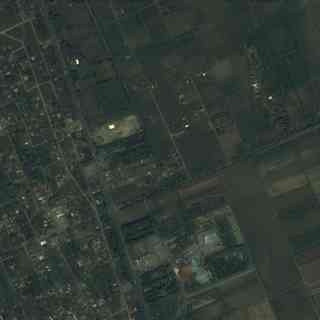}};
		\end{annotatedFigure}\vspace{0.5mm}
		\begin{annotatedFigure}
			{\adjincludegraphics[width=\linewidth,  trim={{.3\width} {.0\width} {.2\width} {.5\width}},clip]	{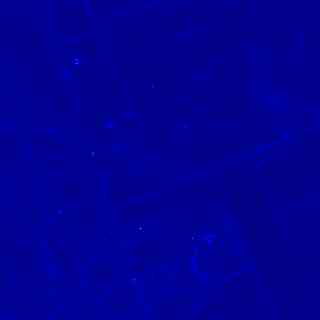}};
		\end{annotatedFigure}\vspace{0.5mm}
		\begin{annotatedFigure}
			{\adjincludegraphics[width=\linewidth,  trim={{.3\width} {.0\width} {.2\width} {.5\width}},clip]	{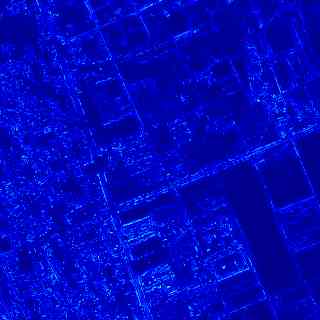}};
		\end{annotatedFigure}\vspace{0.5mm}
		\begin{annotatedFigure}
			{\adjincludegraphics[width=\linewidth,  trim={{.7\width} {.33\width} {.0\width} {.3\width}},clip]	{fig/imgIK/imgIK_rd_glp_reg_fs.jpg}};
		\end{annotatedFigure}\vspace{0.5mm}
		\begin{annotatedFigure}
			{\adjincludegraphics[width=\linewidth,  trim={{.7\width} {.33\width} {.0\width} {.3\width}},clip]	{fig/imgIK/imgIK_rmse_glp_reg_fs.jpg}};
		\end{annotatedFigure}\vspace{0.5mm}
		\begin{annotatedFigure}
			{\adjincludegraphics[width=\linewidth,  trim={{.7\width} {.33\width} {.0\width} {.3\width}},clip]	{fig/imgIK/imgIK_sam_glp_reg_fs.jpg}};
		\end{annotatedFigure}\vspace{1mm}
	\end{minipage}\label{fig:imgIK_rd:f}}\hspace{0.001mm}
	\subfloat[GSA-Segm]{\begin{minipage}[t]{0.095\linewidth}
		\begin{annotatedFigure}
			{\adjincludegraphics[width=\linewidth,  trim={{.3\width} {.0\width} {.2\width} {.5\width}},clip]	{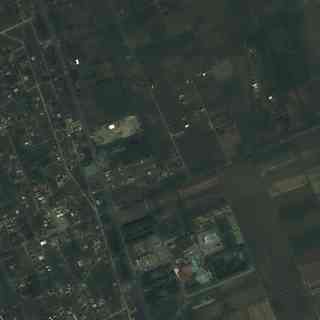}};
		\end{annotatedFigure}\vspace{0.5mm}
		\begin{annotatedFigure}
			{\adjincludegraphics[width=\linewidth,  trim={{.3\width} {.0\width} {.2\width} {.5\width}},clip]	{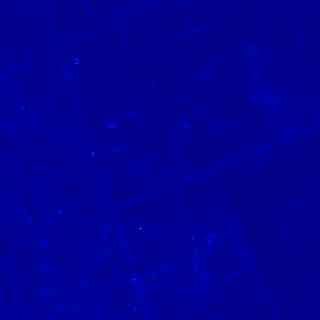}};
		\end{annotatedFigure}\vspace{0.5mm}
		\begin{annotatedFigure}
			{\adjincludegraphics[width=\linewidth,  trim={{.3\width} {.0\width} {.2\width} {.5\width}},clip]	{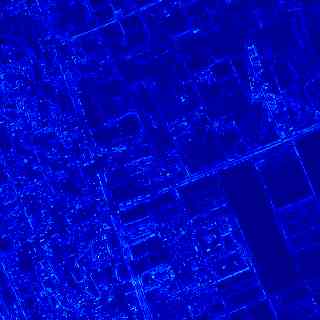}};
		\end{annotatedFigure}\vspace{0.5mm}
		\begin{annotatedFigure}
			{\adjincludegraphics[width=\linewidth,  trim={{.7\width} {.33\width} {.0\width} {.3\width}},clip]	{fig/imgIK/imgIK_rd_seg_gsa.jpg}};
		\end{annotatedFigure}\vspace{0.5mm}
		\begin{annotatedFigure}
			{\adjincludegraphics[width=\linewidth,  trim={{.7\width} {.33\width} {.0\width} {.3\width}},clip]	{fig/imgIK/imgIK_rmse_seg_gsa.jpg}};
		\end{annotatedFigure}\vspace{0.5mm}
		\begin{annotatedFigure}
			{\adjincludegraphics[width=\linewidth,  trim={{.7\width} {.33\width} {.0\width} {.3\width}},clip]	{fig/imgIK/imgIK_sam_seg_gsa.jpg}};
		\end{annotatedFigure}\vspace{1mm}
	\end{minipage}\label{fig:imgIK_rd:g}}\hspace{0.001mm}
	\subfloat[GLP-Segm]{\begin{minipage}[t]{0.095\linewidth}
		\begin{annotatedFigure}
			{\adjincludegraphics[width=\linewidth,  trim={{.3\width} {.0\width} {.2\width} {.5\width}},clip]	{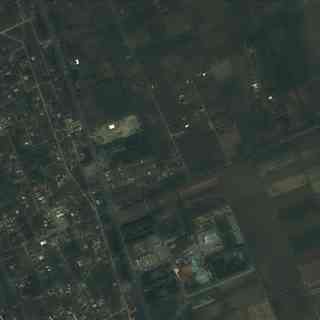}};
		\end{annotatedFigure}\vspace{0.5mm}
		\begin{annotatedFigure}
			{\adjincludegraphics[width=\linewidth,  trim={{.3\width} {.0\width} {.2\width} {.5\width}},clip]	{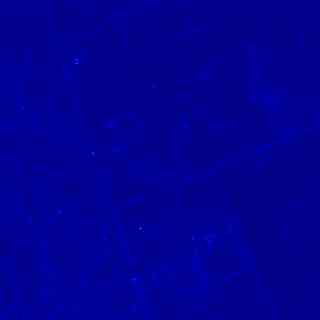}};
		\end{annotatedFigure}\vspace{0.5mm}
		\begin{annotatedFigure}
			{\adjincludegraphics[width=\linewidth,  trim={{.3\width} {.0\width} {.2\width} {.5\width}},clip]	{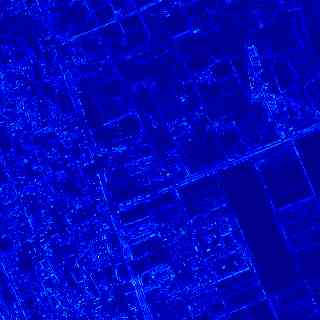}};
		\end{annotatedFigure}\vspace{0.5mm}
		\begin{annotatedFigure}
			{\adjincludegraphics[width=\linewidth,  trim={{.7\width} {.33\width} {.0\width} {.3\width}},clip]	{fig/imgIK/imgIK_rd_seg_glp.jpg}};
		\end{annotatedFigure}\vspace{0.5mm}
		\begin{annotatedFigure}
			{\adjincludegraphics[width=\linewidth,  trim={{.7\width} {.33\width} {.0\width} {.3\width}},clip]	{fig/imgIK/imgIK_rmse_seg_glp.jpg}};
		\end{annotatedFigure}\vspace{0.5mm}
		\begin{annotatedFigure}
			{\adjincludegraphics[width=\linewidth,  trim={{.7\width} {.33\width} {.0\width} {.3\width}},clip]	{fig/imgIK/imgIK_sam_seg_glp.jpg}};
		\end{annotatedFigure}\vspace{1mm}
	\end{minipage}\label{fig:imgIK_rd:h}}\hspace{0.001mm}
	\subfloat[\hspace{-0.5mm}Target-CNN]{\begin{minipage}[t]{0.095\linewidth}
		\begin{annotatedFigure}
			{\adjincludegraphics[width=\linewidth,  trim={{.3\width} {.0\width} {.2\width} {.5\width}},clip]	{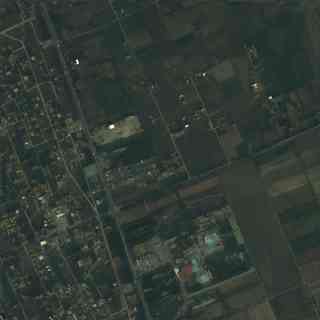}};
		\end{annotatedFigure}\vspace{0.5mm}
		\begin{annotatedFigure}
			{\adjincludegraphics[width=\linewidth,  trim={{.3\width} {.0\width} {.2\width} {.5\width}},clip]	{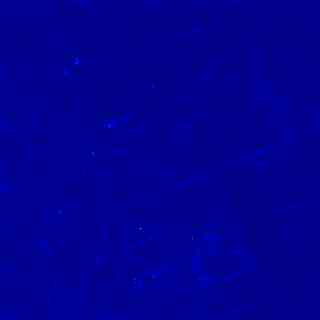}};
		\end{annotatedFigure}\vspace{0.5mm}
		\begin{annotatedFigure}
			{\adjincludegraphics[width=\linewidth,  trim={{.3\width} {.0\width} {.2\width} {.5\width}},clip]	{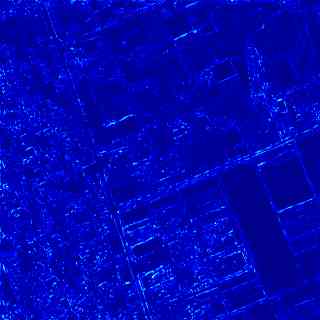}};
		\end{annotatedFigure}\vspace{0.5mm}
		\begin{annotatedFigure}
			{\adjincludegraphics[width=\linewidth,  trim={{.7\width} {.33\width} {.0\width} {.3\width}},clip]	{fig/imgIK/imgIK_rd_pancnn.jpg}};
		\end{annotatedFigure}\vspace{0.5mm}
		\begin{annotatedFigure}
			{\adjincludegraphics[width=\linewidth,  trim={{.7\width} {.33\width} {.0\width} {.3\width}},clip]	{fig/imgIK/imgIK_rmse_pancnn.jpg}};
		\end{annotatedFigure}\vspace{0.5mm}
		\begin{annotatedFigure}
			{\adjincludegraphics[width=\linewidth,  trim={{.7\width} {.33\width} {.0\width} {.3\width}},clip]	{fig/imgIK/imgIK_sam_pancnn.jpg}};
		\end{annotatedFigure}\vspace{1mm}
	\end{minipage}\label{fig:imgIK_rd:i}}\hspace{0.001mm}
	\subfloat[Ours]{\begin{minipage}[t]{0.095\linewidth}
		\begin{annotatedFigure}
			{\adjincludegraphics[width=\linewidth,  trim={{.3\width} {.0\width} {.2\width} {.5\width}},clip]	{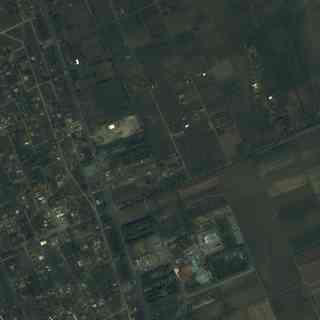}};
		\end{annotatedFigure}\vspace{0.5mm}
		\begin{annotatedFigure}
			{\adjincludegraphics[width=\linewidth,  trim={{.3\width} {.0\width} {.2\width} {.5\width}},clip]	{fig/imgIK/imgIK_rmse_pancnn.jpg}};
		\end{annotatedFigure}\vspace{0.5mm}
		\begin{annotatedFigure}
			{\adjincludegraphics[width=\linewidth,  trim={{.3\width} {.0\width} {.2\width} {.5\width}},clip]	{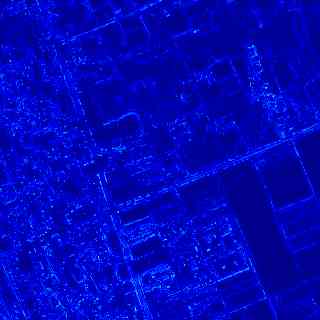}};
		\end{annotatedFigure}\vspace{0.5mm}
		\begin{annotatedFigure}
			{\adjincludegraphics[width=\linewidth,  trim={{.7\width} {.33\width} {.0\width} {.3\width}},clip]	{fig/imgIK/imgIK_rd_ours.jpg}};
		\end{annotatedFigure}\vspace{0.5mm}
		\begin{annotatedFigure}
			{\adjincludegraphics[width=\linewidth,  trim={{.7\width} {.33\width} {.0\width} {.3\width}},clip]	{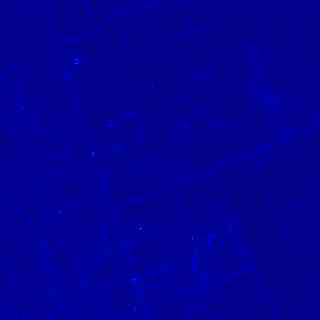}};
		\end{annotatedFigure}\vspace{0.5mm}
		\begin{annotatedFigure}
			{\adjincludegraphics[width=\linewidth,  trim={{.7\width} {.33\width} {.0\width} {.3\width}},clip]	{fig/imgIK/imgIK_sam_ours15.jpg}};
		\end{annotatedFigure}\vspace{1mm}
	\end{minipage}\label{fig:imgIK_rd:j}}\hspace{0.001mm}
	\begin{minipage}{1\linewidth}
		\centering
		\includegraphics[width=0.5\linewidth]{fig/colorbar.jpg}
	\end{minipage}
	\caption{Cropped patches of the pansharpened results from different methods on the IKONOS dataset at reduced resolution. The results of the two patches are shown at the top three rows and the bottom three rows, respectively. For each patch, the top row shows the RGB channels of the pansharpened results. The middle row shows the average intensity difference map between the reconstructed and the ground truth MSI. And the bottom row shows the average spectral difference map between the reconstructed and the ground truth MSI.The first column shows the ground truth MSI (GT), LR MSI (M), and panchromatic (P) images for the two patches, respectively.}
	\label{fig:imgIK_rd}
\end{figure*}

\begin{table}[htbp]
	\caption{Performance comparison on the IKONOS dataset at reduced resolution.}
	\label{tab:Ikonos_rd}
	\centering
	\begin{tabular}{c| c c c c}
		\hline
		& PSNR & SAM & ERGAS & $Q2^n$ \\
		\hline
		GSA & 34.1721 & 3.3972 & 2.2761 & 0.8583 \\
		PRACS & 32.4655 & 3.4449 & 2.7967 & 0.7932 \\
		BDSD-PC & 34.3254 & 3.4859 & 2.2274 & 0.8764 \\
		MTF-CBD & 32.6083 & 3.5253 & 2.7150 & 0.8409 \\
		GLP-Reg-FS & 34.1657 & 3.4006 & 2.2739 & 0.8633\\
		GSA-Segm & 34.2212 & 3.3154 & 2.2482 & 0.8701 \\
		GLP-Segm & 34.1316 & 3.324 & 2.2852 & 0.8645 \\
		Ours &\textbf{34.6342}&\textbf{3.2122}&\textbf{2.1559}& \textbf{0.8790}\\
		\hline
		Target-CNN & 35.0009 & 2.8289 & 2.0617 & 0.9037\\
		HR MSI & $+\infty$ & 0 & 0 & 1 \\
		\hline  
	\end{tabular}%
\end{table}

\begin{figure*}[htbp]
	\subfloat[\hspace{-0.5mm}GT,$\mathbf{M}$,$\mathbf{P}$]{\begin{minipage}[t]{0.095\linewidth}
			\begin{annotatedFigure}
				{\adjincludegraphics[width=\linewidth,   trim={{.0\width} {.0\width} {.5\width} {.5\width}},clip]	{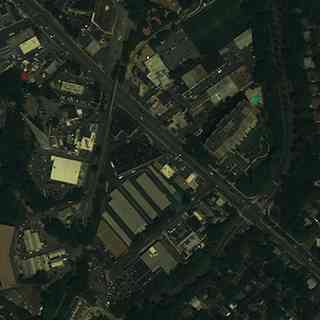}};
			\end{annotatedFigure}\vspace{0.5mm}
			\begin{annotatedFigure}
				{\adjincludegraphics[width=\linewidth,   trim={{.0\width} {.0\width} {.5\width} {.5\width}},clip]	{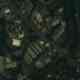}};
			\end{annotatedFigure}\vspace{0.5mm}
			\begin{annotatedFigure}
				{\adjincludegraphics[width=\linewidth,   trim={{.0\width} {.0\width} {.5\width} {.5\width}},clip]	{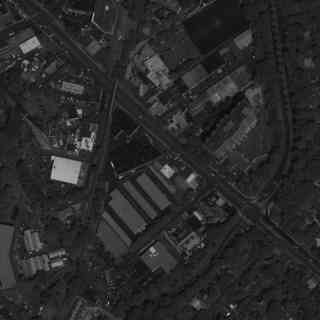}};
			\end{annotatedFigure}\vspace{0.5mm}
			\begin{annotatedFigure}
				{\adjincludegraphics[width=\linewidth,   trim={{.5\width} {.4\width} {.0\width} {.05\width}},clip]	{fig/imgWV2/imgWV2_rd_msi.jpg}};
			\end{annotatedFigure}\vspace{0.5mm}
			\begin{annotatedFigure}
				{\adjincludegraphics[width=\linewidth,   trim={{.5\width} {.4\width} {.0\width} {.05\width}},clip]	{fig/imgWV2/imgWV2_rd_msi_lr.jpg}};
			\end{annotatedFigure}\vspace{0.5mm}
			\begin{annotatedFigure}
				{\adjincludegraphics[width=\linewidth,   trim={{.5\width} {.4\width} {.0\width} {.05\width}},clip]	{fig/imgWV2/imgWV2_rd_pan.jpg}};
			\end{annotatedFigure}\vspace{1mm}
	\end{minipage}\label{fig:wv2_rd:a}}\hspace{0.001mm}
	\subfloat[GSA]{\begin{minipage}[t]{0.095\linewidth}
		\begin{annotatedFigure}
			{\adjincludegraphics[width=\linewidth,   trim={{.0\width} {.0\width} {.5\width} {.5\width}},clip]	{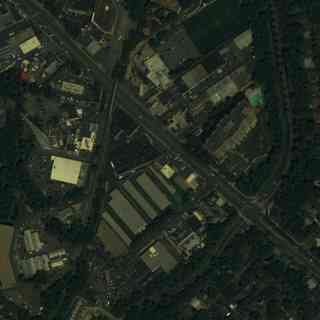}};
		\end{annotatedFigure}\vspace{0.5mm}
		\begin{annotatedFigure}
			{\adjincludegraphics[width=\linewidth,   trim={{.0\width} {.0\width} {.5\width} {.5\width}},clip]	{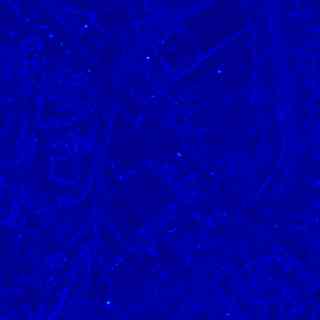}};
		\end{annotatedFigure}\vspace{0.5mm}
		\begin{annotatedFigure}
			{\adjincludegraphics[width=\linewidth,   trim={{.0\width} {.0\width} {.5\width} {.5\width}},clip]	{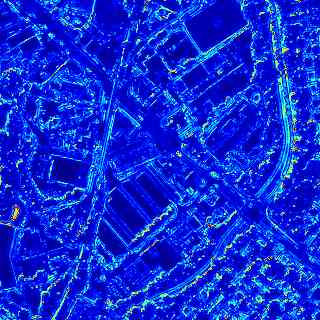}};
		\end{annotatedFigure}\vspace{0.5mm}
		\begin{annotatedFigure}
			{\adjincludegraphics[width=\linewidth,   trim={{.5\width} {.4\width} {.0\width} {.05\width}},clip]	{fig/imgWV2/imgWV2_rd_gsa.jpg}};
		\end{annotatedFigure}\vspace{0.5mm}
		\begin{annotatedFigure}
			{\adjincludegraphics[width=\linewidth,   trim={{.5\width} {.4\width} {.0\width} {.05\width}},clip]	{fig/imgWV2/imgWV2_rmse_gsa.jpg}};
		\end{annotatedFigure}\vspace{0.5mm}
		\begin{annotatedFigure}
			{\adjincludegraphics[width=\linewidth,   trim={{.5\width} {.4\width} {.0\width} {.05\width}},clip]	{fig/imgWV2/imgWV2_sam_gsa.jpg}};
		\end{annotatedFigure}\vspace{1mm}
	\end{minipage}\label{fig:wv2_rd:b}}\hspace{0.001mm}
	\subfloat[PRACS]{\begin{minipage}[t]{0.095\linewidth}
		\begin{annotatedFigure}
			{\adjincludegraphics[width=\linewidth,   trim={{.0\width} {.0\width} {.5\width} {.5\width}},clip]	{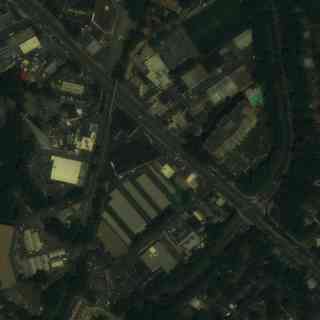}};
		\end{annotatedFigure}\vspace{0.5mm}
		\begin{annotatedFigure}
			{\adjincludegraphics[width=\linewidth,   trim={{.0\width} {.0\width} {.5\width} {.5\width}},clip]	{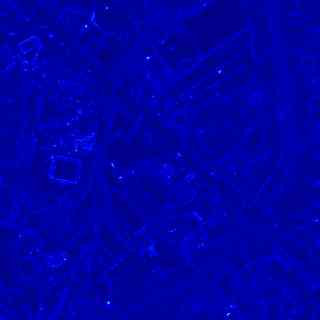}};
		\end{annotatedFigure}\vspace{0.5mm}
		\begin{annotatedFigure}
			{\adjincludegraphics[width=\linewidth,   trim={{.0\width} {.0\width} {.5\width} {.5\width}},clip]	{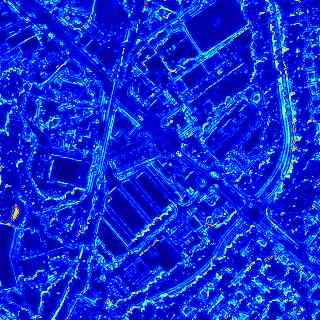}};
		\end{annotatedFigure}\vspace{0.5mm}
		\begin{annotatedFigure}
			{\adjincludegraphics[width=\linewidth,   trim={{.5\width} {.4\width} {.0\width} {.05\width}},clip]	{fig/imgWV2/imgWV2_rd_pracs.jpg}};
		\end{annotatedFigure}\vspace{0.5mm}
		\begin{annotatedFigure}
			{\adjincludegraphics[width=\linewidth,   trim={{.5\width} {.4\width} {.0\width} {.05\width}},clip]	{fig/imgWV2/imgWV2_rmse_pracs.jpg}};
		\end{annotatedFigure}\vspace{0.5mm}
		\begin{annotatedFigure}
			{\adjincludegraphics[width=\linewidth,   trim={{.5\width} {.4\width} {.0\width} {.05\width}},clip]	{fig/imgWV2/imgWV2_sam_pracs.jpg}};
		\end{annotatedFigure}\vspace{1mm}
	\end{minipage}\label{fig:wv2_rd:c}}\hspace{0.001mm}	
	\subfloat[BDSD-PC]{\begin{minipage}[t]{0.095\linewidth}
		\begin{annotatedFigure}
			{\adjincludegraphics[width=\linewidth,   trim={{.0\width} {.0\width} {.5\width} {.5\width}},clip]	{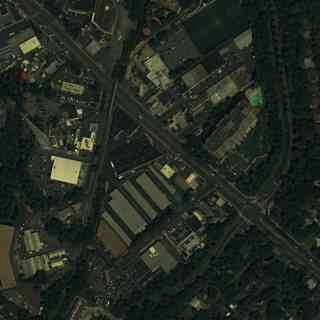}};
		\end{annotatedFigure}\vspace{0.5mm}
		\begin{annotatedFigure}
			{\adjincludegraphics[width=\linewidth,   trim={{.0\width} {.0\width} {.5\width} {.5\width}},clip]	{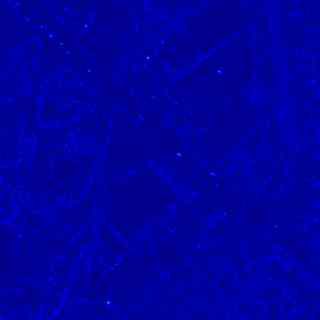}};
		\end{annotatedFigure}\vspace{0.5mm}
		\begin{annotatedFigure}
			{\adjincludegraphics[width=\linewidth,   trim={{.0\width} {.0\width} {.5\width} {.5\width}},clip]	{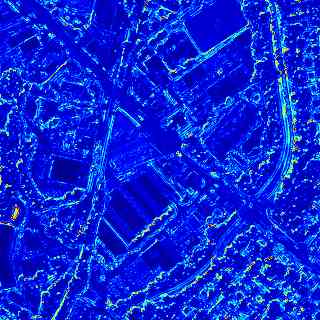}};
		\end{annotatedFigure}\vspace{0.5mm}
		\begin{annotatedFigure}
			{\adjincludegraphics[width=\linewidth,   trim={{.5\width} {.4\width} {.0\width} {.05\width}},clip]	{fig/imgWV2/imgWV2_rd_bdsd_pc.jpg}};
		\end{annotatedFigure}\vspace{0.5mm}
		\begin{annotatedFigure}
			{\adjincludegraphics[width=\linewidth,   trim={{.5\width} {.4\width} {.0\width} {.05\width}},clip]	{fig/imgWV2/imgWV2_rmse_bdsd_pc.jpg}};
		\end{annotatedFigure}\vspace{0.5mm}
		\begin{annotatedFigure}
			{\adjincludegraphics[width=\linewidth,   trim={{.5\width} {.4\width} {.0\width} {.05\width}},clip]	{fig/imgWV2/imgWV2_sam_bdsd_pc.jpg}};
		\end{annotatedFigure}\vspace{1mm}
	\end{minipage}\label{fig:wv2_rd:d}}\hspace{0.001mm}	
	\subfloat[MTF-CBD]{\begin{minipage}[t]{0.095\linewidth}
		\begin{annotatedFigure}
			{\adjincludegraphics[width=\linewidth,   trim={{.0\width} {.0\width} {.5\width} {.5\width}},clip]	{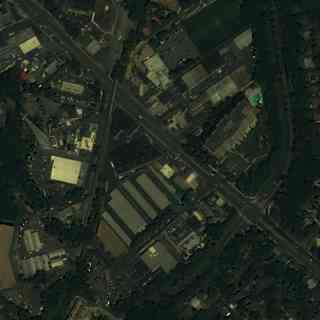}};
		\end{annotatedFigure}\vspace{0.5mm}
		\begin{annotatedFigure}
			{\adjincludegraphics[width=\linewidth,   trim={{.0\width} {.0\width} {.5\width} {.5\width}},clip]	{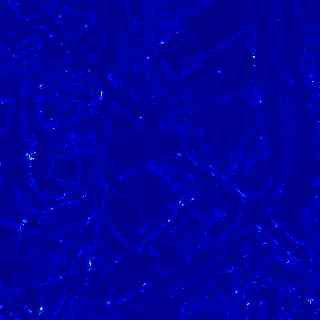}};
		\end{annotatedFigure}\vspace{0.5mm}
		\begin{annotatedFigure}
			{\adjincludegraphics[width=\linewidth,   trim={{.0\width} {.0\width} {.5\width} {.5\width}},clip]	{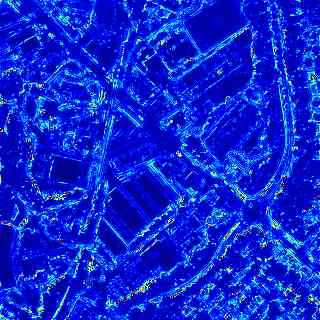}};
		\end{annotatedFigure}\vspace{0.5mm}
		\begin{annotatedFigure}
			{\adjincludegraphics[width=\linewidth,   trim={{.5\width} {.4\width} {.0\width} {.05\width}},clip]	{fig/imgWV2/imgWV2_rd_cbd.jpg}};
		\end{annotatedFigure}\vspace{0.5mm}
		\begin{annotatedFigure}
			{\adjincludegraphics[width=\linewidth,   trim={{.5\width} {.4\width} {.0\width} {.05\width}},clip]	{fig/imgWV2/imgWV2_rmse_cbd.jpg}};
		\end{annotatedFigure}\vspace{0.5mm}
		\begin{annotatedFigure}
			{\adjincludegraphics[width=\linewidth,   trim={{.5\width} {.4\width} {.0\width} {.05\width}},clip]	{fig/imgWV2/imgWV2_sam_cbd.jpg}};
		\end{annotatedFigure}\vspace{1mm}
	\end{minipage}\label{fig:wv2_rd:e}}\hspace{0.001mm}	
	\subfloat[\hspace{-1.1mm}GLP-Reg-FS]{\begin{minipage}[t]{0.095\linewidth}
		\begin{annotatedFigure}
			{\adjincludegraphics[width=\linewidth,   trim={{.0\width} {.0\width} {.5\width} {.5\width}},clip]	{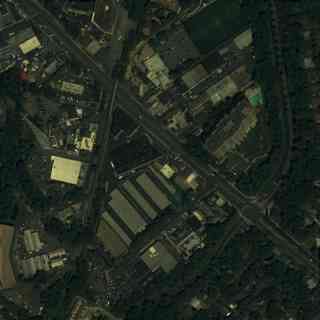}};
		\end{annotatedFigure}\vspace{0.5mm}
		\begin{annotatedFigure}
			{\adjincludegraphics[width=\linewidth,   trim={{.0\width} {.0\width} {.5\width} {.5\width}},clip]	{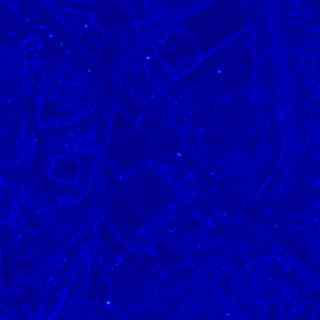}};
		\end{annotatedFigure}\vspace{0.5mm}
		\begin{annotatedFigure}
			{\adjincludegraphics[width=\linewidth,   trim={{.0\width} {.0\width} {.5\width} {.5\width}},clip]	{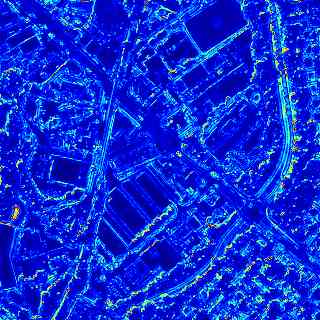}};
		\end{annotatedFigure}\vspace{0.5mm}
		\begin{annotatedFigure}
			{\adjincludegraphics[width=\linewidth,   trim={{.5\width} {.4\width} {.0\width} {.05\width}},clip]	{fig/imgWV2/imgWV2_rd_glp_reg_fs.jpg}};
		\end{annotatedFigure}\vspace{0.5mm}
		\begin{annotatedFigure}
			{\adjincludegraphics[width=\linewidth,   trim={{.5\width} {.4\width} {.0\width} {.05\width}},clip]	{fig/imgWV2/imgWV2_rmse_glp_reg_fs.jpg}};
		\end{annotatedFigure}\vspace{0.5mm}
		\begin{annotatedFigure}
			{\adjincludegraphics[width=\linewidth,   trim={{.5\width} {.4\width} {.0\width} {.05\width}},clip]	{fig/imgWV2/imgWV2_sam_glp_reg_fs.jpg}};
		\end{annotatedFigure}\vspace{1mm}
	\end{minipage}\label{fig:wv2_rd:f}}\hspace{0.001mm}	
	\subfloat[GSA-Segm]{\begin{minipage}[t]{0.095\linewidth}
		\begin{annotatedFigure}
			{\adjincludegraphics[width=\linewidth,   trim={{.0\width} {.0\width} {.5\width} {.5\width}},clip]	{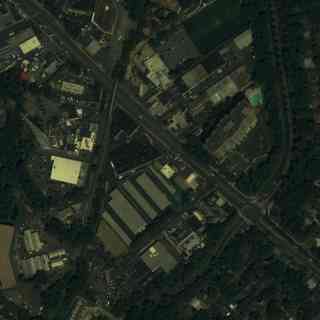}};
		\end{annotatedFigure}\vspace{0.5mm}
		\begin{annotatedFigure}
			{\adjincludegraphics[width=\linewidth,   trim={{.0\width} {.0\width} {.5\width} {.5\width}},clip]	{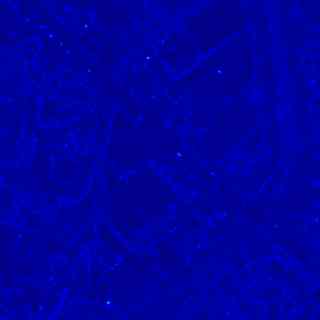}};
		\end{annotatedFigure}\vspace{0.5mm}
		\begin{annotatedFigure}
			{\adjincludegraphics[width=\linewidth,   trim={{.0\width} {.0\width} {.5\width} {.5\width}},clip]	{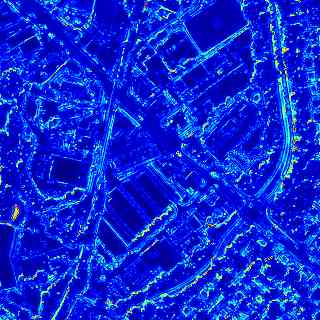}};
		\end{annotatedFigure}\vspace{0.5mm}
		\begin{annotatedFigure}
			{\adjincludegraphics[width=\linewidth,   trim={{.5\width} {.4\width} {.0\width} {.05\width}},clip]	{fig/imgWV2/imgWV2_rd_seg_gsa.jpg}};
		\end{annotatedFigure}\vspace{0.5mm}
		\begin{annotatedFigure}
			{\adjincludegraphics[width=\linewidth,   trim={{.5\width} {.4\width} {.0\width} {.05\width}},clip]	{fig/imgWV2/imgWV2_rmse_seg_gsa.jpg}};
		\end{annotatedFigure}\vspace{0.5mm}
		\begin{annotatedFigure}
			{\adjincludegraphics[width=\linewidth,   trim={{.5\width} {.4\width} {.0\width} {.05\width}},clip]	{fig/imgWV2/imgWV2_sam_seg_gsa.jpg}};
		\end{annotatedFigure}\vspace{1mm}
	\end{minipage}\label{fig:wv2_rd:g}}\hspace{0.001mm}	
	\subfloat[GLP-Segm]{\begin{minipage}[t]{0.095\linewidth}
		\begin{annotatedFigure} 
			{\adjincludegraphics[width=\linewidth,   trim={{.0\width} {.0\width} {.5\width} {.5\width}},clip]	{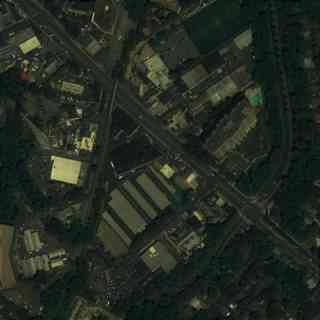}};
		\end{annotatedFigure}\vspace{0.5mm}
		\begin{annotatedFigure}
			{\adjincludegraphics[width=\linewidth,   trim={{.0\width} {.0\width} {.5\width} {.5\width}},clip]	{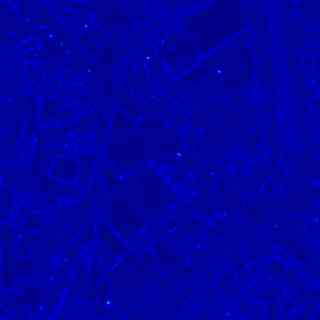}};
		\end{annotatedFigure}\vspace{0.5mm}
		\begin{annotatedFigure}
			{\adjincludegraphics[width=\linewidth,   trim={{.0\width} {.0\width} {.5\width} {.5\width}},clip]	{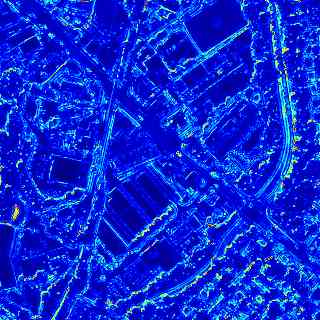}};
		\end{annotatedFigure}\vspace{0.5mm}
		\begin{annotatedFigure}
			{\adjincludegraphics[width=\linewidth,   trim={{.5\width} {.4\width} {.0\width} {.05\width}},clip]	{fig/imgWV2/imgWV2_rd_seg_glp.jpg}};
		\end{annotatedFigure}\vspace{0.5mm}
		\begin{annotatedFigure}
			{\adjincludegraphics[width=\linewidth,   trim={{.5\width} {.4\width} {.0\width} {.05\width}},clip]	{fig/imgWV2/imgWV2_rmse_seg_glp.jpg}};
		\end{annotatedFigure}\vspace{0.5mm}
		\begin{annotatedFigure}
			{\adjincludegraphics[width=\linewidth,   trim={{.5\width} {.4\width} {.0\width} {.05\width}},clip]	{fig/imgWV2/imgWV2_sam_seg_glp.jpg}};
		\end{annotatedFigure}\vspace{1mm}
	\end{minipage}\label{fig:wv2_rd:h}}\hspace{0.001mm}	
	\subfloat[\hspace{-0.5mm}Target-CNN]{\begin{minipage}[t]{0.095\linewidth}
		\begin{annotatedFigure} 
			{\adjincludegraphics[width=\linewidth,   trim={{.0\width} {.0\width} {.5\width} {.5\width}},clip]	{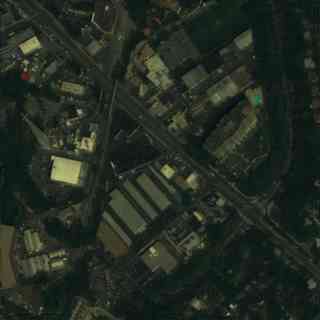}};
		\end{annotatedFigure}\vspace{0.5mm}
		\begin{annotatedFigure}
			{\adjincludegraphics[width=\linewidth,   trim={{.0\width} {.0\width} {.5\width} {.5\width}},clip]	{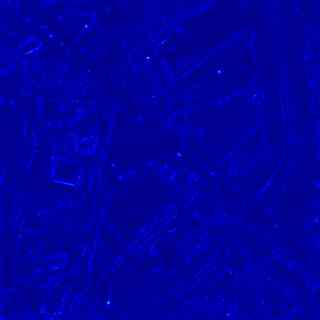}};
		\end{annotatedFigure}\vspace{0.5mm}
		\begin{annotatedFigure}
			{\adjincludegraphics[width=\linewidth,   trim={{.0\width} {.0\width} {.5\width} {.5\width}},clip]	{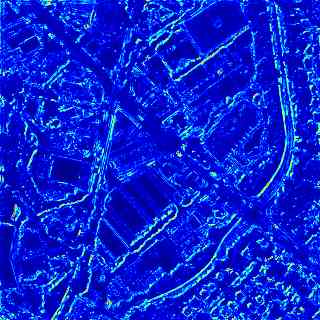}};
		\end{annotatedFigure}\vspace{0.5mm}
		\begin{annotatedFigure}
			{\adjincludegraphics[width=\linewidth,   trim={{.5\width} {.4\width} {.0\width} {.05\width}},clip]	{fig/imgWV2/imgWV2_rd_pancnn.jpg}};
		\end{annotatedFigure}\vspace{0.5mm}
		\begin{annotatedFigure}
			{\adjincludegraphics[width=\linewidth,   trim={{.5\width} {.4\width} {.0\width} {.05\width}},clip]	{fig/imgWV2/imgWV2_rmse_pancnn.jpg}};
		\end{annotatedFigure}\vspace{0.5mm}
		\begin{annotatedFigure}
			{\adjincludegraphics[width=\linewidth,   trim={{.5\width} {.4\width} {.0\width} {.05\width}},clip]	{fig/imgWV2/imgWV2_sam_pancnn.jpg}};
		\end{annotatedFigure}\vspace{1mm}
	\end{minipage}\label{fig:wv2_rd:i}}\hspace{0.001mm}	
	\subfloat[Ours]{\begin{minipage}[t]{0.095\linewidth}
		\begin{annotatedFigure} 
			{\adjincludegraphics[width=\linewidth,   trim={{.0\width} {.0\width} {.5\width} {.5\width}},clip]	{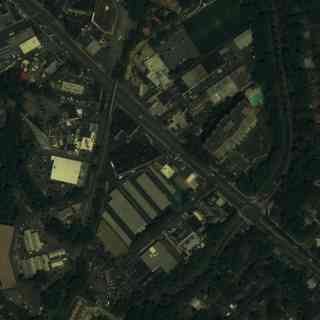}};
		\end{annotatedFigure}\vspace{0.5mm}
		\begin{annotatedFigure}
			{\adjincludegraphics[width=\linewidth,   trim={{.0\width} {.0\width} {.5\width} {.5\width}},clip]	{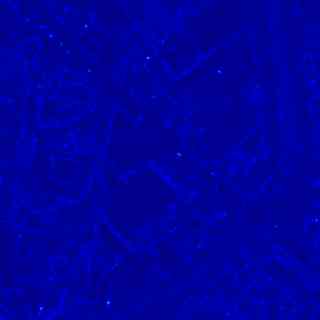}};
		\end{annotatedFigure}\vspace{0.5mm}
		\begin{annotatedFigure}
			{\adjincludegraphics[width=\linewidth,   trim={{.0\width} {.0\width} {.5\width} {.5\width}},clip]	{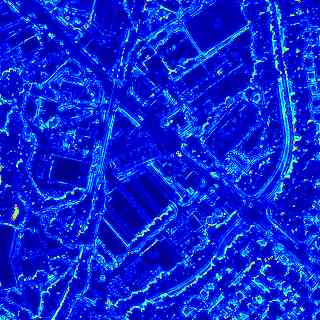}};
		\end{annotatedFigure}\vspace{0.5mm}
		\begin{annotatedFigure}
			{\adjincludegraphics[width=\linewidth,   trim={{.5\width} {.4\width} {.0\width} {.05\width}},clip]	{fig/imgWV2/imgWV2_rd_ours.jpg}};
		\end{annotatedFigure}\vspace{0.5mm}
		\begin{annotatedFigure}
			{\adjincludegraphics[width=\linewidth,   trim={{.5\width} {.4\width} {.0\width} {.05\width}},clip]	{fig/imgWV2/imgWV2_rmse_ours.jpg}};
		\end{annotatedFigure}\vspace{0.5mm}
		\begin{annotatedFigure}
			{\adjincludegraphics[width=\linewidth,   trim={{.5\width} {.4\width} {.0\width} {.05\width}},clip]	{fig/imgWV2/imgWV2_sam_ours.jpg}};
		\end{annotatedFigure}\vspace{1mm}
	\end{minipage}\label{fig:wv2_rd:j}}\hspace{0.001mm}	
	\begin{minipage}{1\linewidth}
		\centering
		\includegraphics[width=0.5\linewidth]{fig/colorbar.jpg}
	\end{minipage}
	\caption{Cropped patches of the pansharpened results from different methods on the WorldView2 dataset at reduced resolution. The results of the two patches are shown at the top three rows and the bottom three rows, respectively. For each patch, the top row shows the RGB channels of the pansharpened results. The middle row shows the average intensity difference map between the reconstructed and the ground truth MSI. And the bottom row shows the average spectral difference map between the reconstructed and the ground truth MSI. The first column shows the ground truth MSI (GT), LR MSI (M), and panchromatic (P) images for the two patches, respectively.}
	\label{fig:wv2_rd}
\end{figure*}

\begin{table}[htbp]
	\caption{Performance comparison on the WorldView-2 dataset at reduced resolution.}
	\label{tab:wv2_rd}
	\centering
	\begin{tabular}{c| c c c c}
		\hline
		& PSNR & SAM & ERGAS & $Q2^n$ \\
		\hline
		GSA & 29.4057 & 8.1089 & 4.9977 & 0.8678 \\
		PRACS & 27.8108 & 7.9880 & 5.8213 & 0.8143 \\
		BDSD-PC & 30.0782 & 7.6924& 4.5188 & \textbf{0.8949}\\
		MTF-CBD & 28.6271 & 7.9131 & 5.4890 & 0.8658 \\
		GLP-Reg-FS & 29.3599 & 8.2072 & 4.9851 & 0.8749\\
		GSA-Segm & 29.8942 & 7.7044 & {4.4614} & 0.8831\\
		GLP-Segm & 29.434 & 7.69 & {4.8364} & 0.87547\\
		Ours & \textbf{30.091}& \textbf{7.27}&\textbf{4.4527} & {0.8909}\\
		\hline
		Target-CNN & 29.0564 & 6.9455 & 4.8480 & 0.8717\\
		HR MSI & $+\infty$ & 0 & 0 & 1 \\
		\hline  
	\end{tabular}%
\end{table}

\begin{figure*}[htbp]
	\subfloat[\hspace{-0.5mm}GT,$\mathbf{M}$,$\mathbf{P}$]{\begin{minipage}[t]{0.095\linewidth}
			\begin{annotatedFigure}
				{\adjincludegraphics[width=\linewidth,   trim={{.1\width} {.0\width} {.55\width} {.7\width}},clip]	{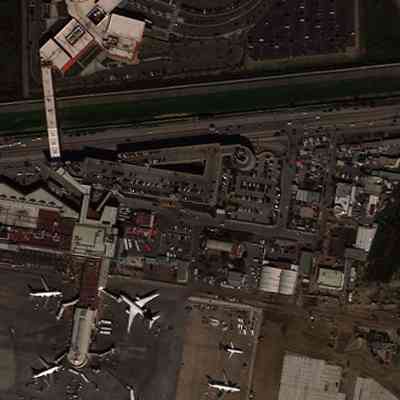}};
			\end{annotatedFigure}\vspace{0.5mm}
			\begin{annotatedFigure}
				{\adjincludegraphics[width=\linewidth,   trim={{.1\width} {.0\width} {.55\width} {.7\width}},clip]	{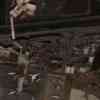}};
			\end{annotatedFigure}\vspace{0.5mm}
			\begin{annotatedFigure}
				{\adjincludegraphics[width=\linewidth,   trim={{.1\width} {.0\width} {.55\width} {.7\width}},clip]	{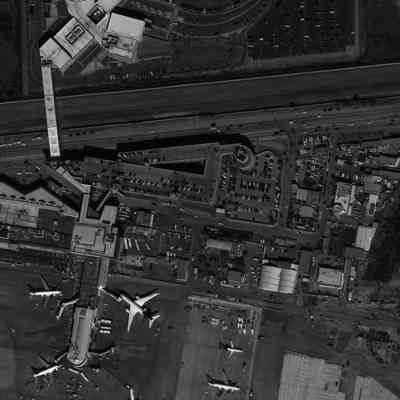}};
			\end{annotatedFigure}\vspace{0.5mm}
			\begin{annotatedFigure}
				{\adjincludegraphics[width=\linewidth,   trim={{.1\width} {.5\width} {.4\width} {.0\width}},clip]	{fig/imgWV3/imgWV3_rd_msi.jpg}};
			\end{annotatedFigure}\vspace{0.5mm}
			\begin{annotatedFigure}
				{\adjincludegraphics[width=\linewidth,   trim={{.1\width} {.5\width} {.4\width} {.0\width}},clip]	{fig/imgWV3/imgWV3_rd_msi_lr.jpg}};
			\end{annotatedFigure}\vspace{0.5mm}
			\begin{annotatedFigure}
				{\adjincludegraphics[width=\linewidth,   trim={{.1\width} {.5\width} {.4\width} {.0\width}},clip]	{fig/imgWV3/imgWV3_rd_pan.jpg}};
			\end{annotatedFigure}\vspace{1mm}
	\end{minipage}\label{fig:wv3_rd:a}}\hspace{0.001mm}
	\subfloat[GSA]{\begin{minipage}[t]{0.095\linewidth}
		\begin{annotatedFigure}
			{\adjincludegraphics[width=\linewidth,  trim={{.1\width} {.0\width} {.55\width} {.7\width}},clip]	{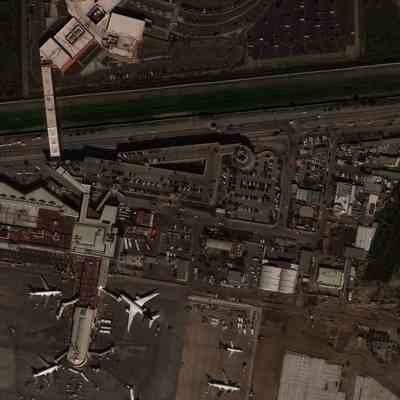}};
		\end{annotatedFigure}\vspace{0.5mm}
		\begin{annotatedFigure}
			{\adjincludegraphics[width=\linewidth,   trim={{.1\width} {.0\width} {.55\width} {.7\width}},clip]	{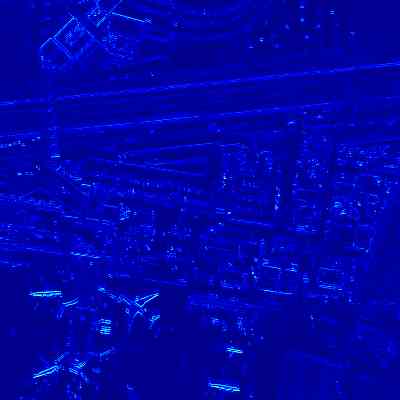}};
		\end{annotatedFigure}\vspace{0.5mm}
		\begin{annotatedFigure}
			{\adjincludegraphics[width=\linewidth,   trim={{.1\width} {.0\width} {.55\width} {.7\width}},clip]	{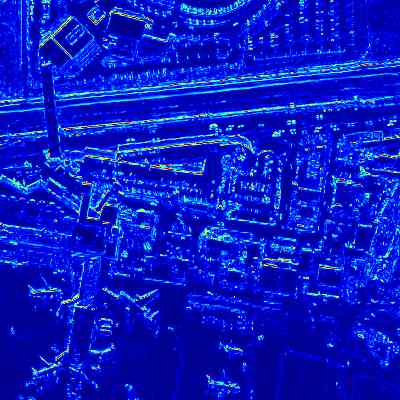}};
		\end{annotatedFigure}\vspace{0.5mm}
		\begin{annotatedFigure}
			{\adjincludegraphics[width=\linewidth,  trim={{.1\width} {.5\width} {.4\width} {.0\width}},clip]	{fig/imgWV3/imgWV3_rd_gsa.jpg}};
		\end{annotatedFigure}\vspace{0.5mm}
		\begin{annotatedFigure}
			{\adjincludegraphics[width=\linewidth,  trim={{.1\width} {.5\width} {.4\width} {.0\width}},clip]	{fig/imgWV3/imgWV3_rmse_gsa.jpg}};
		\end{annotatedFigure}\vspace{0.5mm}
		\begin{annotatedFigure}
			{\adjincludegraphics[width=\linewidth,  trim={{.1\width} {.5\width} {.4\width} {.0\width}},clip]	{fig/imgWV3/imgWV3_sam_gsa.jpg}};
		\end{annotatedFigure}\vspace{1mm}
	\end{minipage}\label{fig:wv3_rd:b}}\hspace{0.001mm}
	\subfloat[PRACS]{\begin{minipage}[t]{0.095\linewidth}
		\begin{annotatedFigure}
			{\adjincludegraphics[width=\linewidth,   trim={{.1\width} {.0\width} {.55\width} {.7\width}},clip]	{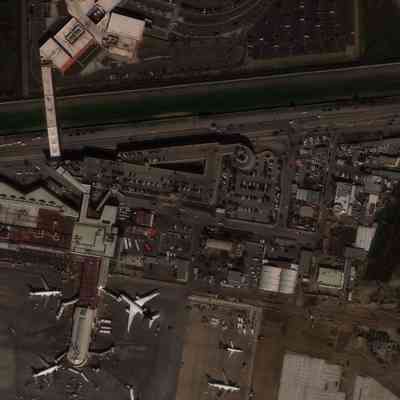}};
		\end{annotatedFigure}\vspace{0.5mm}
		\begin{annotatedFigure}
			{\adjincludegraphics[width=\linewidth,   trim={{.1\width} {.0\width} {.55\width} {.7\width}},clip]	{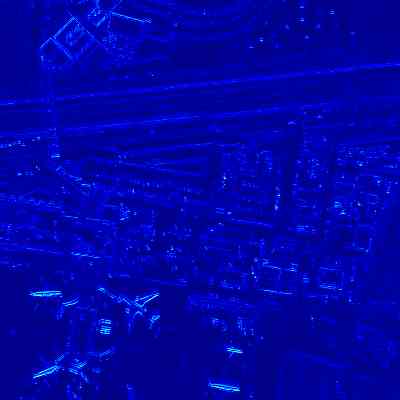}};
		\end{annotatedFigure}\vspace{0.5mm}
		\begin{annotatedFigure}
			{\adjincludegraphics[width=\linewidth,   trim={{.1\width} {.0\width} {.55\width} {.7\width}},clip]	{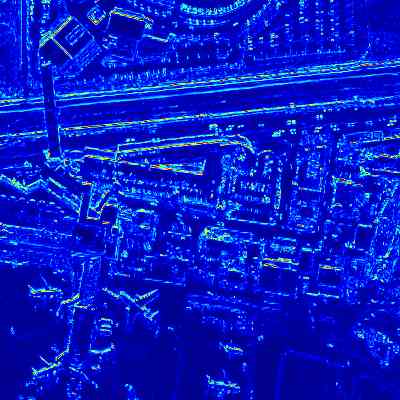}};
		\end{annotatedFigure}\vspace{0.5mm}
		\begin{annotatedFigure}
			{\adjincludegraphics[width=\linewidth,   trim={{.1\width} {.5\width} {.4\width} {.0\width}},clip]	{fig/imgWV3/imgWV3_rd_pracs.jpg}};
		\end{annotatedFigure}\vspace{0.5mm}
		\begin{annotatedFigure}
			{\adjincludegraphics[width=\linewidth,   trim={{.1\width} {.5\width} {.4\width} {.0\width}},clip]	{fig/imgWV3/imgWV3_rmse_pracs.jpg}};
		\end{annotatedFigure}\vspace{0.5mm}
		\begin{annotatedFigure}
			{\adjincludegraphics[width=\linewidth,  trim={{.1\width} {.5\width} {.4\width} {.0\width}},clip]	{fig/imgWV3/imgWV3_sam_pracs.jpg}};
		\end{annotatedFigure}\vspace{1mm}
	\end{minipage}\label{fig:wv3_rd:c}}\hspace{0.001mm}
	\subfloat[BDSD-PC]{\begin{minipage}[t]{0.095\linewidth}
		\begin{annotatedFigure}
			{\adjincludegraphics[width=\linewidth,   trim={{.1\width} {.0\width} {.55\width} {.7\width}},clip]	{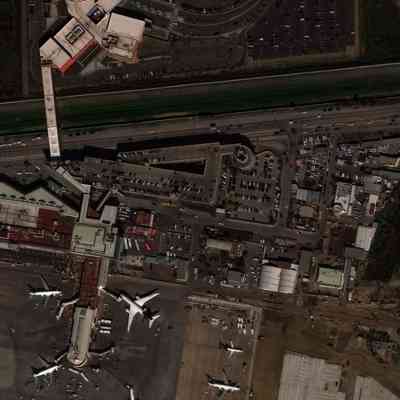}};
		\end{annotatedFigure}\vspace{0.5mm}
		\begin{annotatedFigure}
			{\adjincludegraphics[width=\linewidth,   trim={{.1\width} {.0\width} {.55\width} {.7\width}},clip]	{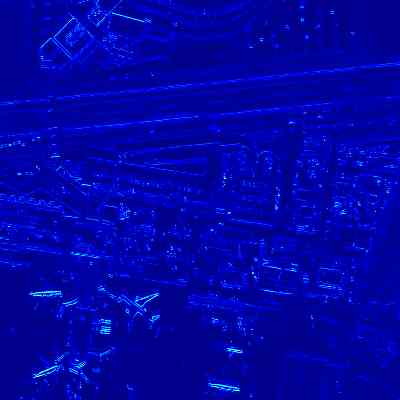}};
		\end{annotatedFigure}\vspace{0.5mm}
		\begin{annotatedFigure}
			{\adjincludegraphics[width=\linewidth,   trim={{.1\width} {.0\width} {.55\width} {.7\width}},clip]	{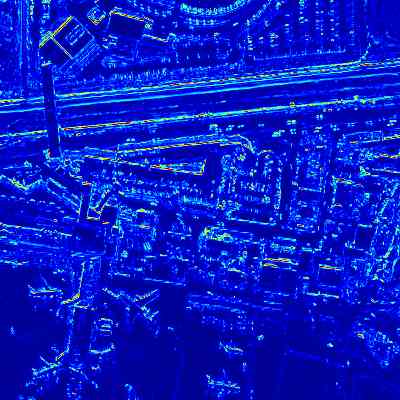}};
		\end{annotatedFigure}\vspace{0.5mm}
		\begin{annotatedFigure}
			{\adjincludegraphics[width=\linewidth,   trim={{.1\width} {.5\width} {.4\width} {.0\width}},clip]	{fig/imgWV3/imgWV3_rd_bdsd_pc.jpg}};
		\end{annotatedFigure}\vspace{0.5mm}
		\begin{annotatedFigure}
			{\adjincludegraphics[width=\linewidth,   trim={{.1\width} {.5\width} {.4\width} {.0\width}},clip]	{fig/imgWV3/imgWV3_rmse_bdsd_pc.jpg}};
		\end{annotatedFigure}\vspace{0.5mm}
		\begin{annotatedFigure}
			{\adjincludegraphics[width=\linewidth,   trim={{.1\width} {.5\width} {.4\width} {.0\width}},clip]	{fig/imgWV3/imgWV3_sam_bdsd_pc.jpg}};
		\end{annotatedFigure}\vspace{1mm}
	\end{minipage}\label{fig:wv3_rd:d}}\hspace{0.001mm}
	\subfloat[MTF-CBD]{\begin{minipage}[t]{0.095\linewidth}
		\begin{annotatedFigure}
			{\adjincludegraphics[width=\linewidth,   trim={{.1\width} {.0\width} {.55\width} {.7\width}},clip]	{fig/imgWV3/imgWV3_rd_bdsd_pc.jpg}};
		\end{annotatedFigure}\vspace{0.5mm}
		\begin{annotatedFigure}
			{\adjincludegraphics[width=\linewidth,   trim={{.1\width} {.0\width} {.55\width} {.7\width}},clip]	{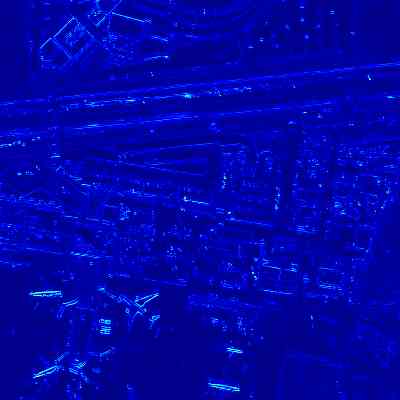}};
		\end{annotatedFigure}\vspace{0.5mm}
		\begin{annotatedFigure}
			{\adjincludegraphics[width=\linewidth,   trim={{.1\width} {.0\width} {.55\width} {.7\width}},clip]	{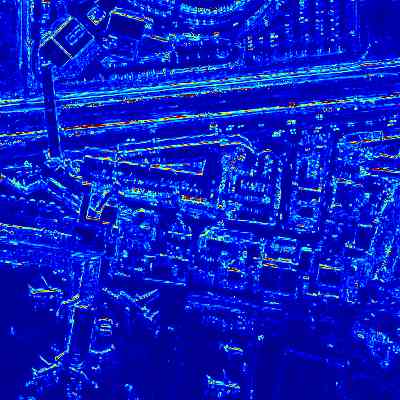}};
		\end{annotatedFigure}\vspace{0.5mm}
		\begin{annotatedFigure}
			{\adjincludegraphics[width=\linewidth,   trim={{.1\width} {.5\width} {.4\width} {.0\width}},clip]	{fig/imgWV3/imgWV3_rd_bdsd_pc.jpg}};
		\end{annotatedFigure}\vspace{0.5mm}
		\begin{annotatedFigure}
			{\adjincludegraphics[width=\linewidth,  trim={{.1\width} {.5\width} {.4\width} {.0\width}},clip]	{fig/imgWV3/imgWV3_rmse_cbd.jpg}};
		\end{annotatedFigure}\vspace{0.5mm}
		\begin{annotatedFigure}
			{\adjincludegraphics[width=\linewidth,   trim={{.1\width} {.5\width} {.4\width} {.0\width}},clip]	{fig/imgWV3/imgWV3_sam_cbd.jpg}};
		\end{annotatedFigure}\vspace{1mm}
	\end{minipage}\label{fig:wv3_rd:e}}\hspace{0.001mm}
	\subfloat[\hspace{-1.1mm}GLP-Reg-FS]{\begin{minipage}[t]{0.095\linewidth}
		\begin{annotatedFigure}
			{\adjincludegraphics[width=\linewidth,   trim={{.1\width} {.0\width} {.55\width} {.7\width}},clip]	{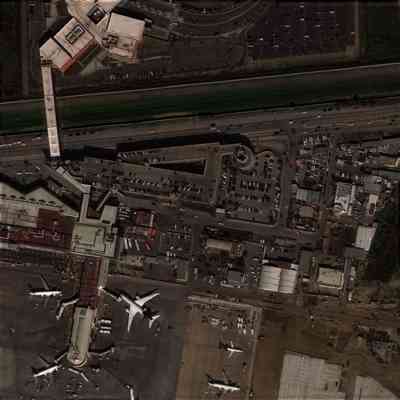}};
		\end{annotatedFigure}\vspace{0.5mm}
		\begin{annotatedFigure}
			{\adjincludegraphics[width=\linewidth,   trim={{.1\width} {.0\width} {.55\width} {.7\width}},clip]	{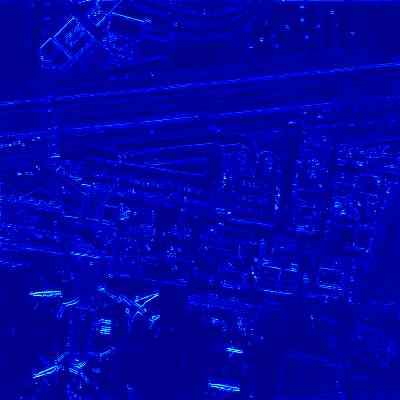}};
		\end{annotatedFigure}\vspace{0.5mm}
		\begin{annotatedFigure}
			{\adjincludegraphics[width=\linewidth,   trim={{.1\width} {.0\width} {.55\width} {.7\width}},clip]	{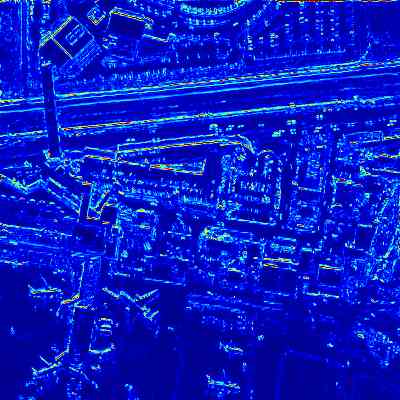}};
		\end{annotatedFigure}\vspace{0.5mm}
		\begin{annotatedFigure}
			{\adjincludegraphics[width=\linewidth,   trim={{.1\width} {.5\width} {.4\width} {.0\width}},clip]	{fig/imgWV3/imgWV3_rd_glp_reg_fs.jpg}};
		\end{annotatedFigure}\vspace{0.5mm}
		\begin{annotatedFigure}
			{\adjincludegraphics[width=\linewidth,   trim={{.1\width} {.5\width} {.4\width} {.0\width}},clip]	{fig/imgWV3/imgWV3_rmse_glp_reg_fs.jpg}};
		\end{annotatedFigure}\vspace{0.5mm}
		\begin{annotatedFigure}
			{\adjincludegraphics[width=\linewidth,   trim={{.1\width} {.5\width} {.4\width} {.0\width}},clip]	{fig/imgWV3/imgWV3_sam_glp_reg_fs.jpg}};
		\end{annotatedFigure}\vspace{1mm}
	\end{minipage}\label{fig:wv3_rd:f}}\hspace{0.001mm}
	\subfloat[GSA-Segm]{\begin{minipage}[t]{0.095\linewidth}
		\begin{annotatedFigure}
			{\adjincludegraphics[width=\linewidth,   trim={{.1\width} {.0\width} {.55\width} {.7\width}},clip]	{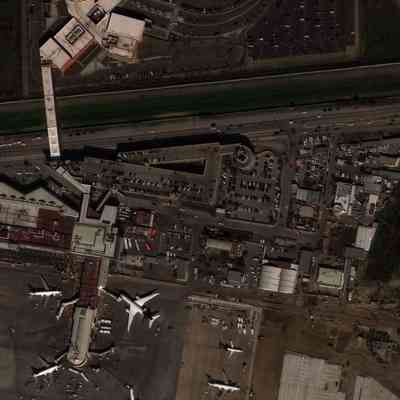}};
		\end{annotatedFigure}\vspace{0.5mm}
		\begin{annotatedFigure}
			{\adjincludegraphics[width=\linewidth,  trim={{.1\width} {.0\width} {.55\width} {.7\width}},clip]	{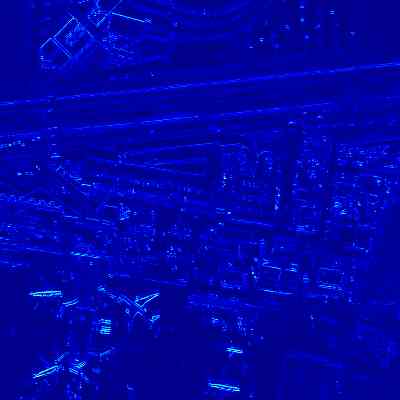}};
		\end{annotatedFigure}\vspace{0.5mm}
		\begin{annotatedFigure}
			{\adjincludegraphics[width=\linewidth,  trim={{.1\width} {.0\width} {.55\width} {.7\width}},clip]	{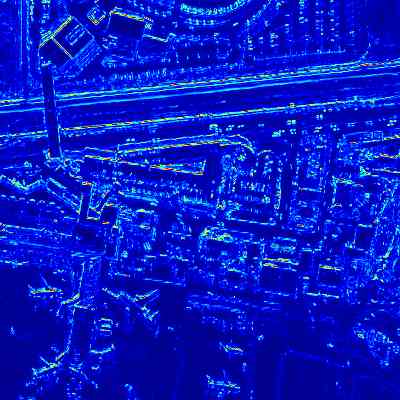}};
		\end{annotatedFigure}\vspace{0.5mm}
		\begin{annotatedFigure}
			{\adjincludegraphics[width=\linewidth,   trim={{.1\width} {.5\width} {.4\width} {.0\width}},clip]	{fig/imgWV3/imgWV3_rd_seg_gsa.jpg}};
		\end{annotatedFigure}\vspace{0.5mm}
		\begin{annotatedFigure}
			{\adjincludegraphics[width=\linewidth,   trim={{.1\width} {.5\width} {.4\width} {.0\width}},clip]	{fig/imgWV3/imgWV3_rmse_seg_gsa.jpg}};
		\end{annotatedFigure}\vspace{0.5mm}
		\begin{annotatedFigure}
			{\adjincludegraphics[width=\linewidth,   trim={{.1\width} {.5\width} {.4\width} {.0\width}},clip]	{fig/imgWV3/imgWV3_sam_seg_gsa.jpg}};
		\end{annotatedFigure}\vspace{1mm}
	\end{minipage}\label{fig:wv3_rd:g}}\hspace{0.001mm}
	\subfloat[GLP-Segm]{\begin{minipage}[t]{0.095\linewidth}
		\begin{annotatedFigure}
			{\adjincludegraphics[width=\linewidth,   trim={{.1\width} {.0\width} {.55\width} {.7\width}},clip]	{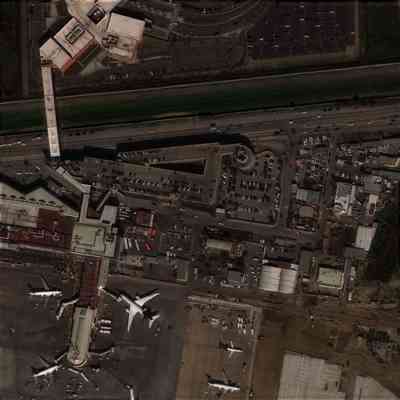}};
		\end{annotatedFigure}\vspace{0.5mm}
		\begin{annotatedFigure}
			{\adjincludegraphics[width=\linewidth,   trim={{.1\width} {.0\width} {.55\width} {.7\width}},clip]	{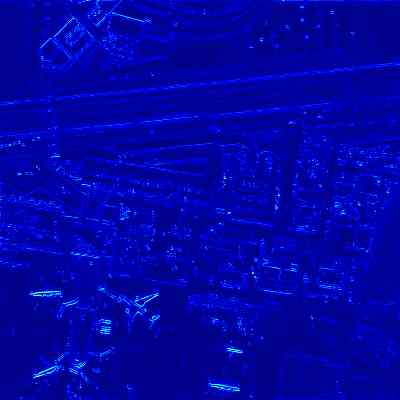}};
		\end{annotatedFigure}\vspace{0.5mm}
		\begin{annotatedFigure}
			{\adjincludegraphics[width=\linewidth,   trim={{.1\width} {.0\width} {.55\width} {.7\width}},clip]	{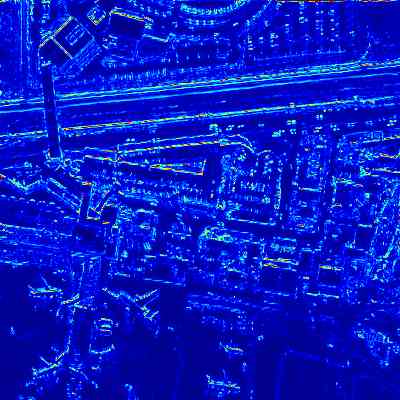}};
		\end{annotatedFigure}\vspace{0.5mm}
		\begin{annotatedFigure}
			{\adjincludegraphics[width=\linewidth,   trim={{.1\width} {.5\width} {.4\width} {.0\width}},clip]	{fig/imgWV3/imgWV3_rd_seg_glp.jpg}};
		\end{annotatedFigure}\vspace{0.5mm}
		\begin{annotatedFigure}
			{\adjincludegraphics[width=\linewidth,   trim={{.1\width} {.5\width} {.4\width} {.0\width}},clip]	{fig/imgWV3/imgWV3_rmse_seg_glp.jpg}};
		\end{annotatedFigure}\vspace{0.5mm}
		\begin{annotatedFigure}
			{\adjincludegraphics[width=\linewidth,   trim={{.1\width} {.5\width} {.4\width} {.0\width}},clip]	{fig/imgWV3/imgWV3_sam_seg_glp.jpg}};
		\end{annotatedFigure}\vspace{1mm}
	\end{minipage}\label{fig:wv3_rd:h}}\hspace{0.001mm}
	\subfloat[\hspace{-0.5mm}Target-CNN]{\begin{minipage}[t]{0.095\linewidth}
		\begin{annotatedFigure}
			{\adjincludegraphics[width=\linewidth,  trim={{.1\width} {.0\width} {.55\width} {.7\width}},clip]	{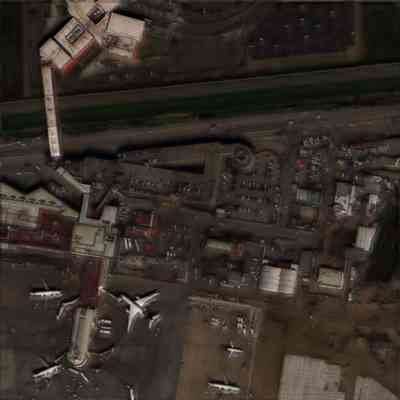}};
		\end{annotatedFigure}\vspace{0.5mm}
		\begin{annotatedFigure}
			{\adjincludegraphics[width=\linewidth,   trim={{.1\width} {.0\width} {.55\width} {.7\width}},clip]	{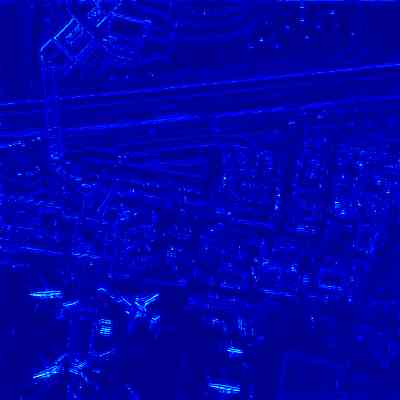}};
		\end{annotatedFigure}\vspace{0.5mm}
		\begin{annotatedFigure}
			{\adjincludegraphics[width=\linewidth,   trim={{.1\width} {.0\width} {.55\width} {.7\width}},clip]	{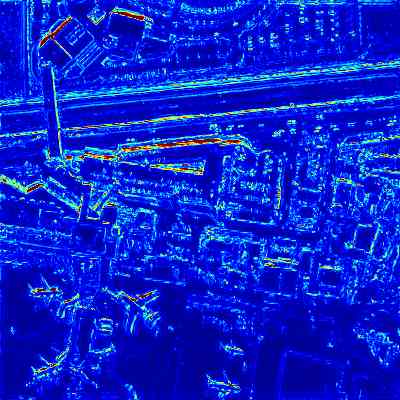}};
		\end{annotatedFigure}\vspace{0.5mm}
		\begin{annotatedFigure}
			{\adjincludegraphics[width=\linewidth,  trim={{.1\width} {.5\width} {.4\width} {.0\width}},clip]	{fig/imgWV3/imgWV3_rd_pancnn.jpg}};
		\end{annotatedFigure}\vspace{0.5mm}
		\begin{annotatedFigure}
			{\adjincludegraphics[width=\linewidth,   trim={{.1\width} {.5\width} {.4\width} {.0\width}},clip]	{fig/imgWV3/imgWV3_rmse_pancnn.jpg}};
		\end{annotatedFigure}\vspace{0.5mm}
		\begin{annotatedFigure}
			{\adjincludegraphics[width=\linewidth,   trim={{.1\width} {.5\width} {.4\width} {.0\width}},clip]	{fig/imgWV3/imgWV3_sam_pancnn.jpg}};
		\end{annotatedFigure}\vspace{1mm}
	\end{minipage}\label{fig:wv3_rd:i}}\hspace{0.001mm}
	\subfloat[Ours]{\begin{minipage}[t]{0.095\linewidth}
		\begin{annotatedFigure}
			{\adjincludegraphics[width=\linewidth,   trim={{.1\width} {.0\width} {.55\width} {.7\width}},clip]	{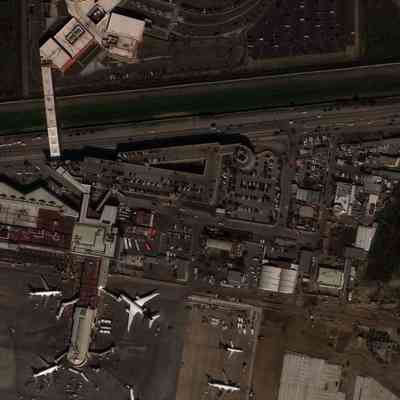}};
		\end{annotatedFigure}\vspace{0.5mm}
		\begin{annotatedFigure}
			{\adjincludegraphics[width=\linewidth,   trim={{.1\width} {.0\width} {.55\width} {.7\width}},clip]	{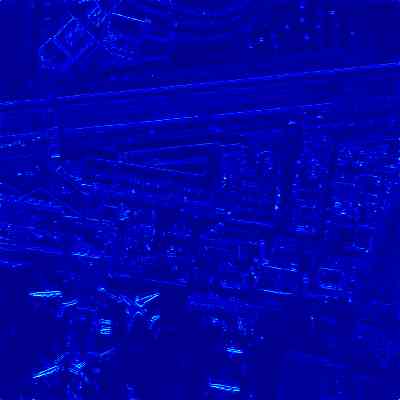}};
		\end{annotatedFigure}\vspace{0.5mm}
		\begin{annotatedFigure}
			{\adjincludegraphics[width=\linewidth,   trim={{.1\width} {.0\width} {.55\width} {.7\width}},clip]	{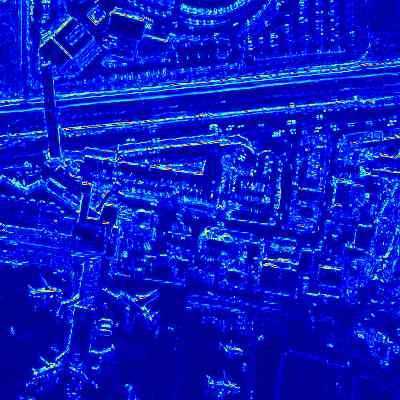}};
		\end{annotatedFigure}\vspace{0.5mm}
		\begin{annotatedFigure}
			{\adjincludegraphics[width=\linewidth,   trim={{.1\width} {.5\width} {.4\width} {.0\width}},clip]	{fig/imgWV3/imgWV3_rd_ours_2005.jpg}};
		\end{annotatedFigure}\vspace{0.5mm}
		\begin{annotatedFigure}
			{\adjincludegraphics[width=\linewidth,   trim={{.1\width} {.5\width} {.4\width} {.0\width}},clip]	{fig/imgWV3/imgWV3_rmse_ours_2.jpg}};
		\end{annotatedFigure}\vspace{0.5mm}
		\begin{annotatedFigure}
			{\adjincludegraphics[width=\linewidth,   trim={{.1\width} {.5\width} {.4\width} {.0\width}},clip]	{fig/imgWV3/imgWV3_sam_ours.jpg}};
		\end{annotatedFigure}\vspace{1mm}
	\end{minipage}\label{fig:wv3_rd:j}}\hspace{0.001mm}
	\begin{minipage}{1\linewidth}
		\centering
		\includegraphics[width=0.5\linewidth]{fig/colorbar.jpg}
	\end{minipage}
	\caption{Cropped patches of the pansharpened results from different methods on the WorldView3 dataset at reduced resolution. The results of the two patches are shown at the top three rows and the bottom three rows, respectively. For each patch, the top row shows the RGB channels of the pansharpened results. The middle row shows the average intensity difference map between the reconstructed and the ground truth MSI. And the bottom row shows the average spectral difference map between the reconstructed and the ground truth MSI. The first column shows the ground truth MSI (GT), LR MSI (M), and panchromatic (P) images for the two patches, respectively.}
	\label{fig:wv3_rd}
\end{figure*}

\begin{table}[htbp]
	\caption{Performance comparison on the WorldView-3 dataset at reduced resolution.}
	\label{tab:wv3_rd}
	\centering
	\begin{tabular}{c| c c c c}
		\hline
		& PSNR & SAM & ERGAS & $Q2^n$ \\
		\hline
		GSA & 24.6003 & 6.1272 & 7.2409 & 0.7823 \\
		PRACS & 24.8127 & 6.1096 & 7.0256 & 0.7816 \\
		BDSD-PC & 24.3359 &6.1978 &7.475 & 0.7897 \\
		MTF-CBD & 23.9003 & 6.7134 & 7.8236 & 0.7736 \\
		GLP-Reg-FS & 24.4321 & 6.3271 & 7.3874 & 0.7783 \\
		GSA-Segm& 24.5905 & 5.8829 & 7.2274 & 0.796 \\
		GLP-Segm & 24.5872 & 5.93 & 7.2179 & 0.78211\\
		Ours &\textbf{25.0487}&\textbf{5.8202} &\textbf{6.8566} &\textbf{0.8141} \\
		\hline
		Target-CNN & 24.4776 & 7.4056 & 7.3207 & 0.7607 \\
		HR MSI & $+\infty$ & 0 & 0 & 1 \\
		\hline  
	\end{tabular}%
\end{table}

The visual results on the Geoeye-1 dataset at reduced resolution are presented in Fig.~\ref{fig:geoeye_rd}. The reconstructed two patches are shown at the top three rows and the bottom three rows, respectively. For better inspection, we show the patches of the final results constructed with its red, green and blue channels, such that both the spatial distortion and spectral distortion are visible. To better visualize the difference, we show the average intensity difference map and spectral difference map (calculated from SAM) across all the bands between the reconstructed MSI and ground truth MSI. The results are color-coded according to the colorbar at the bottom of the figure. From left to right, the colors illustrate the differences from the minimum to the maximum. We can observe that CS-based PRACS~\cite{choi2010new} loss a lot of detail information in the recovered MSI, thus the result is quite blur as shown in Fig.~\ref{fig:geoeye_rd:c}. BDSD-PC~\cite{vivone2019robust} recovers more details than PRACS. From the intensity and the spectral difference maps shown in Fig.~\ref{fig:geoeye_rd:b}, we observe that GSA~\cite{aiazzi2007improving} performs the best among all the CS-based approaches in this case. 
The MTF-based method, MTF-CBD~\cite{aiazzi2006mtf}, recovers more details than the CS-based PRACS~\cite{choi2010new}. However, it also introduces some spectral distortions as shown in the spectral difference maps in Fig.~\ref{fig:geoeye_rd:e}. GLP-Reg-FS~\cite{vivone2018full} preserves the spectral information better than MTF-CBD. The context-based approach, GSA-Segm shown in Fig.~\ref{fig:geoeye_rd:g} and GLP-Segm shown in Fig.~\ref{fig:geoeye_rd:h}~\cite{restaino2017context} have some blur effects in this scenario. This is because it estimates the injection coefficients according to the raw MSI with mixed pixels which may reduce its reconstruction performance. The deep-learning based approach, Target-CNN~\cite{scarpa2018target}, works well, because the model is trained on its ground truth Geoeye-1 MSI in Fig.~\ref{fig:geoeye_rd:a}. The proposed UP-SAM, recovers more details than the supervised Target-CNN, the CS-based and MRA-based methods, while at the same time preserving spectral information well. This is because the extraction and injection functions of the proposed method are defined based on the attention representations that indicating spectral characteristics of the MSI with sub-pixel accuracy.

For quantitative comparison, the evaluation results are reported in Table~\ref{tab:geoeye_rd}. If the reconstructed MSI is the same as the ground truth MSI, we should obtain the measured results listed in the last row of the table. We can observe that the proposed method performs the best among all the unsupervised methods, and it is comparable to (if not better than) the supervised method, Target-CNN, trained with the ground truth MSI.

Similar behaviors can be observed on the Ikonos dataset as demonstrated in Fig.~\ref{fig:imgIK_rd} and Table~\ref{tab:Ikonos_rd}. Again, the proposed method works the best among all the CS-based and MRA-based methods, and presents comparable results than those from the supervised method.

The results of a more challenging dataset, WorldView2, are shown in Fig.~\ref{fig:wv2_rd} and Table~\ref{tab:wv2_rd}. The MSI in this dataset has 8 spectral bands, thus it is more difficult to preserve the spectral information than for the Geoeye-1 or Ikonos datasets. Both the supervised deep learning based method, Target-CNN, and the proposed unsupervised method, UP-SAM, are able to preserve the spectral information better than the other approaches. However, the Target-CNN has low scores in terms of PSNR, ERGAS, and $Q2^n$. From Fig.~\ref{fig:wv2_rd:i}, we could also observe that the reconstructed MSI from Target-CNN tends to miss high-frequency details even when the HR MSI is adopted during the training procedure. This is because the trained injection function may not be sufficient for pixels possessing different spectral characteristics. On the contrary, although the proposed method is unsupervised, it could reconstruct a sharper MSI as shown in Fig.~\ref{fig:wv2_rd:j}. This is due to the design that both the extraction and injection functions of the proposed method vary according to pixels' spectral characteristics.

The last dataset WorldView3 is the most challenging one, because there is a slight mismatch between the panchromatic and MSI. Also this dataset has not been used to train the supervised Target-CNN. From Fig.~\ref{fig:wv3_rd} and Table~\ref{tab:wv3_rd}, we can observe that, most unsupervised methods work better than the Target-CNN. This is because the spectral characteristics of the WorldView3 dataset are distinctive from those of the other three datasets, which have not been learned by Target-CNN during the training procedure. Therefore, the trained mapping function fails in this situation. On the contrary, due to the self-attention mechanism, the proposed method constantly performs the best among all the methods regardless of the type of images tested.

\begin{table*}[htbp]
	\caption{Performance comparison on different datasets at full resolution.}
	\label{tab:full}
	\centering
	\begin{tabular}{l | c c c | c c c | c c c | c c c}
		\hline
		{}&\multicolumn{3}{c|}{GeoEye-1}&\multicolumn{3}{c|}{IKONOS}&\multicolumn{3}{c|}{WorldView2}&\multicolumn{3}{c}{WorldView3}\\
		\hline
		& D$_{\lambda}$ & D$_S$ & QNR& D$_{\lambda}$ & D$_S$ & QNR& D$_{\lambda}$ & D$_S$ & QNR& D$_{\lambda}$ & D$_S$ & QNR\\
		\hline
		GSA & \textbf{\underline{0.1058}} & 0.1801 & 0.7332& 0.1341 & 0.1981 & 0.6944& 0.0486 & 0.15 & 0.8089& 0.0518 & 0.1788 & 0.7787\\
		PRACS & \textbf{0.0598} & \textbf{0.0206} & \textbf{0.9208} & \textbf{0.0535} & \textbf{0.1099} & \textbf{0.8425}& \textbf{0.0109} & \textbf{0.0981} & \textbf{0.892}& \textbf{\underline{0.0214}} & \textbf{0.0864} & \textbf{0.8941}\\
		BDSD-PC & 0.1078 & 0.1703 & 0.7402& 0.1105 & 0.1514 & 0.7548& 0.0464 & {0.1457} & 0.8147& \textbf{0.0116} &0.2032 & 0.7876\\
		MTF-CBD & 0.2412 & \textbf{\underline{0.1308}} & 0.6596 & 0.1763 & \textbf{\underline{0.1275}} & 0.7186& 0.0967 & \textbf{\underline{0.0993}} & 0.8136& 0.0584 & 0.1198 & 0.8288\\
		GLP-Reg-FS & 0.1111 & 0.1743 & 0.7339 & 0.1566 & 0.2016 & 0.6733& 0.0766 & 0.1464 & 0.7882& 0.0518 & 0.1314 &  0.8237\\
		GSA-Segm& 0.1530 & 0.1878 & 0.6680& 0.1220 & 0.1658 & 0.7325& 0.0617& 0.156 & 0.792 & 0.0583 & 0.1456 & 0.8046\\
		GLP-Segm& 0.1177 & 0.1709 & 0.7315& 0.1516 & 0.1986 & 0.6799& 0.0708& 0.1446& 0.7948& 0.0439 & 0.1219 & 0.8395\\
		Ours & 0.1117&0.1661 &\textbf{\underline{0.7407}}& \textbf{\underline{0.0996}}&0.1499 &\textbf{\underline{0.7654}}&\textbf{\underline{0.0459}} &0.1429 & \textbf{\underline{0.8178}}& 0.0333& \textbf{\underline{0.1083}}&\textbf{\underline{0.8619}} \\
		\hline
		Target-CNN & 0.0237 & 0.0620 & 0.9157& 0.0241 & 0.0450 & 0.9320& 0.0575 & 0.0351 & 0.9094& 0.0939 & 0.0584 & 0.8532\\
		HR MSI & 0 & 0 & 1& 0 & 0 & 1& 0 & 0 & 1&0 & 0 & 1\\
		\hline  
	\end{tabular}%
\end{table*}

\begin{figure*}[htbp]
	\centering
	\begin{minipage}{0.95\linewidth}
		\begin{minipage}{1\linewidth}
			\subfloat[LR MSI]{\begin{minipage}[t]{0.19\linewidth}
					\begin{annotatedFigure}
						{\adjincludegraphics[width=\linewidth,   trim={{.04\width} {.02\width} {.885\width} {.92\width}},clip]	{fig/imgGE/imgGE_rd_msi_sc.jpg}};
					\end{annotatedFigure}\vspace{0.5mm}
					\begin{annotatedFigure}
						{\adjincludegraphics[width=\linewidth,  trim={{.95\width} {.95\width} {.0\width} {.01\width}},clip]{fig/imgGE/imgGE_rd_msi_sc.jpg}};
					\end{annotatedFigure}\vspace{0.5mm}
				\end{minipage}\label{fig:geoeye:a}}\hspace{0.001mm}
			\subfloat[GSA]{\begin{minipage}[t]{0.19\linewidth}
					\begin{annotatedFigure}
						{\adjincludegraphics[width=\linewidth,  trim={{.04\width} {.02\width} {.885\width} {.92\width}},clip]	{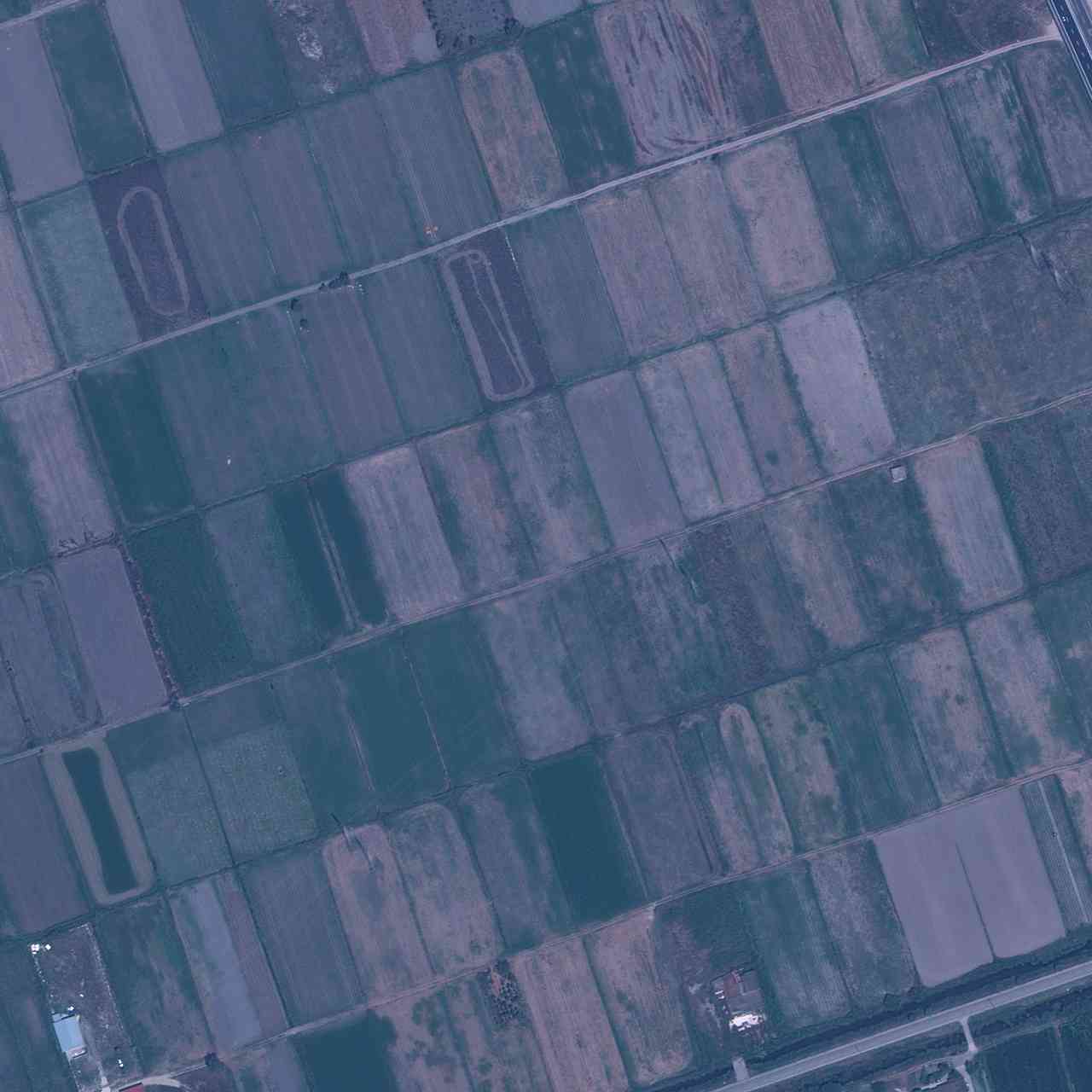}};
					\end{annotatedFigure}\vspace{0.5mm}
					\begin{annotatedFigure}
						{\adjincludegraphics[width=\linewidth, trim={{.95\width} {.95\width} {.0\width} {.01\width}},clip]	{fig/imgGE/imgGE_full_gsa_sc.jpg}};
					\end{annotatedFigure}\vspace{0.5mm}
				\end{minipage}\label{fig:geoeye:b}}\hspace{0.001mm}
			\subfloat[PRACS]{\begin{minipage}[t]{0.19\linewidth}
					\begin{annotatedFigure}
						{\adjincludegraphics[width=\linewidth,   trim={{.04\width} {.02\width} {.885\width} {.92\width}},clip]		{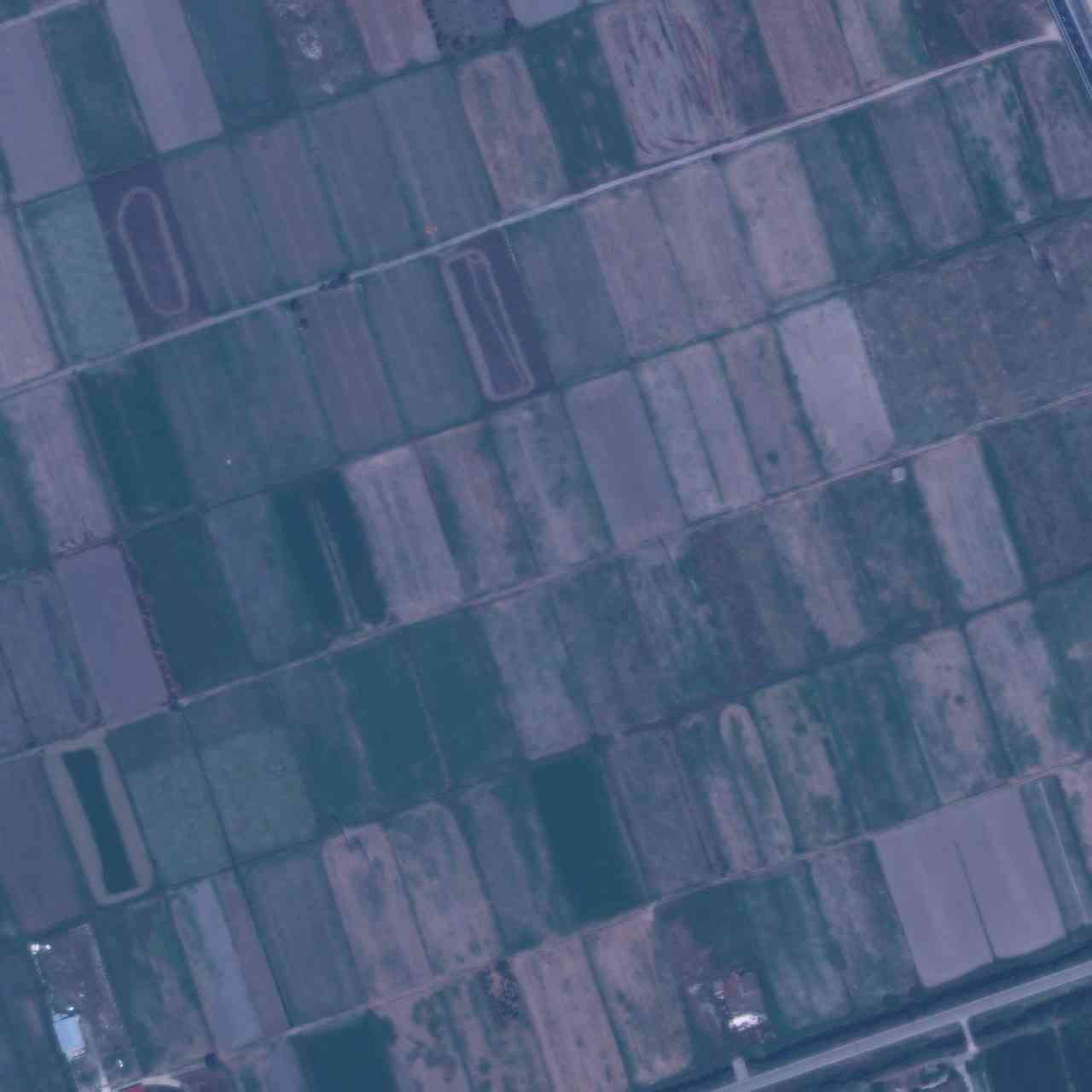}};
					\end{annotatedFigure}\vspace{0.5mm}
					\begin{annotatedFigure}
						{\adjincludegraphics[width=\linewidth,   trim={{.95\width} {.95\width} {.0\width} {.01\width}},clip]	{fig/imgGE/imgGE_full_pracs_sc.jpg}};
					\end{annotatedFigure}\vspace{0.5mm}
				\end{minipage}\label{fig:geoeye:c}}\hspace{0.001mm}
			\subfloat[BDSD-PC]{\begin{minipage}[t]{0.19\linewidth}
					\begin{annotatedFigure}
						{\adjincludegraphics[width=\linewidth,   trim={{.04\width} {.02\width} {.885\width} {.92\width}},clip] {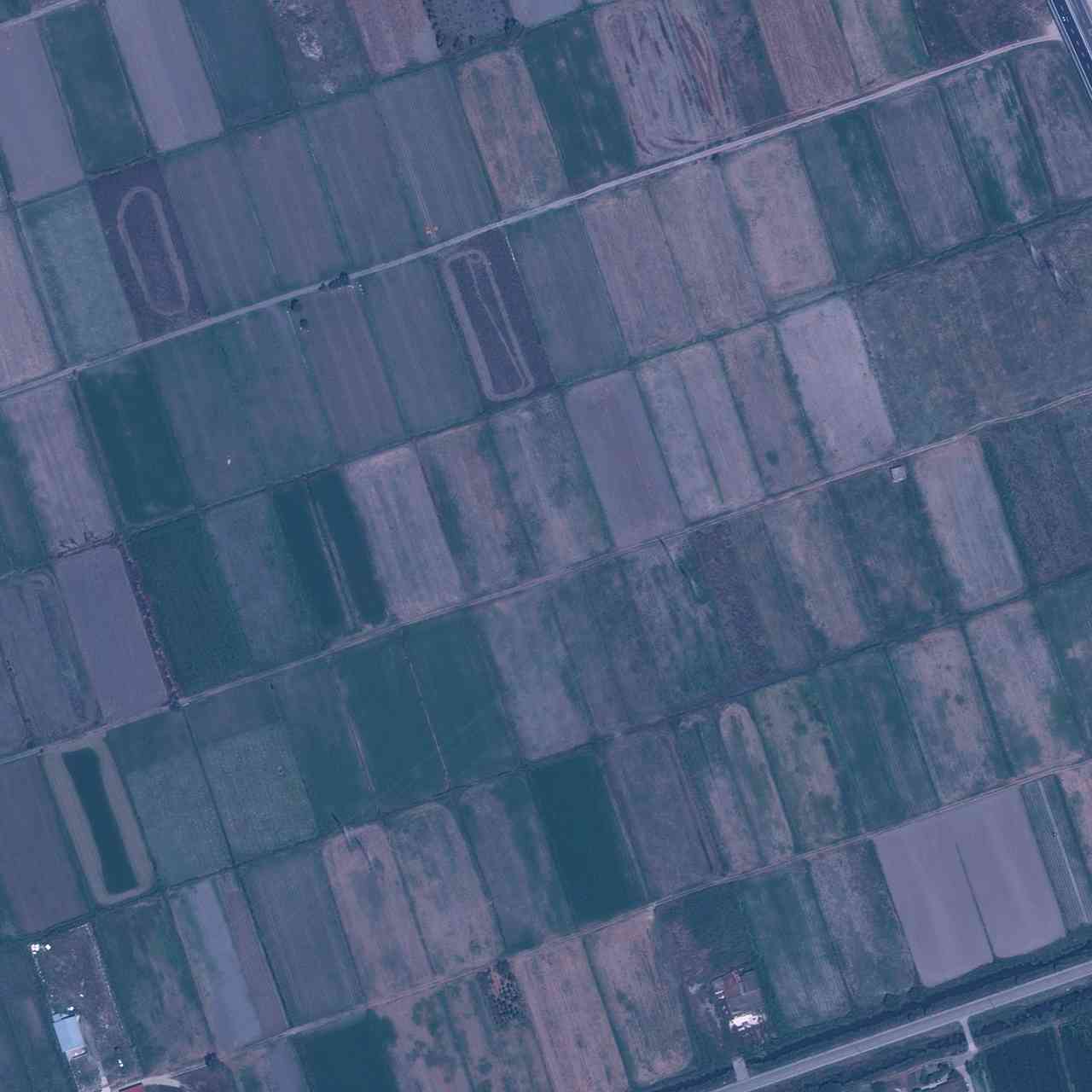}};
					\end{annotatedFigure}\vspace{0.5mm}
					\begin{annotatedFigure}
						{\adjincludegraphics[width=\linewidth,   trim={{.95\width} {.95\width} {.0\width} {.01\width}},clip] {fig/imgGE/imgGE_full_bdsd_pc_sc.jpg}};
					\end{annotatedFigure}\vspace{0.5mm}
			\end{minipage}\label{fig:geoeye:d}}\hspace{0.001mm}
			\subfloat[MTF-CBD]{\begin{minipage}[t]{0.19\linewidth}
					\begin{annotatedFigure}
						{\adjincludegraphics[width=\linewidth,   trim={{.04\width} {.02\width} {.885\width} {.92\width}},clip]	{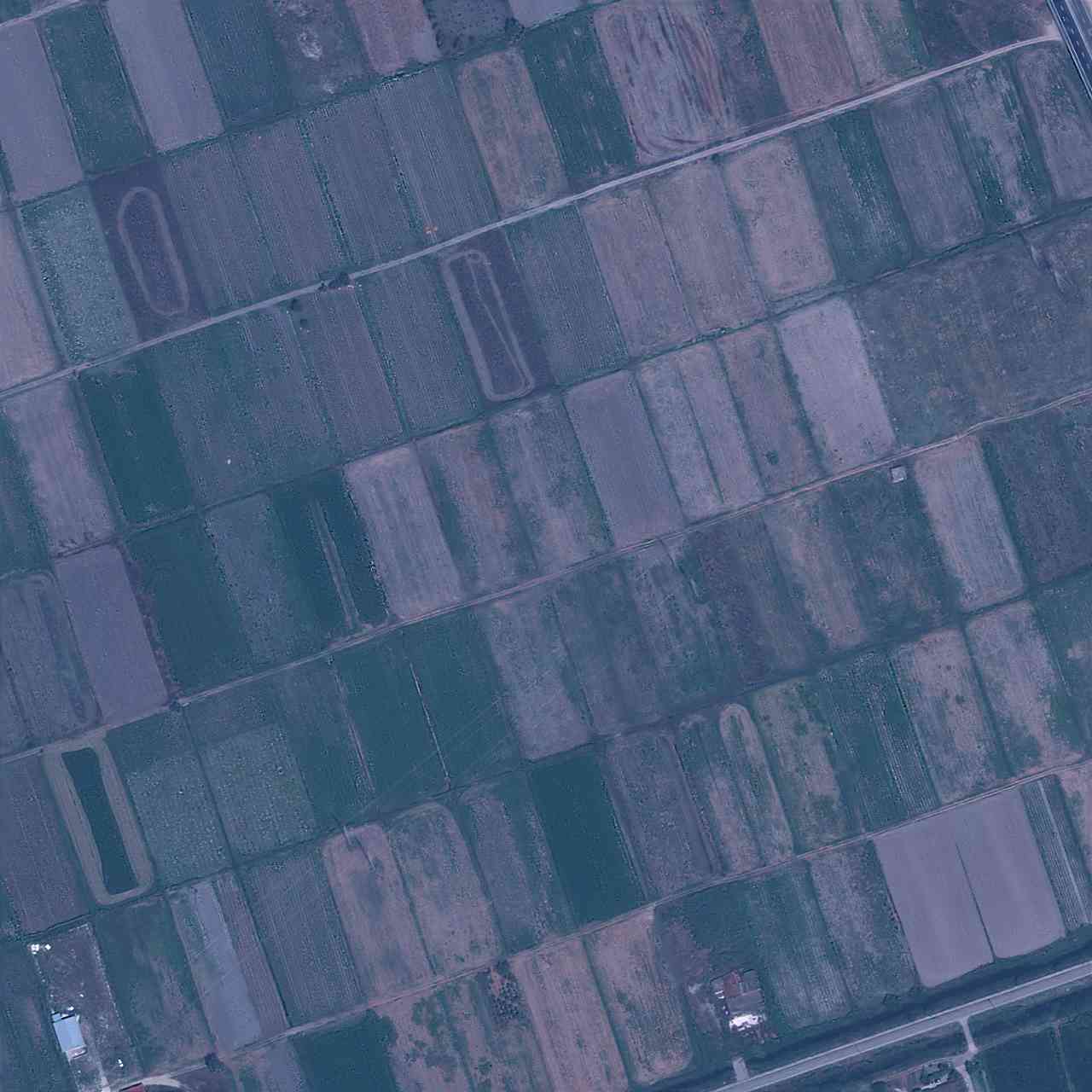}};
					\end{annotatedFigure}\vspace{0.5mm}
					\begin{annotatedFigure}
						{\adjincludegraphics[width=\linewidth,   trim={{.95\width} {.95\width} {.0\width} {.01\width}},clip]	{fig/imgGE/imgGE_full_cbd_sc.jpg}};
					\end{annotatedFigure}\vspace{0.5mm}
				\end{minipage}\label{fig:geoeye:e}}
		\end{minipage}
		\begin{minipage}{1\linewidth}
			\subfloat[GLP-Reg-FS]{\begin{minipage}[t]{0.19\linewidth}
					\begin{annotatedFigure}
						{\adjincludegraphics[width=\linewidth,    trim={{.04\width} {.02\width} {.885\width} {.92\width}},clip]	{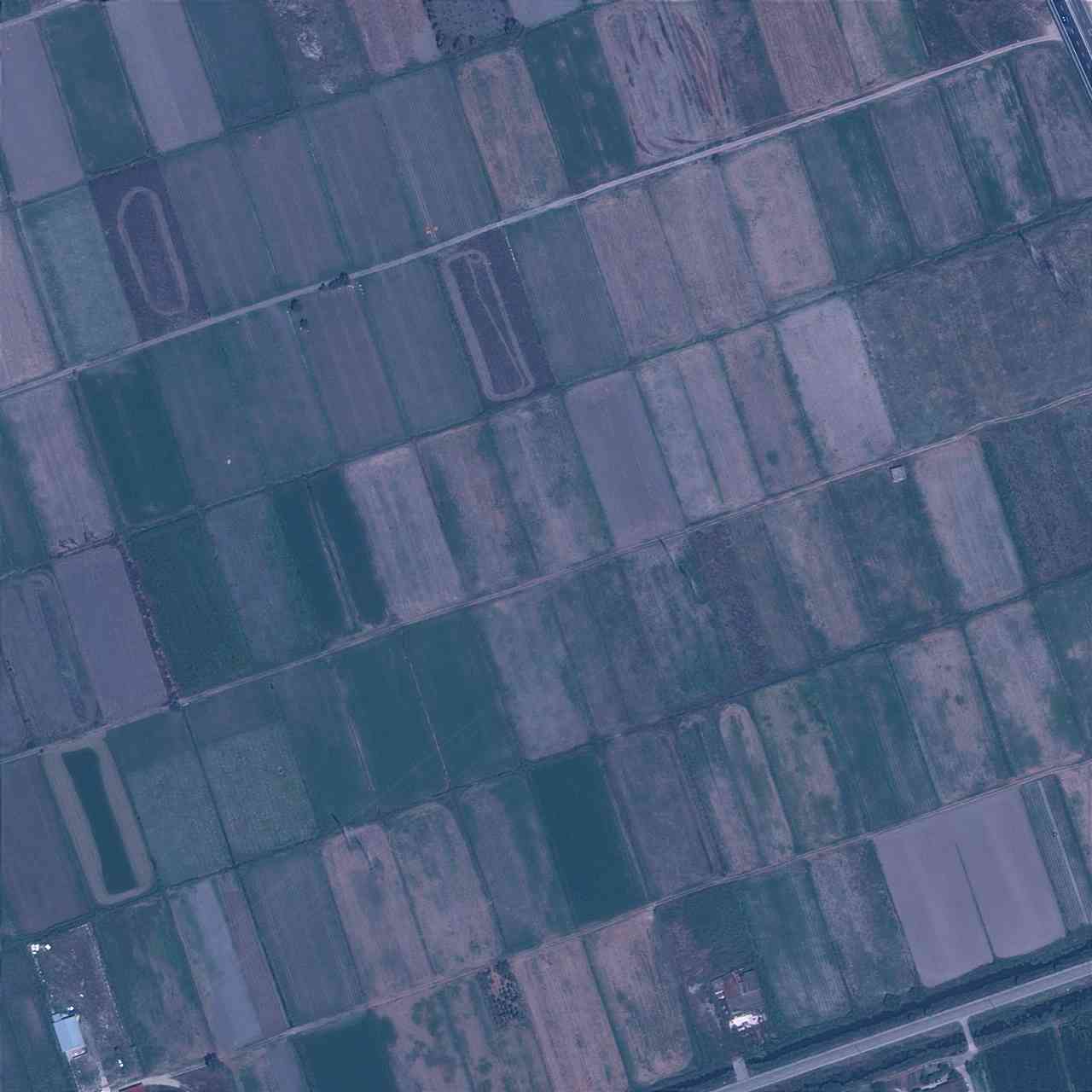}};
					\end{annotatedFigure}\vspace{0.5mm}
					\begin{annotatedFigure}
						{\adjincludegraphics[width=\linewidth,    trim={{.95\width} {.95\width} {.0\width} {.01\width}},clip]	{fig/imgGE/imgGE_full_glp_reg_fs_sc.jpg}};
					\end{annotatedFigure}\vspace{0.5mm}
				\end{minipage}\label{fig:geoeye:f}}\hspace{0.001mm}
			\subfloat[GSA-Segm]{\begin{minipage}[t]{0.19\linewidth}
					\begin{annotatedFigure}
						{\adjincludegraphics[width=\linewidth, trim={{.04\width} {.02\width} {.885\width} {.92\width}},clip] {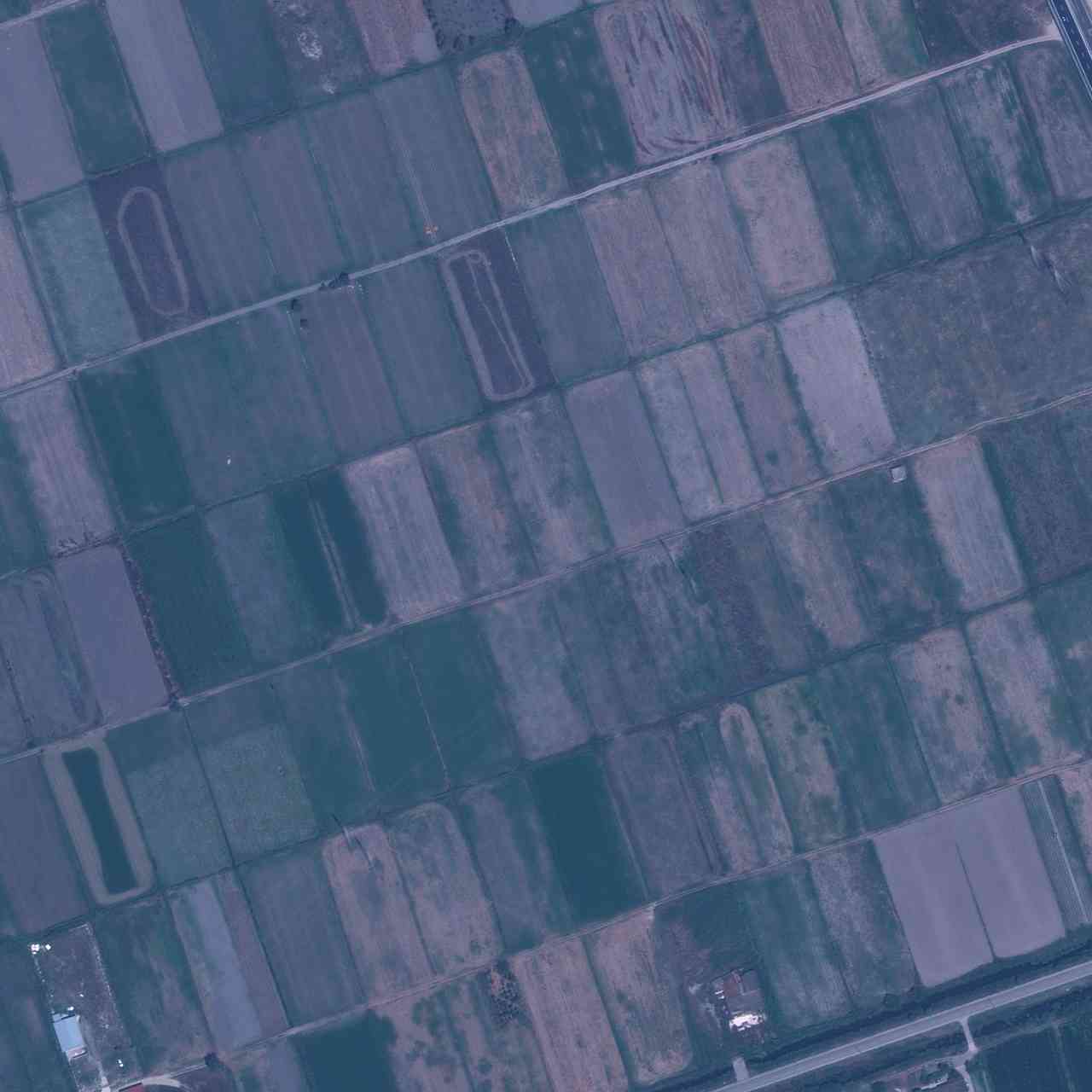}};
					\end{annotatedFigure}\vspace{0.5mm}
					\begin{annotatedFigure}
						{\adjincludegraphics[width=\linewidth,    trim={{.95\width} {.95\width} {.0\width} {.01\width}},clip]	{fig/imgGE/imgGE_full_seg_gsa_sc.jpg}};
					\end{annotatedFigure}\vspace{0.5mm}
				\end{minipage}\label{fig:geoeye:g}}\hspace{0.001mm}
			\subfloat[GLP-Segm]{\begin{minipage}[t]{0.19\linewidth}
					\begin{annotatedFigure}
						{\adjincludegraphics[width=\linewidth,   trim={{.04\width} {.02\width} {.885\width} {.92\width}},clip]		{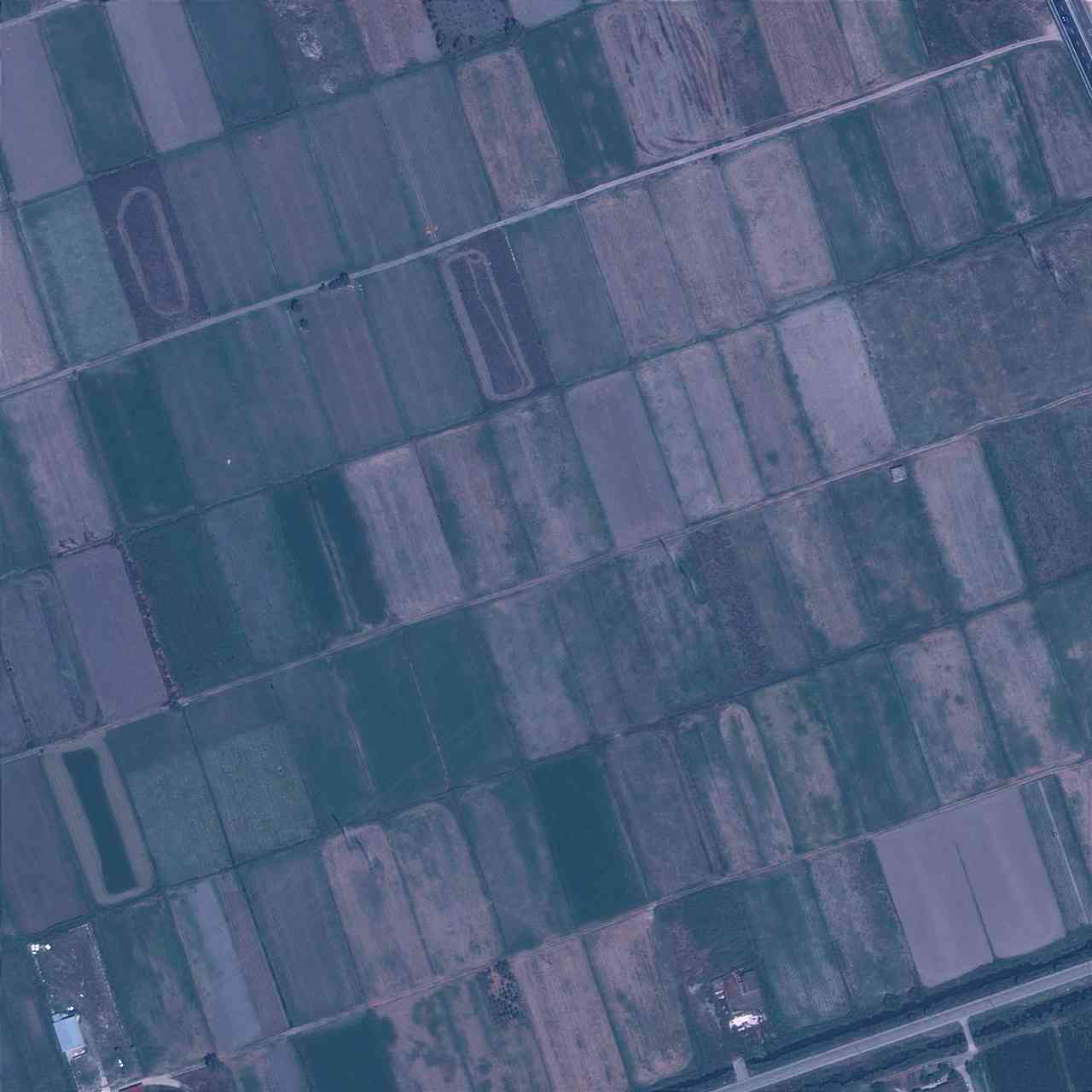}};
					\end{annotatedFigure}\vspace{0.5mm}
					\begin{annotatedFigure}
						{\adjincludegraphics[width=\linewidth,   trim={{.95\width} {.95\width} {.0\width} {.01\width}},clip]	{fig/imgGE/imgGE_full_seg_glp_sc.jpg}};
					\end{annotatedFigure}\vspace{0.5mm}
				\end{minipage}\label{fig:geoeye:h}}\hspace{0.001mm}
			\subfloat[Target-CNN]{\begin{minipage}[t]{0.19\linewidth}
					\begin{annotatedFigure}
						{\adjincludegraphics[width=\linewidth,  trim={{.04\width} {.02\width} {.885\width} {.92\width}},clip]		{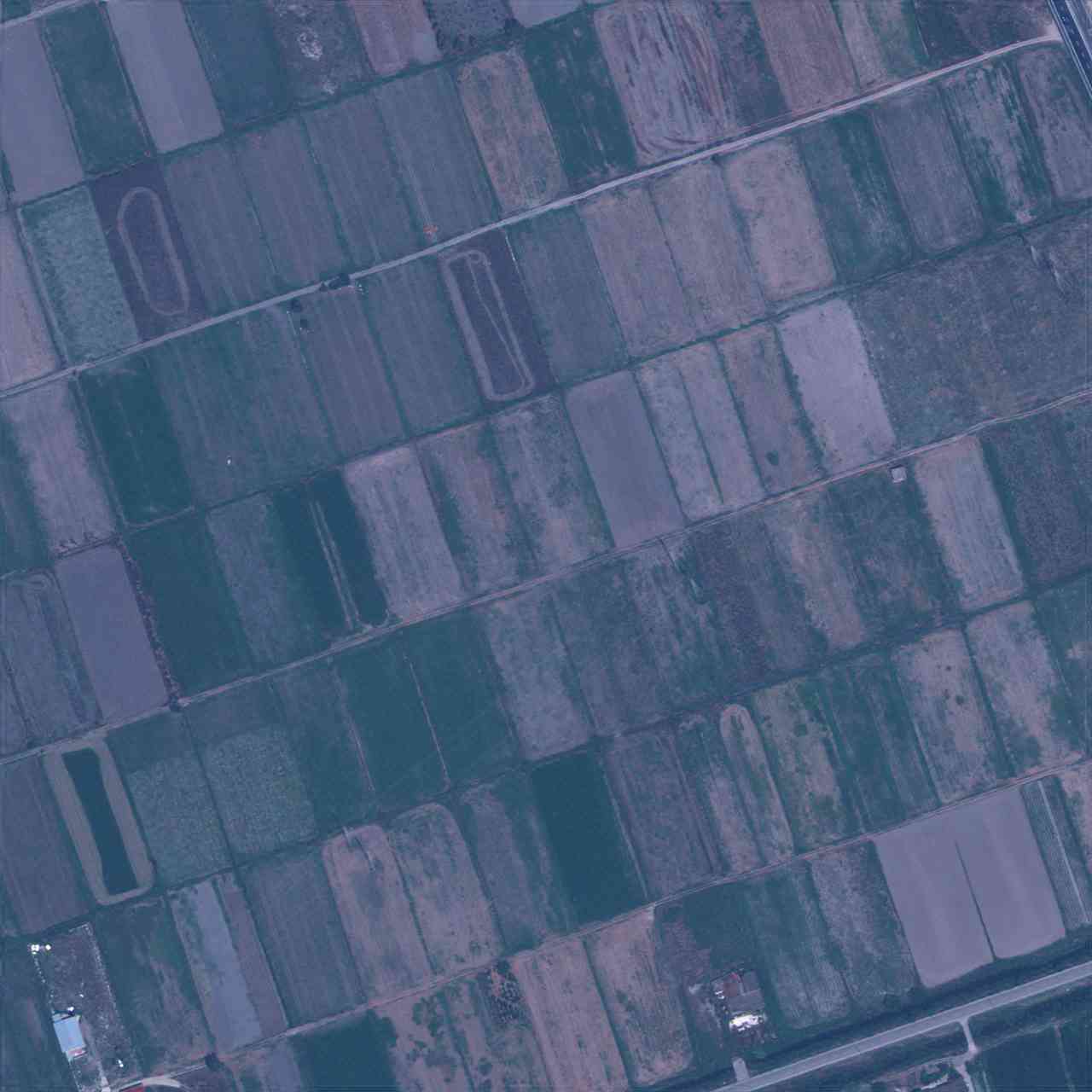}};
					\end{annotatedFigure}\vspace{0.5mm}
					\begin{annotatedFigure}
						{\adjincludegraphics[width=\linewidth,  trim={{.95\width} {.95\width} {.0\width} {.01\width}},clip]	{fig/imgGE/imgGE_full_pancnn_sc.jpg}};
					\end{annotatedFigure}\vspace{0.5mm}
				\end{minipage}\label{fig:geoeye:i}}\hspace{0.001mm}
			\subfloat[Ours]{\begin{minipage}[t]{0.19\linewidth}
					\begin{annotatedFigure}
						{\adjincludegraphics[width=\linewidth,   trim={{.04\width} {.02\width} {.885\width} {.92\width}},clip]	{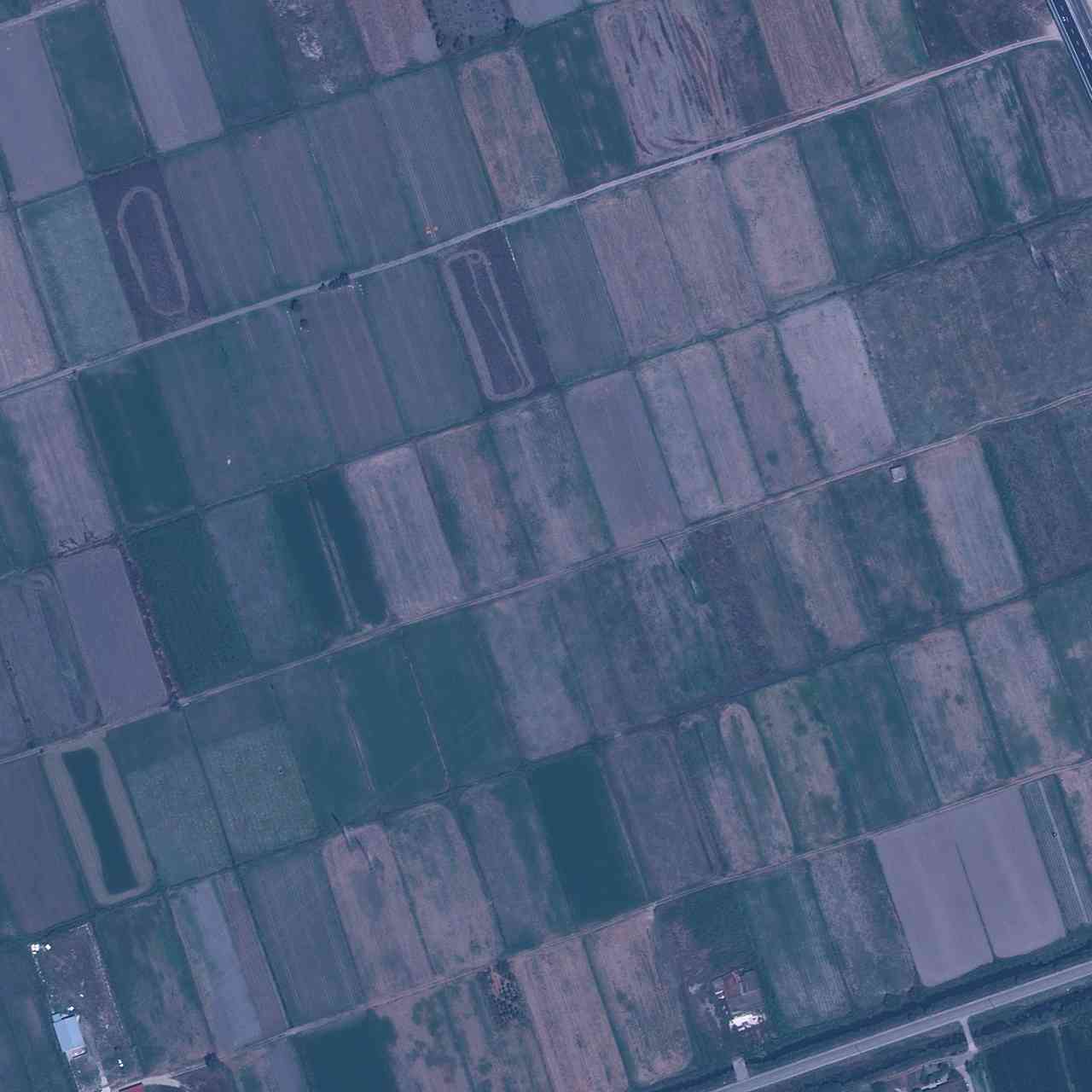}};
					\end{annotatedFigure}\vspace{0.5mm}
					\begin{annotatedFigure}
						{\adjincludegraphics[width=\linewidth,   trim={{.95\width} {.95\width} {.0\width} {.01\width}},clip]	{fig/imgGE/imgGE_full_ours_sc.jpg}};
					\end{annotatedFigure}\vspace{0.5mm}
				\end{minipage}\label{fig:geoeye:j}}
		\end{minipage}
		\caption{RGB channels of the pansharpened results from different methods on GeoEye-1 at full resolution.}
		\label{fig:geoeye}
	\end{minipage}
	\begin{minipage}{0.95\linewidth}
		\begin{minipage}{1\linewidth}
			\subfloat[\hspace{-0.5mm}LR MSI]{\begin{minipage}[t]{0.19\linewidth}
					\begin{annotatedFigure}
						{\adjincludegraphics[width=\linewidth,  trim={{.4\width} {.108\width} {.505\width} {.805\width}},clip]	{fig/imgIK/imgIK_rd_msi.jpg}};
					\end{annotatedFigure}\vspace{0.5mm}
					\begin{annotatedFigure}
						{\adjincludegraphics[width=\linewidth,  trim={{.15\width} {.76\width} {.75\width} {.18\width}},clip]	{fig/imgIK/imgIK_rd_msi.jpg}};
					\end{annotatedFigure}\vspace{0.5mm}
				\end{minipage}\label{fig:ik:a}}\hspace{0.001mm}
			\subfloat[GSA]{\begin{minipage}[t]{0.19\linewidth}
					\begin{annotatedFigure}
						{\adjincludegraphics[width=\linewidth,  trim={{.4\width} {.108\width} {.505\width} {.805\width}},clip]	{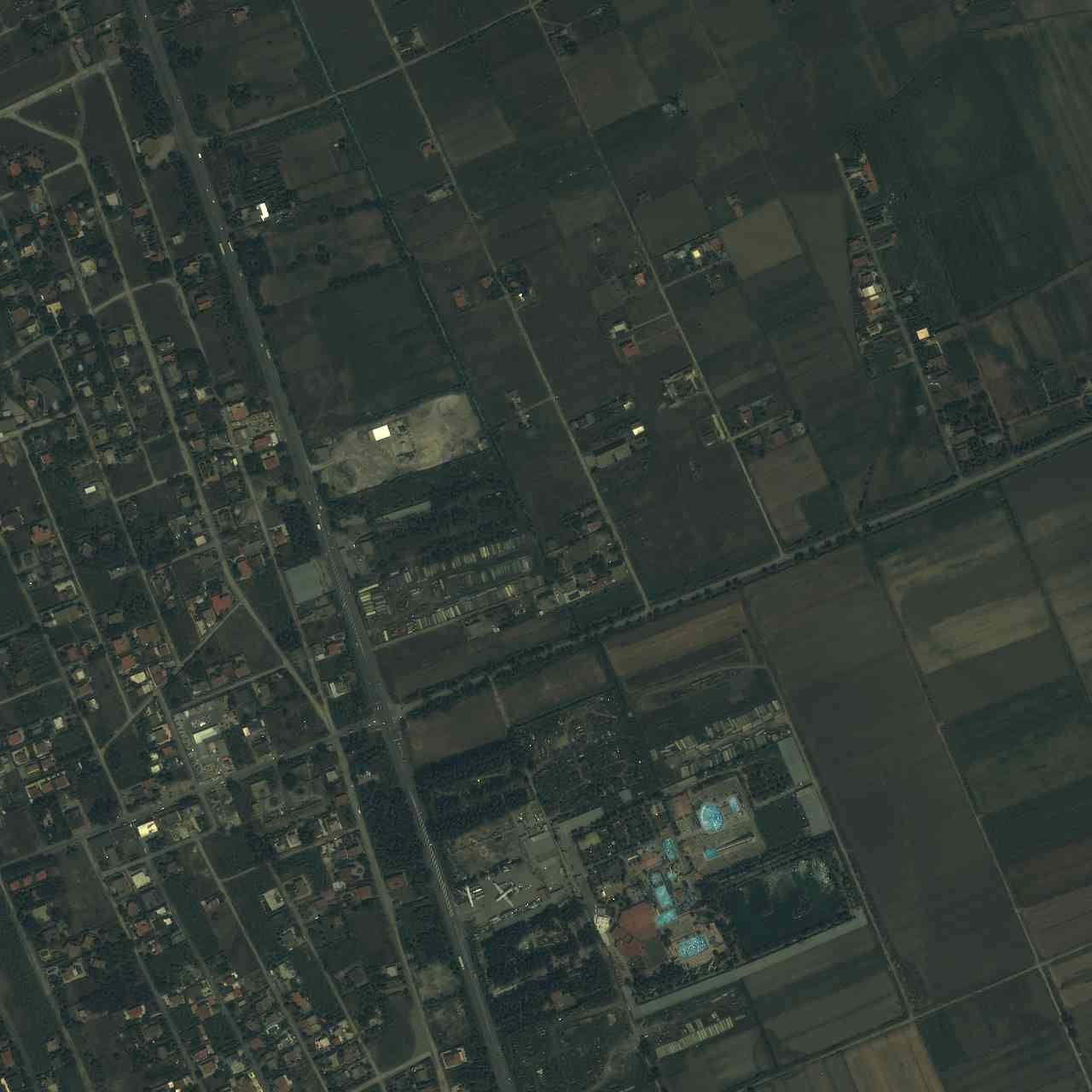}};
					\end{annotatedFigure}\vspace{0.5mm}
					\begin{annotatedFigure}
						{\adjincludegraphics[width=\linewidth,  trim={{.15\width} {.76\width} {.75\width} {.18\width}},clip]	{fig/imgIK/imgIK_full_gsa.jpg}};
					\end{annotatedFigure}\vspace{0.5mm}
				\end{minipage}\label{fig:ik:b}}\hspace{0.001mm}
			\subfloat[PRACS]{\begin{minipage}[t]{0.19\linewidth}
					\begin{annotatedFigure}
						{\adjincludegraphics[width=\linewidth,  trim={{.4\width} {.108\width} {.505\width} {.805\width}},clip]	{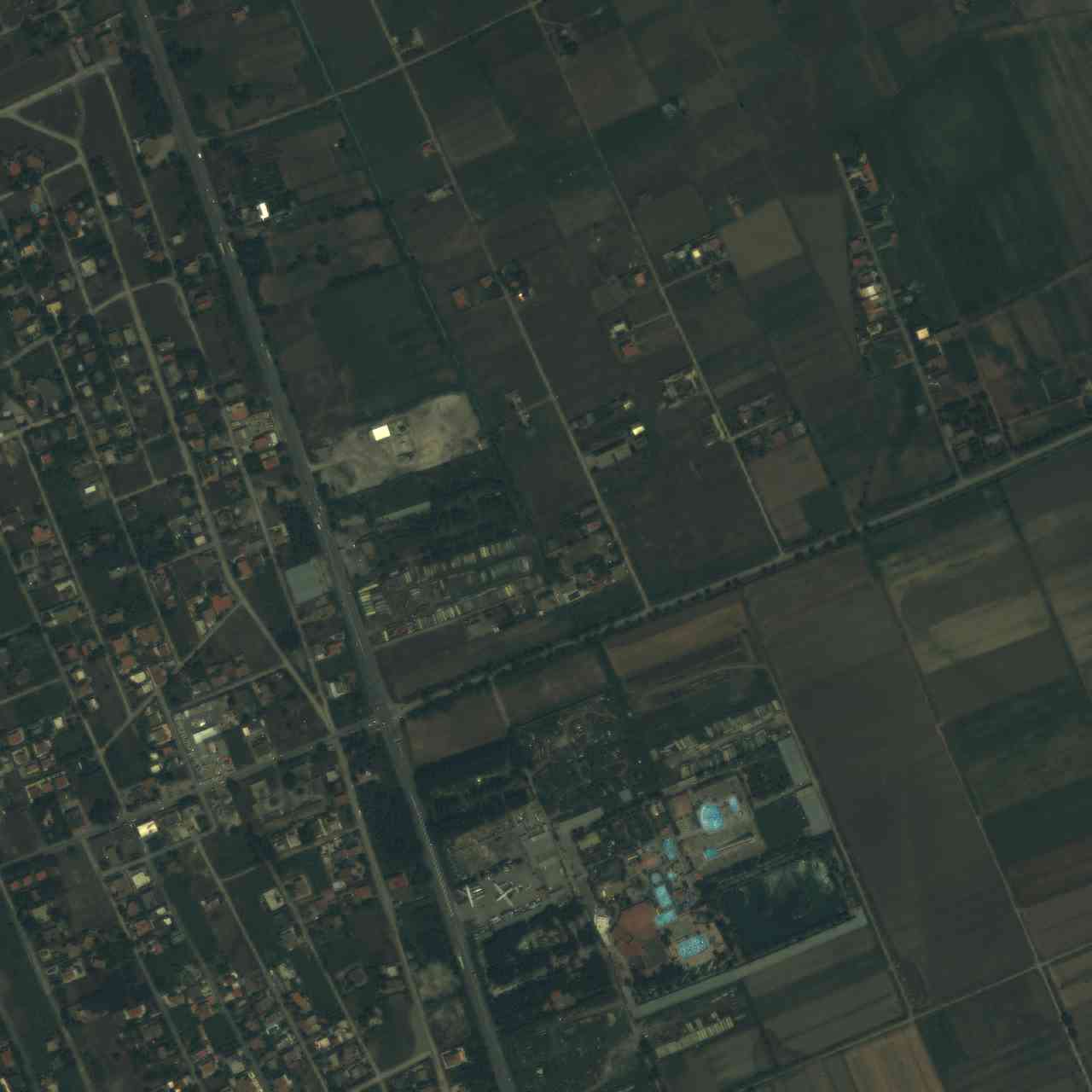}};
					\end{annotatedFigure}\vspace{0.5mm}
					\begin{annotatedFigure}
						{\adjincludegraphics[width=\linewidth,   trim={{.15\width} {.76\width} {.75\width} {.18\width}},clip]	{fig/imgIK/imgIK_full_pracs.jpg}};
					\end{annotatedFigure}\vspace{0.5mm}
				\end{minipage}\label{fig:ik:c}}\hspace{0.001mm}
			\subfloat[BDSD-PC]{\begin{minipage}[t]{0.19\linewidth}
					\begin{annotatedFigure}
						{\adjincludegraphics[width=\linewidth,  trim={{.4\width} {.108\width} {.505\width} {.805\width}},clip] {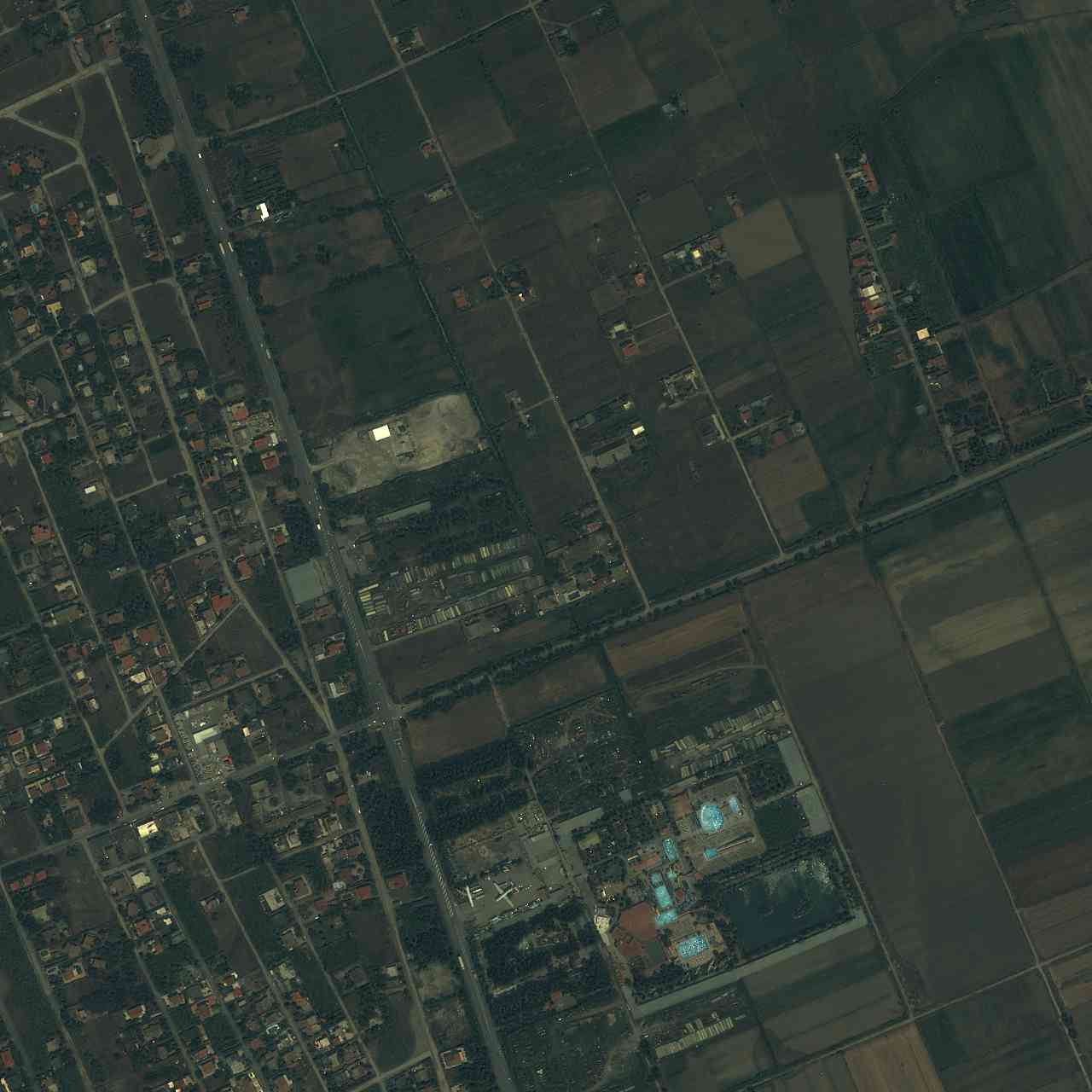}};
					\end{annotatedFigure}\vspace{0.5mm}
					\begin{annotatedFigure}
						{\adjincludegraphics[width=\linewidth,  trim={{.15\width} {.76\width} {.75\width} {.18\width}},clip]	{fig/imgIK/imgIK_full_bdsd_pc.jpg}};
					\end{annotatedFigure}\vspace{0.5mm}
				\end{minipage}\label{fig:ik:d}}\hspace{0.001mm}
			\subfloat[MTF-CBD]{\begin{minipage}[t]{0.19\linewidth}
					\begin{annotatedFigure}
						{\adjincludegraphics[width=\linewidth,  trim={{.4\width} {.108\width} {.505\width} {.805\width}},clip]	{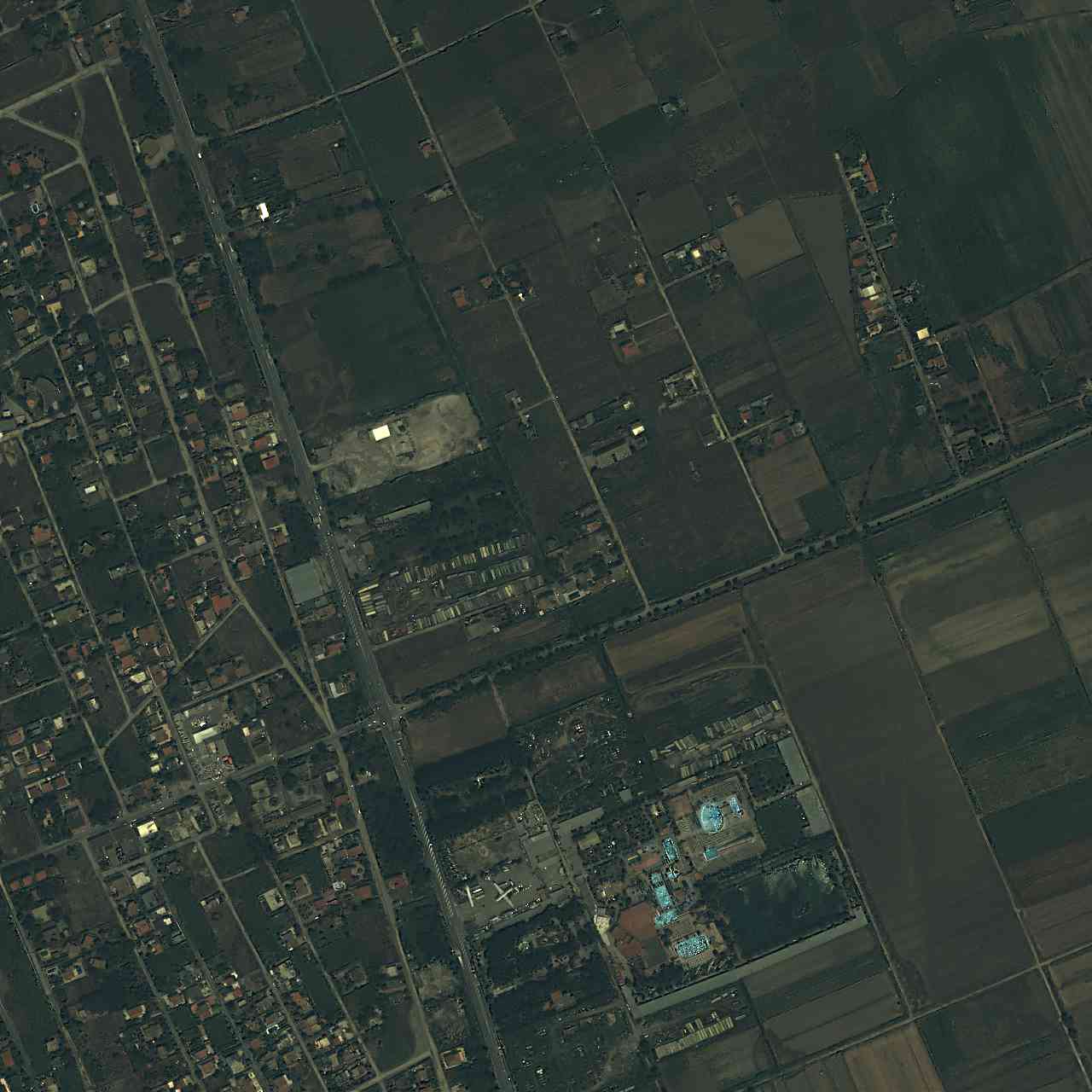}};
					\end{annotatedFigure}\vspace{0.5mm}
					\begin{annotatedFigure}
						{\adjincludegraphics[width=\linewidth,  trim={{.15\width} {.76\width} {.75\width} {.18\width}},clip]	{fig/imgIK/imgIK_full_cbd.jpg}};
					\end{annotatedFigure}\vspace{0.5mm}
				\end{minipage}\label{fig:ik:e}}
		\end{minipage}\\
		\begin{minipage}{1\linewidth}
			\subfloat[GLP-Reg-FS]{\begin{minipage}[t]{0.19\linewidth}
					\begin{annotatedFigure}
						{\adjincludegraphics[width=\linewidth,   trim={{.4\width} {.108\width} {.505\width} {.805\width}},clip]	{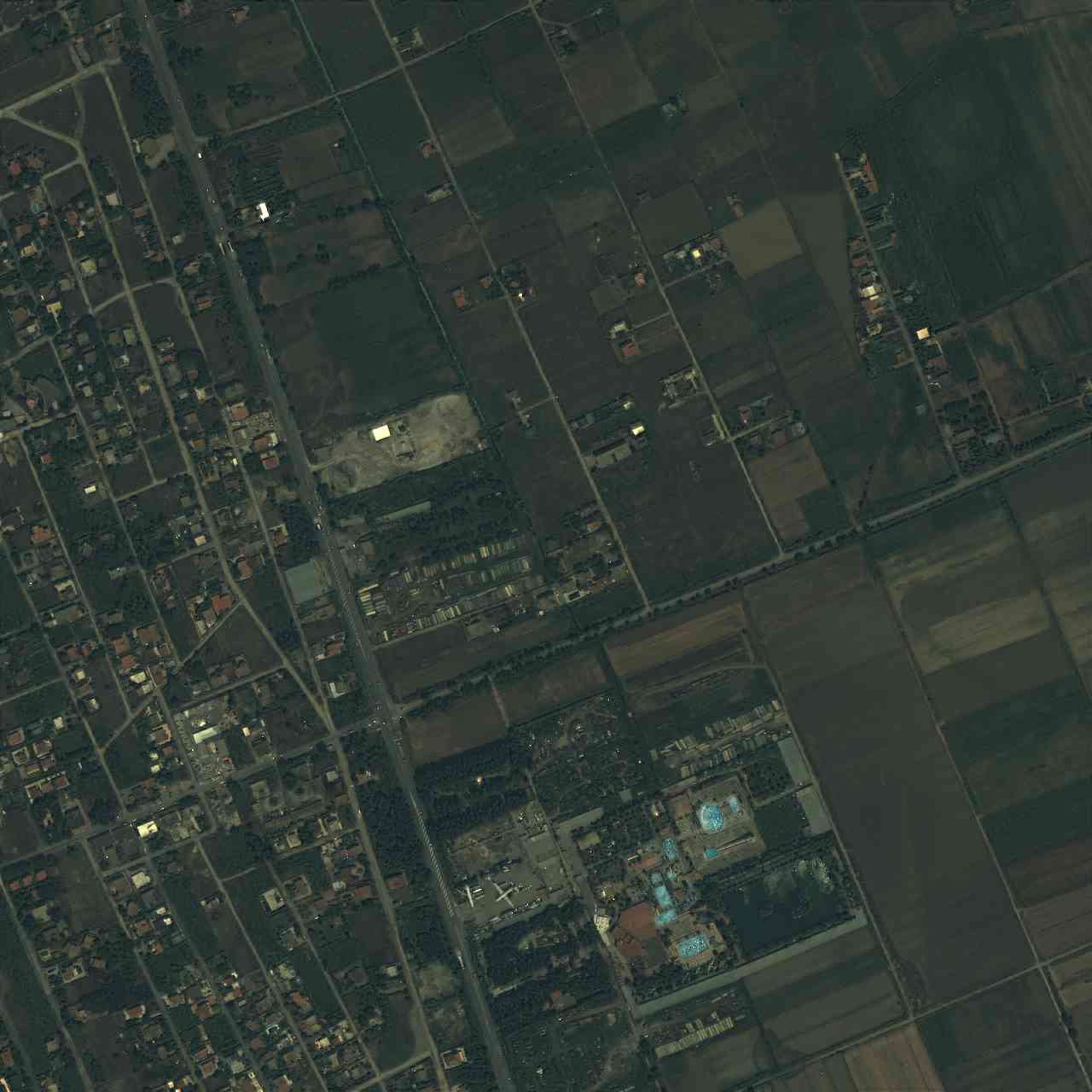}};
					\end{annotatedFigure}\vspace{0.5mm}
					\begin{annotatedFigure}
						{\adjincludegraphics[width=\linewidth,   trim={{.15\width} {.76\width} {.75\width} {.18\width}},clip]	{fig/imgIK/imgIK_full_glp_reg_fs.jpg}};
					\end{annotatedFigure}\vspace{0.5mm}
				\end{minipage}\label{fig:ik:f}}\hspace{0.001mm}
			\subfloat[GSA-Segm]{\begin{minipage}[t]{0.19\linewidth}
					\begin{annotatedFigure}
						{\adjincludegraphics[width=\linewidth,  trim={{.4\width} {.108\width} {.505\width} {.805\width}},clip]	{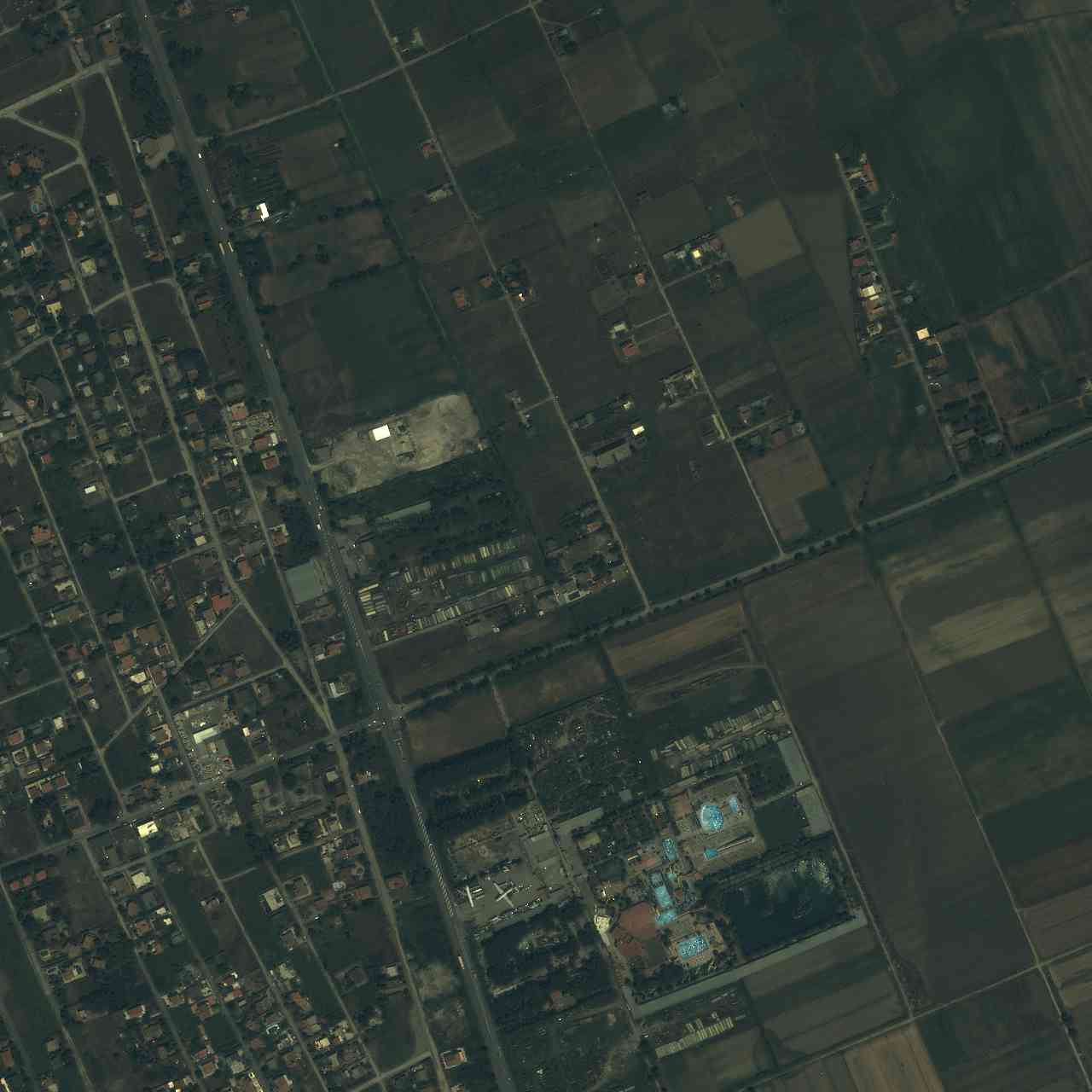}};
					\end{annotatedFigure}\vspace{0.5mm}
					\begin{annotatedFigure}
						{\adjincludegraphics[width=\linewidth,   trim={{.15\width} {.76\width} {.75\width} {.18\width}},clip]	{fig/imgIK/imgIK_full_seg_gsa.jpg}};
					\end{annotatedFigure}\vspace{0.5mm}
				\end{minipage}\label{fig:ik:g}}\hspace{0.001mm}
			\subfloat[GLP-Segm]{\begin{minipage}[t]{0.19\linewidth}
					\begin{annotatedFigure}
						{\adjincludegraphics[width=\linewidth,   trim={{.4\width} {.108\width} {.505\width} {.805\width}},clip]	{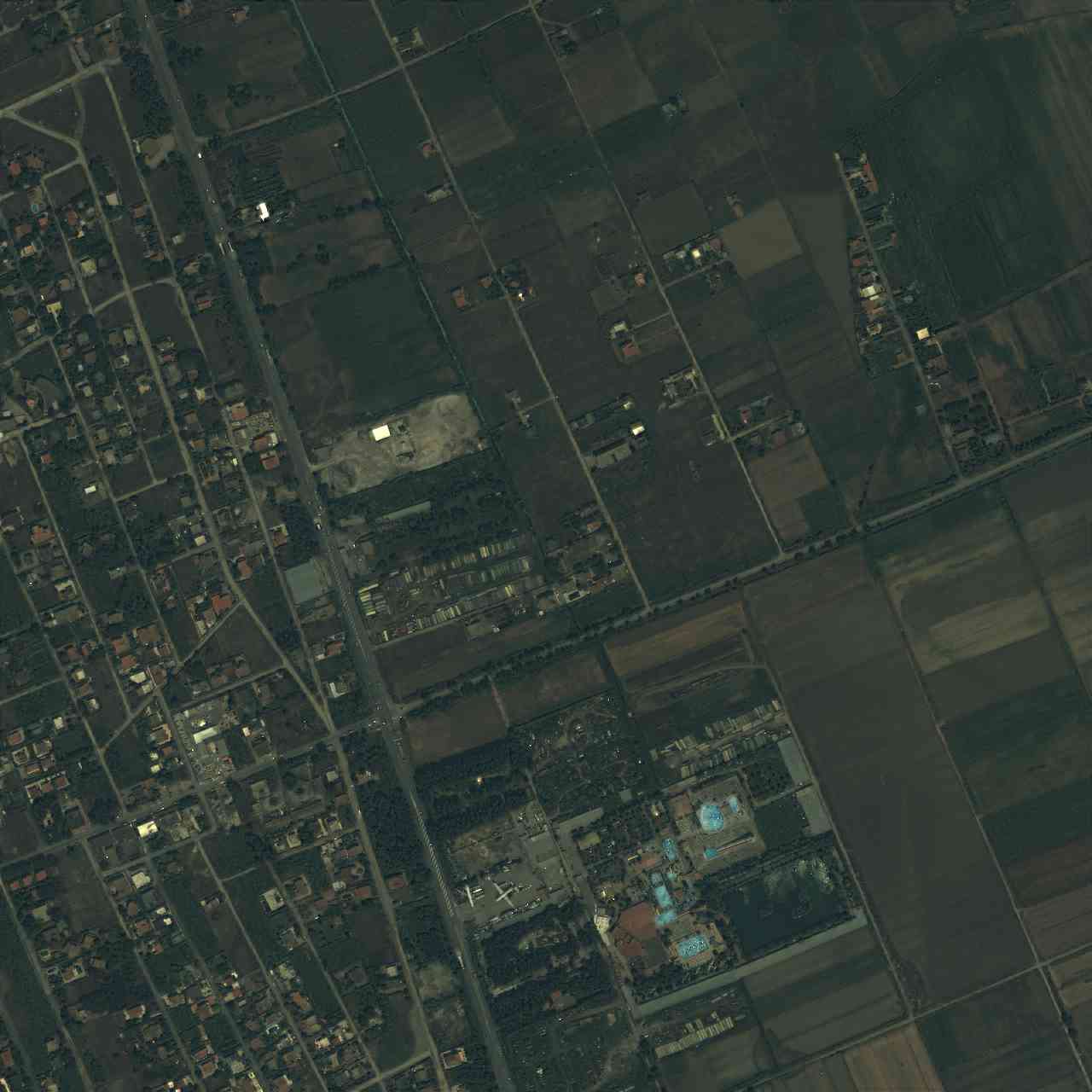}};
					\end{annotatedFigure}\vspace{0.5mm}
					\begin{annotatedFigure}
						{\adjincludegraphics[width=\linewidth,   trim={{.15\width} {.76\width} {.75\width} {.18\width}},clip]	{fig/imgIK/imgIK_full_seg_glp.jpg}};
					\end{annotatedFigure}\vspace{0.5mm}
				\end{minipage}\label{fig:ik:h}}\hspace{0.001mm}
			\subfloat[\hspace{-0.5mm}Target-CNN]{\begin{minipage}[t]{0.19\linewidth}
					\begin{annotatedFigure}
						{\adjincludegraphics[width=\linewidth,  trim={{.4\width} {.108\width} {.505\width} {.805\width}},clip]	{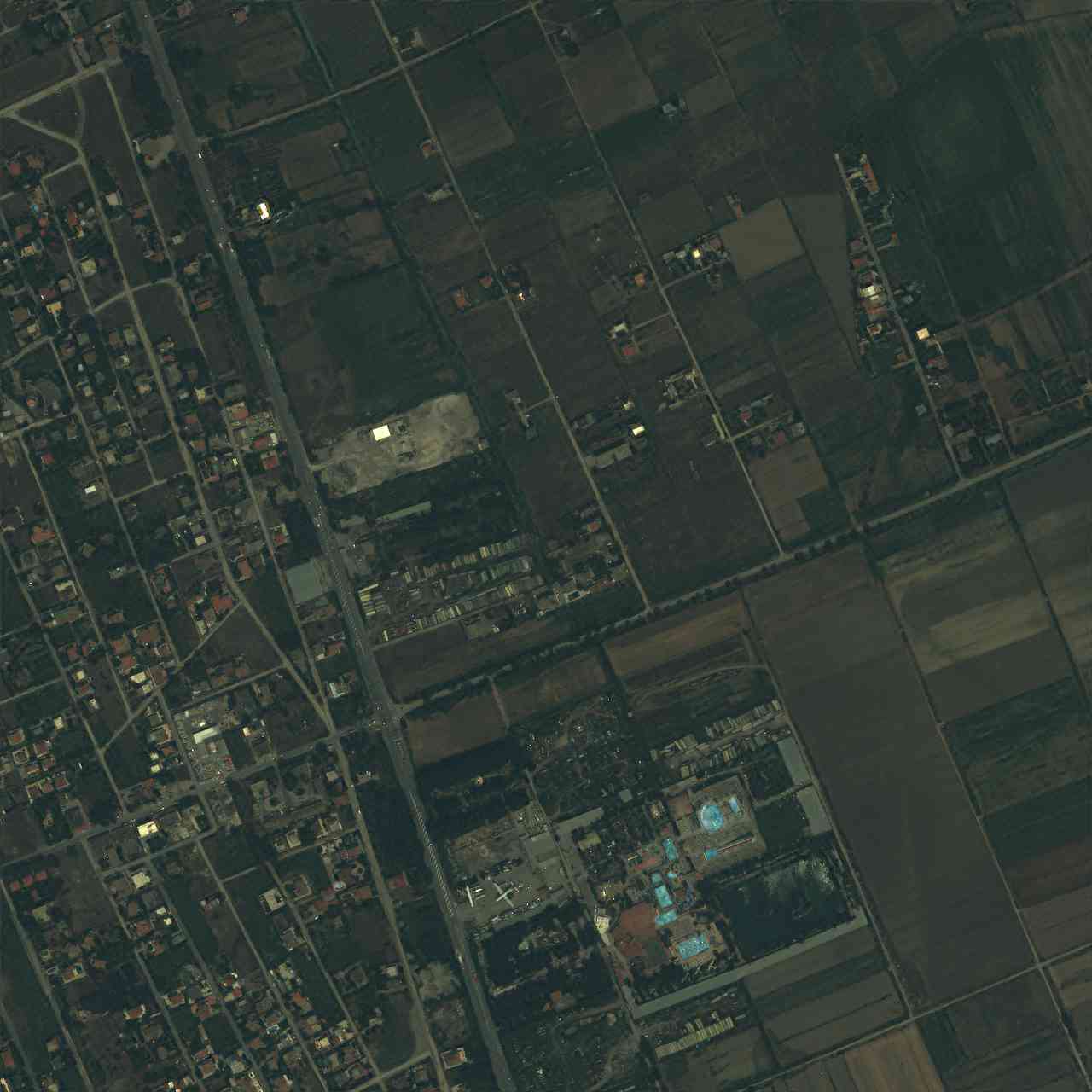}};
					\end{annotatedFigure}\vspace{0.5mm}
					\begin{annotatedFigure}
						{\adjincludegraphics[width=\linewidth,  trim={{.15\width} {.76\width} {.75\width} {.18\width}},clip]	{fig/imgIK/imgIK_full_pancnn.jpg}};
					\end{annotatedFigure}\vspace{0.5mm}
				\end{minipage}\label{fig:ik:i}}\hspace{0.001mm}
			\subfloat[Ours]{\begin{minipage}[t]{0.19\linewidth}
					\begin{annotatedFigure}
						{\adjincludegraphics[width=\linewidth,   trim={{.4\width} {.108\width} {.505\width} {.805\width}},clip]	{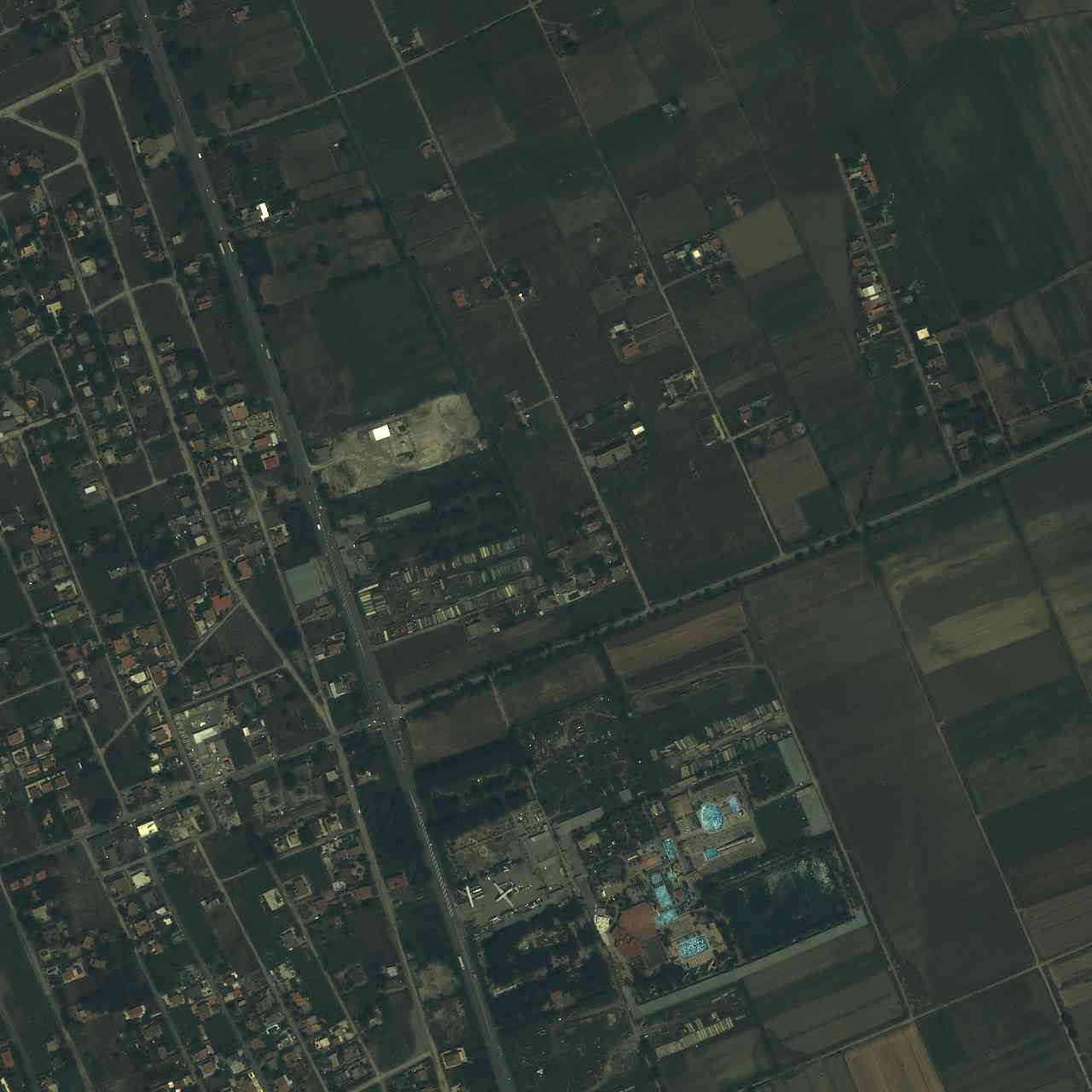}};
					\end{annotatedFigure}\vspace{0.5mm}
					\begin{annotatedFigure}
						{\adjincludegraphics[width=\linewidth,   trim={{.15\width} {.76\width} {.75\width} {.18\width}},clip]	{fig/imgIK/imgIK_full_ours.jpg}};
					\end{annotatedFigure}\vspace{0.5mm}
				\end{minipage}\label{fig:ik:j}}
		\end{minipage}
		\caption{RGB channels of the pansharpened results from different methods on IKONOS at full resolution.}
		\label{fig:ik}
\end{minipage}
\end{figure*}

\begin{figure*}[htbp]
	\centering
	\begin{minipage}{0.95\linewidth}
		\begin{minipage}{1\linewidth}
			\subfloat[LR MSI]{\begin{minipage}[t]{0.19\linewidth}
					\begin{annotatedFigure}
						{\adjincludegraphics[width=\linewidth,  trim={{.455\width} {.44\width} {.45\width} {.475\width}},clip] {fig/imgWV2/imgWV2_rd_msi.jpg}};
					\end{annotatedFigure}\vspace{0.5mm}
					\begin{annotatedFigure}
						{\adjincludegraphics[width=\linewidth,  trim={{.75\width} {.685\width} {.18\width} {.265\width}},clip] {fig/imgWV2/imgWV2_rd_msi.jpg}};
					\end{annotatedFigure}\vspace{1mm}
				\end{minipage}\label{fig:wv2:a}}\hspace{0.001mm}
			\subfloat[GSA]{\begin{minipage}[t]{0.19\linewidth}
					\begin{annotatedFigure}
						{\adjincludegraphics[width=\linewidth, trim={{.455\width} {.44\width} {.45\width} {.475\width}},clip]	{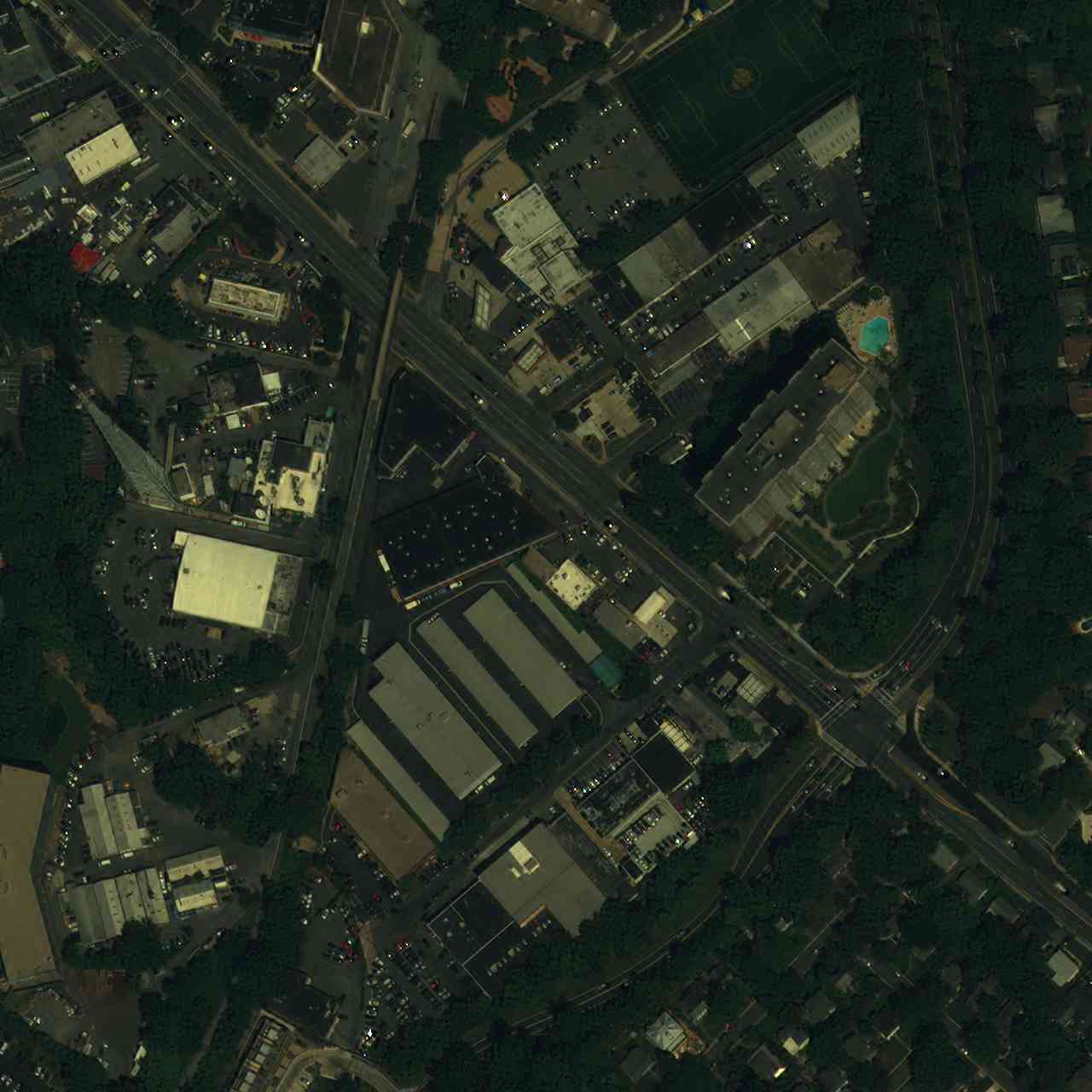}};
					\end{annotatedFigure}\vspace{0.5mm}
					\begin{annotatedFigure}
						{\adjincludegraphics[width=\linewidth,  trim={{.75\width} {.685\width} {.18\width} {.265\width}},clip]	{fig/imgWV2/imgWV2_full_gsa.jpg}};
					\end{annotatedFigure}\vspace{0.5mm}
				\end{minipage}\label{fig:wv2:b}}\hspace{0.001mm}
			\subfloat[PRACS]{\begin{minipage}[t]{0.19\linewidth}
					\begin{annotatedFigure}
						{\adjincludegraphics[width=\linewidth,  trim={{.455\width} {.44\width} {.45\width} {.475\width}},clip]		{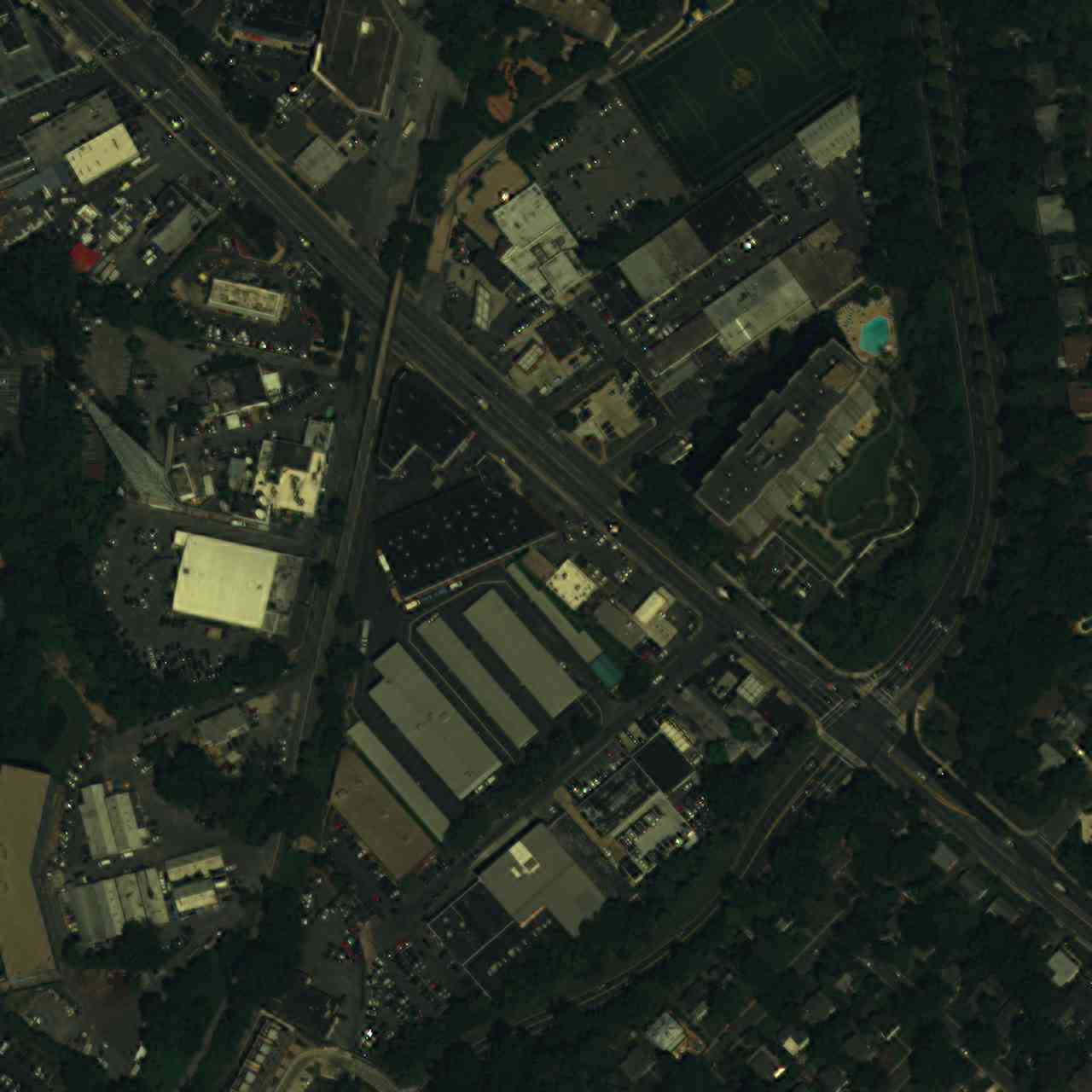}};
					\end{annotatedFigure}\vspace{0.5mm}
					\begin{annotatedFigure}
						{\adjincludegraphics[width=\linewidth,   trim={{.75\width} {.685\width} {.18\width} {.265\width}},clip]	{fig/imgWV2/imgWV2_full_pracs.jpg}};
					\end{annotatedFigure}\vspace{0.5mm}
				\end{minipage}\label{fig:wv2:c}}\hspace{0.001mm}
			\subfloat[BDSD-PC]{\begin{minipage}[t]{0.19\linewidth}
					\begin{annotatedFigure}
						{\adjincludegraphics[width=\linewidth,  trim={{.455\width} {.44\width} {.45\width} {.475\width}},clip] {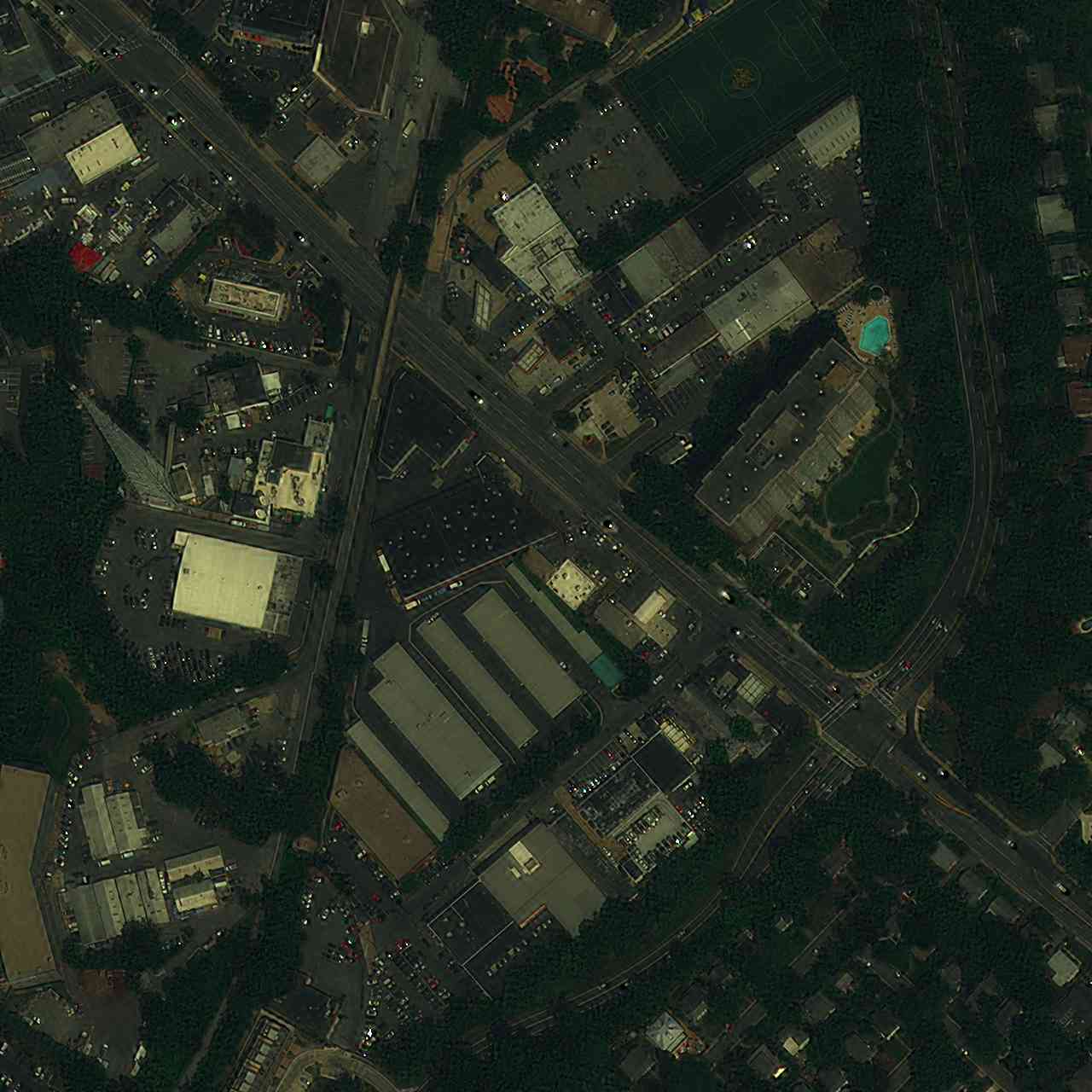}};
					\end{annotatedFigure}\vspace{0.5mm}
					\begin{annotatedFigure}
						{\adjincludegraphics[width=\linewidth,   trim={{.75\width} {.685\width} {.18\width} {.265\width}},clip] {fig/imgWV2/imgWV2_full_bdsd_pc.jpg}};
					\end{annotatedFigure}\vspace{0.5mm}
				\end{minipage}\label{fig:wv2:d}}\hspace{0.001mm}
			\subfloat[MTF-CBD]{\begin{minipage}[t]{0.19\linewidth}
					\begin{annotatedFigure}
						{\adjincludegraphics[width=\linewidth,   trim={{.455\width} {.44\width} {.45\width} {.475\width}},clip]	{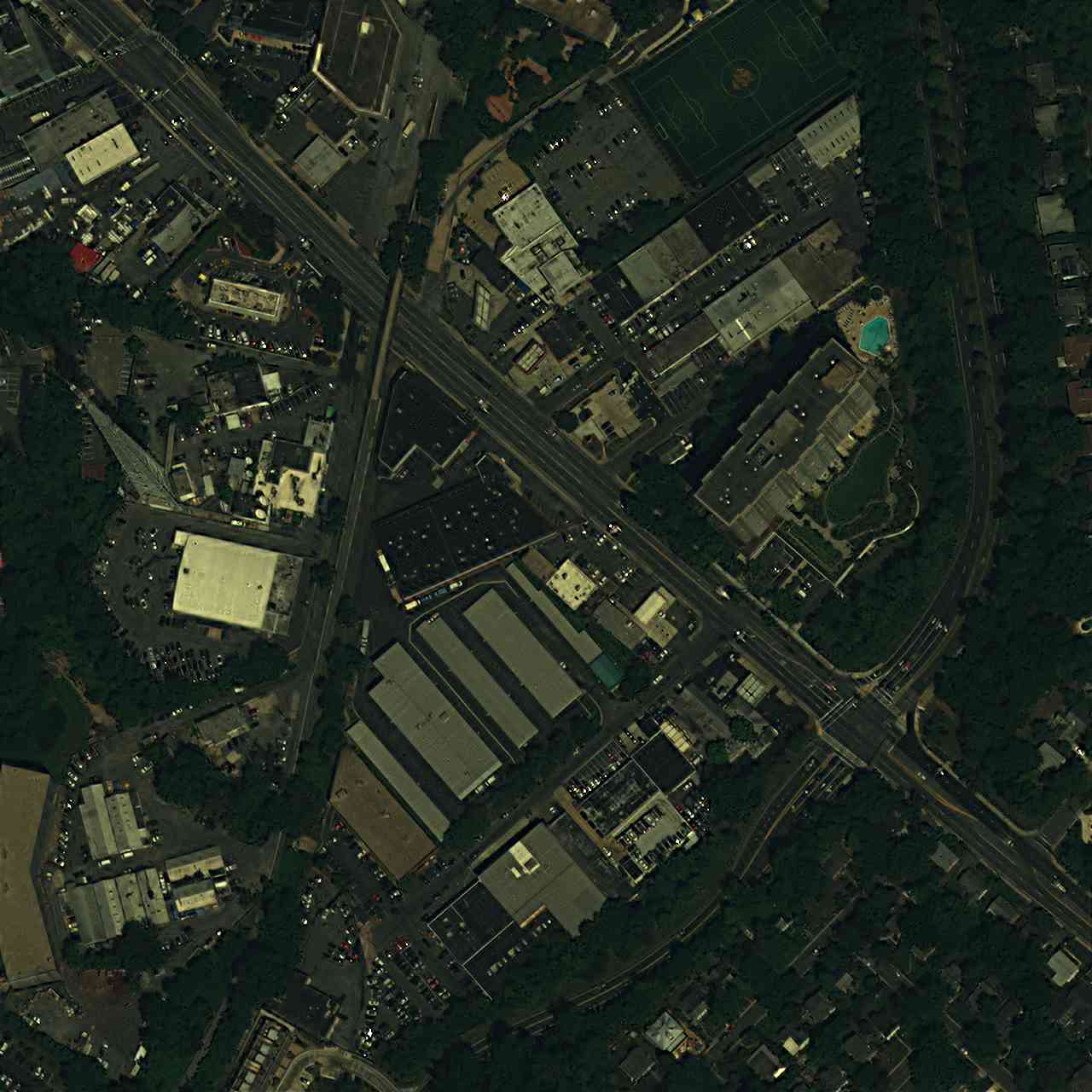}};
					\end{annotatedFigure}\vspace{0.5mm}
					\begin{annotatedFigure}
						{\adjincludegraphics[width=\linewidth,   trim={{.75\width} {.685\width} {.18\width} {.265\width}},clip] {fig/imgWV2/imgWV2_full_cbd.jpg}};
					\end{annotatedFigure}\vspace{0.5mm}
				\end{minipage}\label{fig:wv2:e}}
		\end{minipage}
		\begin{minipage}{1\linewidth}
			\subfloat[GLP-Reg-FS]{\begin{minipage}[t]{0.19\linewidth}
					\begin{annotatedFigure}
						{\adjincludegraphics[width=\linewidth,   trim={{.455\width} {.44\width} {.45\width} {.475\width}},clip] {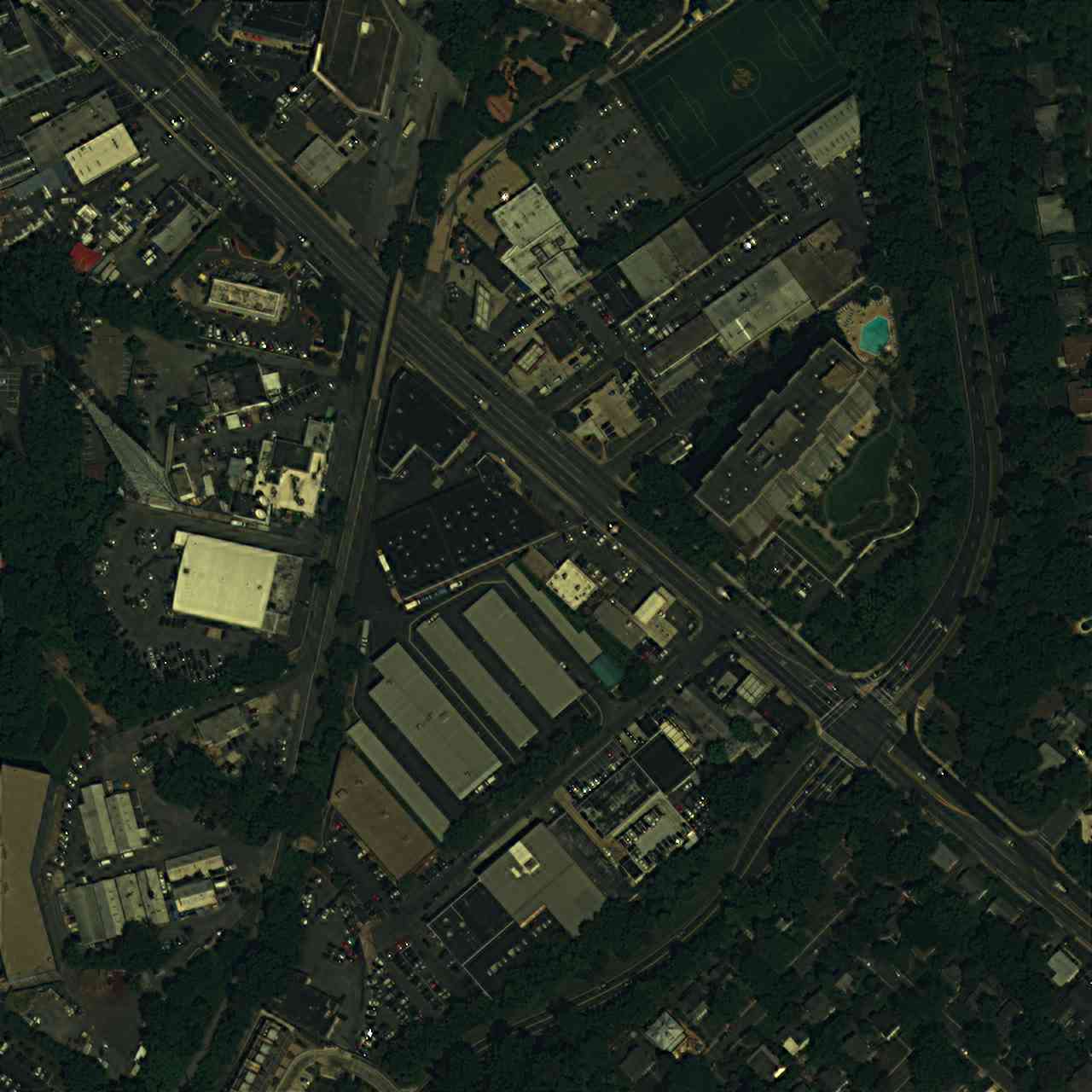}};
					\end{annotatedFigure}\vspace{0.5mm}
					\begin{annotatedFigure}
						{\adjincludegraphics[width=\linewidth,    trim={{.75\width} {.685\width} {.18\width} {.265\width}},clip]	{fig/imgWV2/imgWV2_full_glp_reg_fs.jpg}};
					\end{annotatedFigure}\vspace{0.5mm}
				\end{minipage}\label{fig:wv2:f}}\hspace{0.001mm}
			\subfloat[GSA-Segm]{\begin{minipage}[t]{0.19\linewidth}
					\begin{annotatedFigure}
						{\adjincludegraphics[width=\linewidth, 	trim={{.455\width} {.44\width} {.45\width} {.475\width}},clip] {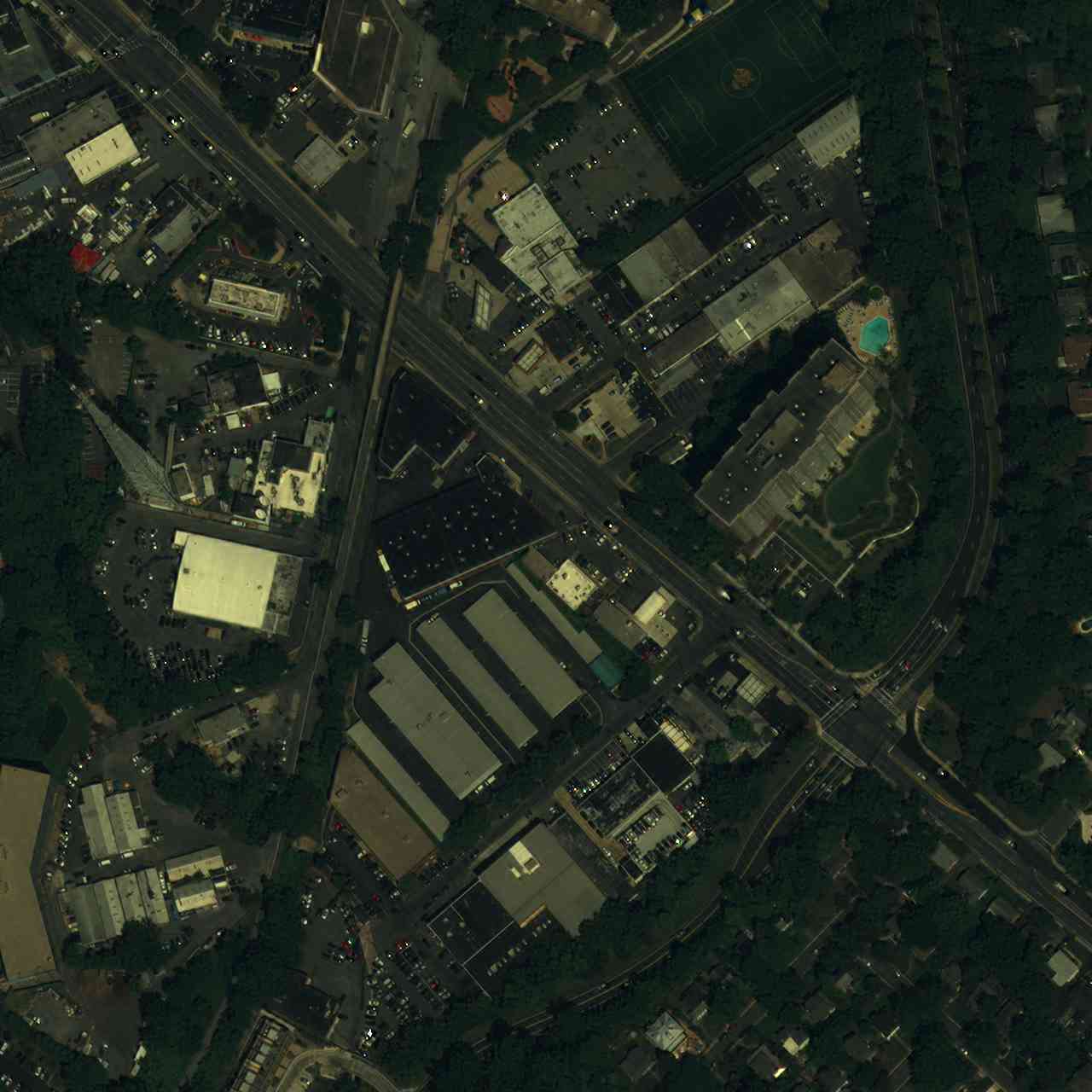}};
					\end{annotatedFigure}\vspace{0.5mm}
					\begin{annotatedFigure}
						{\adjincludegraphics[width=\linewidth,   trim={{.75\width} {.685\width} {.18\width} {.265\width}},clip]	{fig/imgWV2/imgWV2_full_seg_gsa.jpg}};
					\end{annotatedFigure}\vspace{0.5mm}
				\end{minipage}\label{fig:wv2:g}}\hspace{0.001mm}
			\subfloat[GLP-Segm]{\begin{minipage}[t]{0.19\linewidth}
					\begin{annotatedFigure}
						{\adjincludegraphics[width=\linewidth,  trim={{.455\width} {.44\width} {.45\width} {.475\width}},clip]	{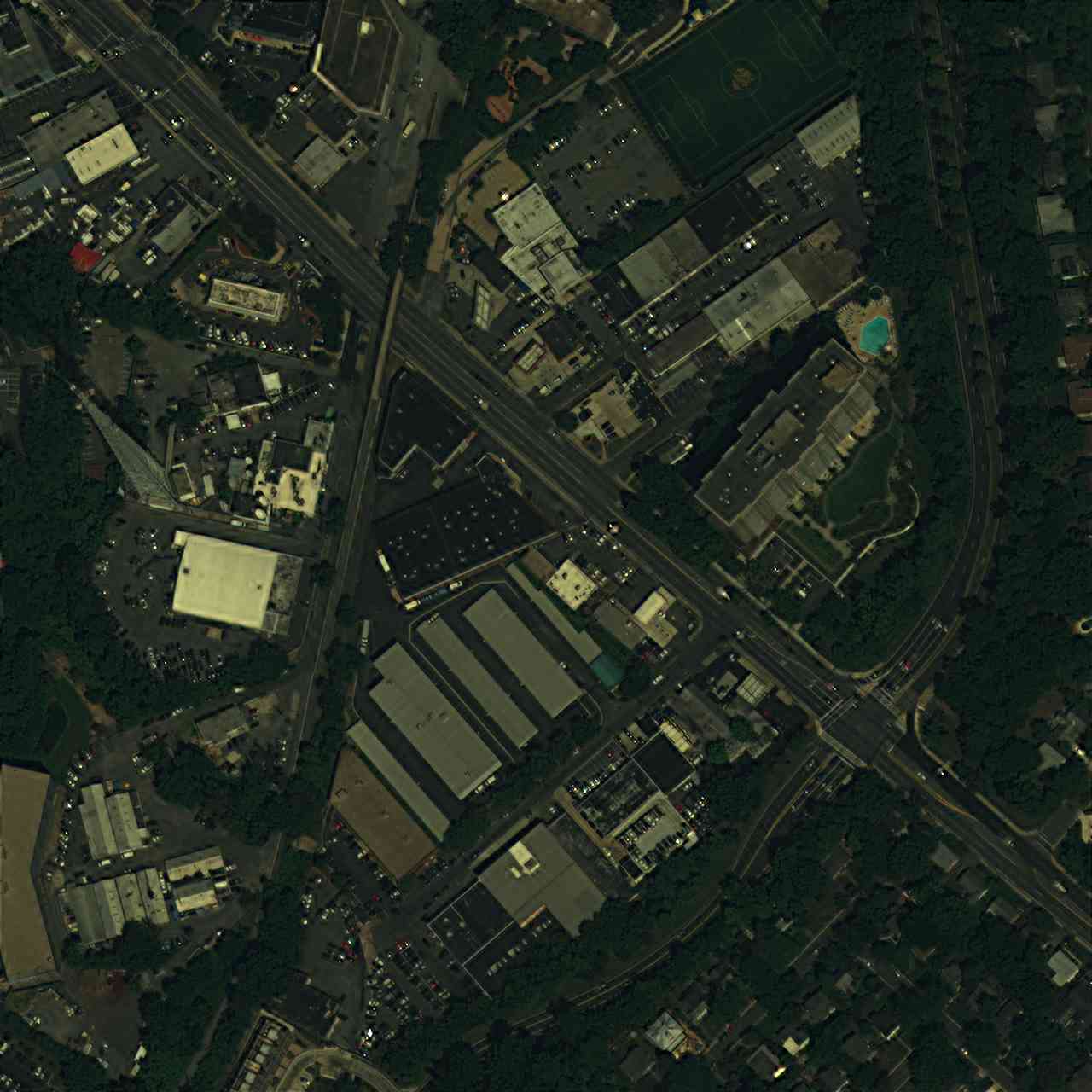}};
					\end{annotatedFigure}\vspace{0.5mm}
					\begin{annotatedFigure}
						{\adjincludegraphics[width=\linewidth,   trim={{.75\width} {.685\width} {.18\width} {.265\width}},clip]	{fig/imgWV2/imgWV2_full_seg_glp.jpg}};
					\end{annotatedFigure}\vspace{0.5mm}
				\end{minipage}\label{fig:wv2:h}}\hspace{0.001mm}
			\subfloat[Target-CNN]{\begin{minipage}[t]{0.19\linewidth}
					\begin{annotatedFigure}
						{\adjincludegraphics[width=\linewidth,  trim={{.455\width} {.44\width} {.45\width} {.475\width}},clip]	{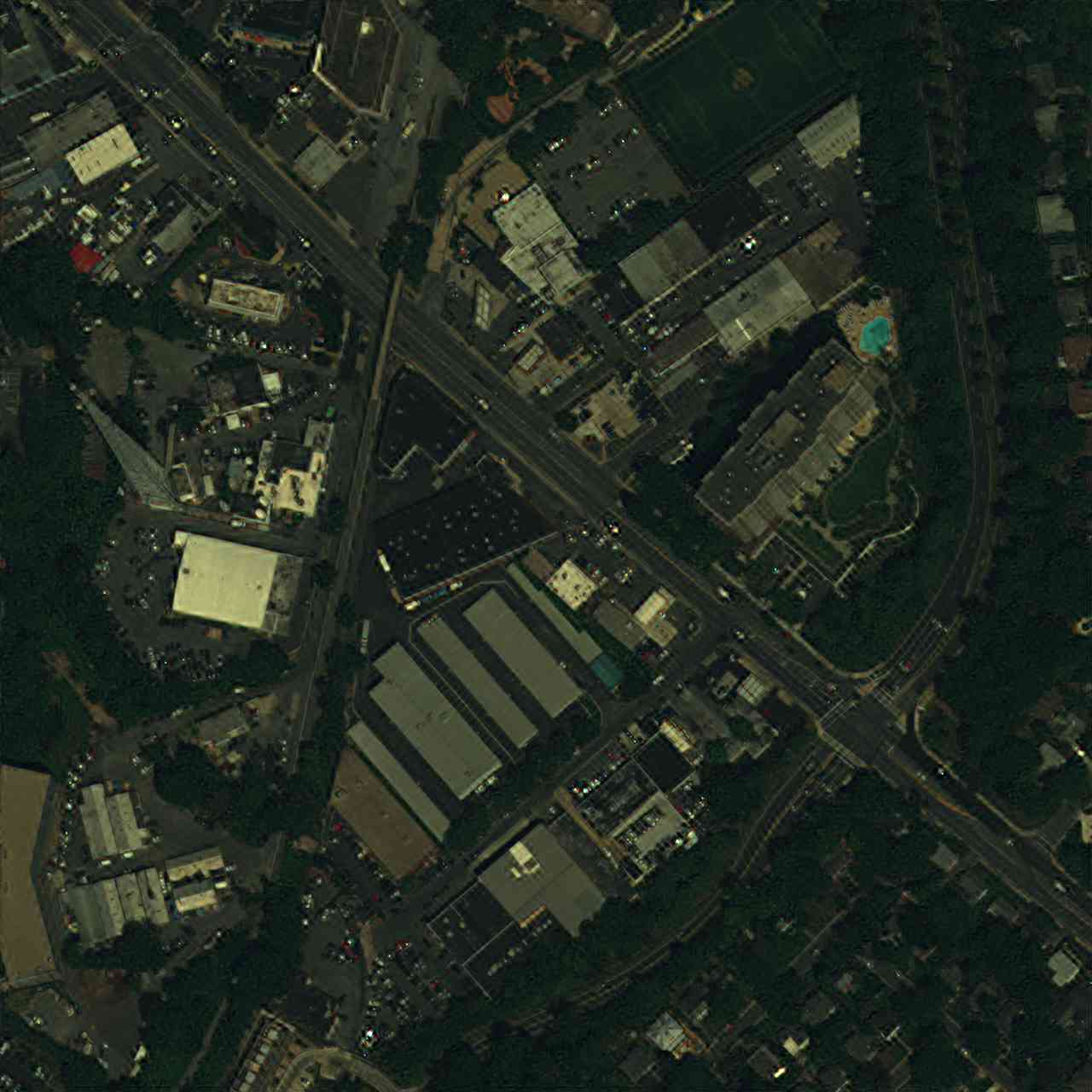}};
					\end{annotatedFigure}\vspace{0.5mm}
					\begin{annotatedFigure}
						{\adjincludegraphics[width=\linewidth,   trim={{.75\width} {.685\width} {.18\width} {.265\width}},clip]	{fig/imgWV2/imgWV2_full_pancnn.jpg}};
					\end{annotatedFigure}\vspace{0.5mm}
				\end{minipage}\label{fig:wv2:i}}\hspace{0.001mm}
			\subfloat[Ours]{\begin{minipage}[t]{0.19\linewidth}
					\begin{annotatedFigure}
						{\adjincludegraphics[width=\linewidth,   trim={{.455\width} {.44\width} {.45\width} {.475\width}},clip]	{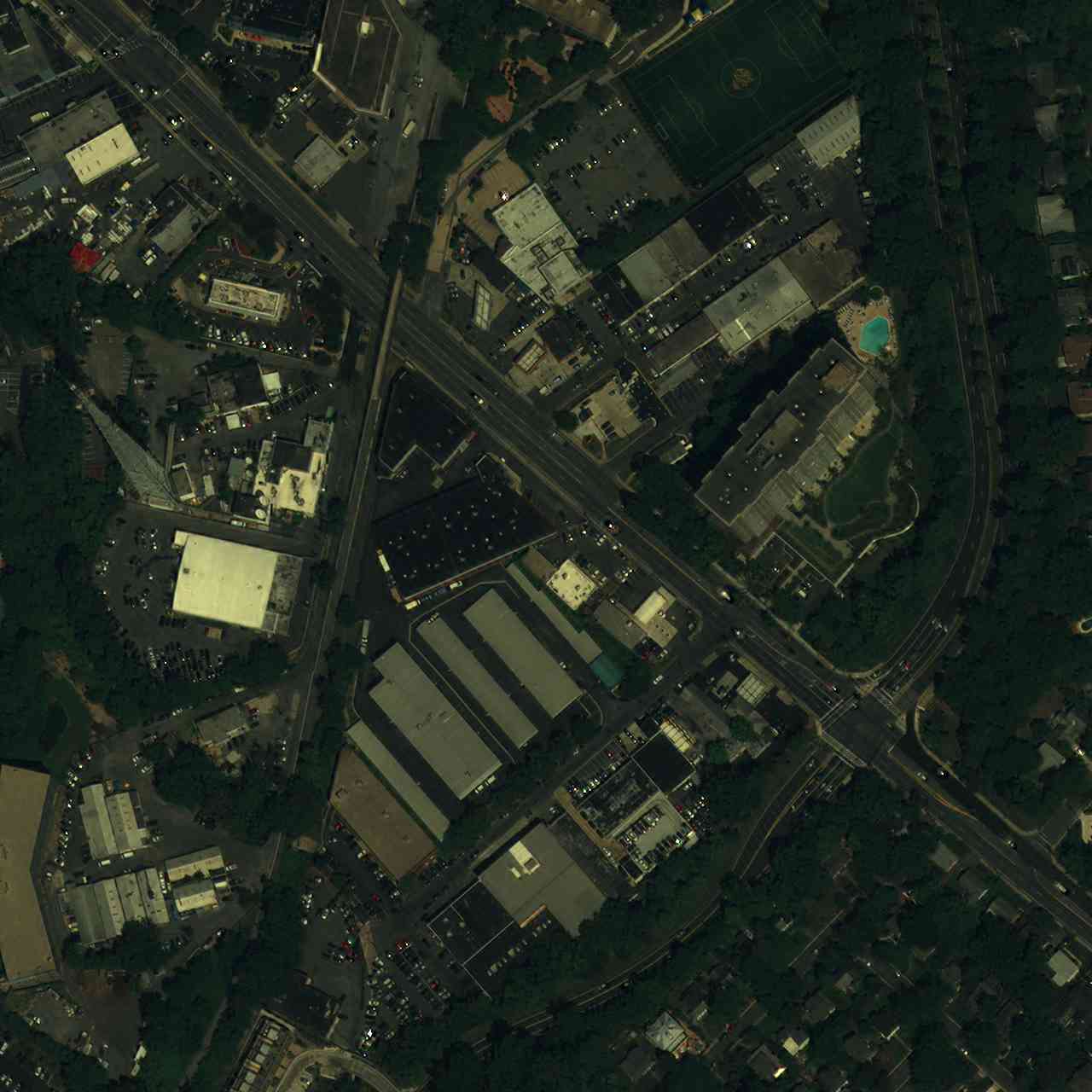}};
					\end{annotatedFigure}\vspace{0.5mm}
					\begin{annotatedFigure}
						{\adjincludegraphics[width=\linewidth,   trim={{.75\width} {.685\width} {.18\width} {.265\width}},clip]	{fig/imgWV2/imgWV2_full_ours2.jpg}};
					\end{annotatedFigure}\vspace{0.5mm}
				\end{minipage}\label{fig:wv2:j}}
		\end{minipage}
		\caption{RGB channels of the pansharpened results from different methods on WV2 at full resolution.}
		\label{fig:wv2}
	\end{minipage}
	\begin{minipage}{0.95\linewidth}
		\begin{minipage}{1\linewidth}
			\subfloat[\hspace{-0.5mm}LR MSI]{\begin{minipage}[t]{0.19\linewidth}
					\begin{annotatedFigure}
						{\adjincludegraphics[width=\linewidth,  trim={{.18\width} {.43\width} {.715\width} {.485\width}},clip]	{fig/imgWV3/imgWV3_rd_msi.jpg}};
					\end{annotatedFigure}\vspace{0.5mm}
					\begin{annotatedFigure}
						{\adjincludegraphics[width=\linewidth,  trim={{.25\width} {.12\width} {.615\width} {.752\width}},clip]	{fig/imgWV3/imgWV3_rd_msi.jpg}};
					\end{annotatedFigure}\vspace{1mm}
				\end{minipage}\label{fig:wv3:a}}\hspace{0.001mm}
			\subfloat[GSA]{\begin{minipage}[t]{0.19\linewidth}
					\begin{annotatedFigure}
						{\adjincludegraphics[width=\linewidth,   trim={{.18\width} {.43\width} {.715\width} {.485\width}},clip]	{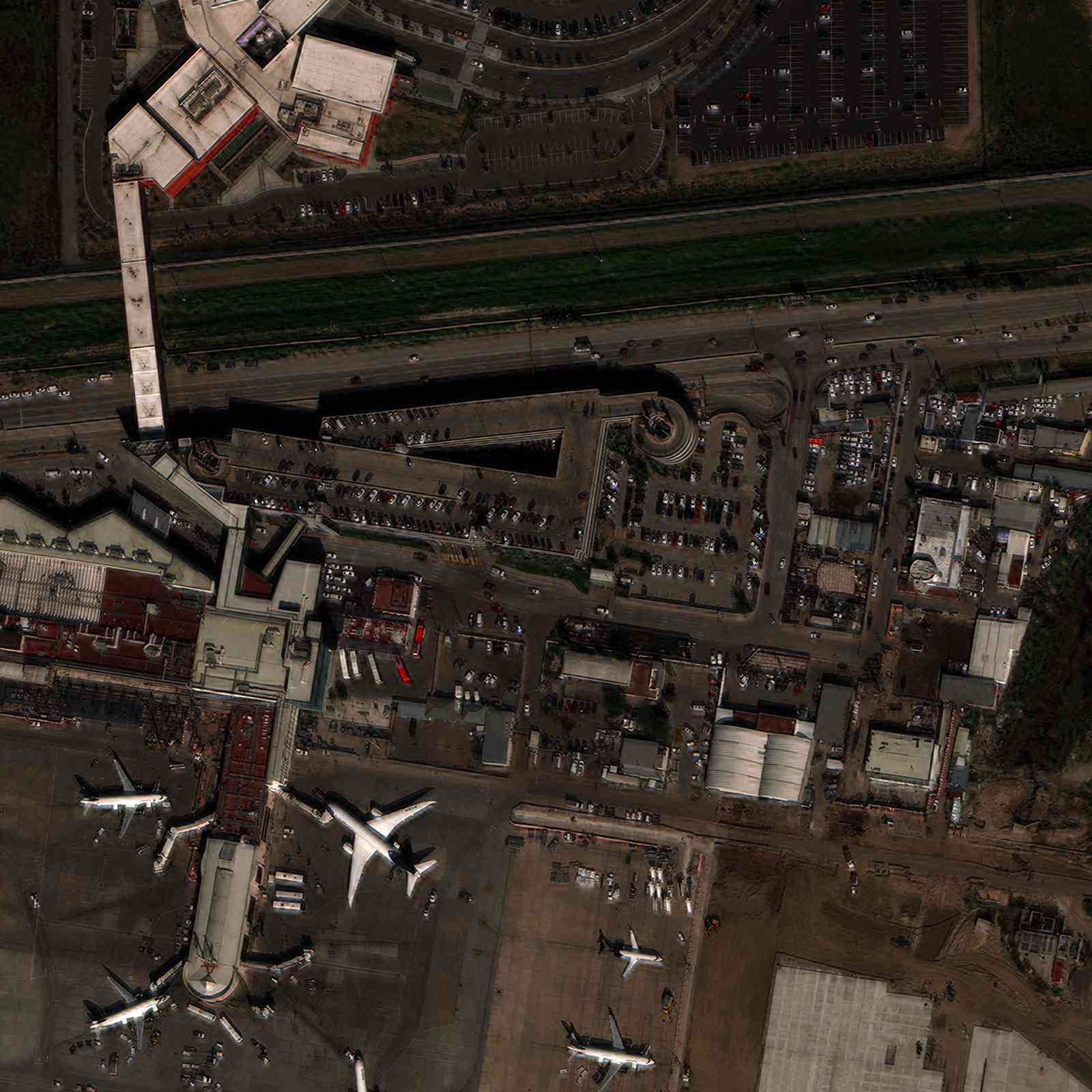}};
					\end{annotatedFigure}\vspace{0.5mm}
					\begin{annotatedFigure}
						{\adjincludegraphics[width=\linewidth,  trim={{.25\width} {.12\width} {.615\width} {.752\width}},clip] {fig/imgWV3/imgWV3_full_gsa.jpg}};
					\end{annotatedFigure}\vspace{0.5mm}
				\end{minipage}\label{fig:wv3:b}}\hspace{0.001mm}
			\subfloat[PRACS]{\begin{minipage}[t]{0.19\linewidth}
					\begin{annotatedFigure}
						{\adjincludegraphics[width=\linewidth,   trim={{.18\width} {.43\width} {.715\width} {.485\width}},clip]	{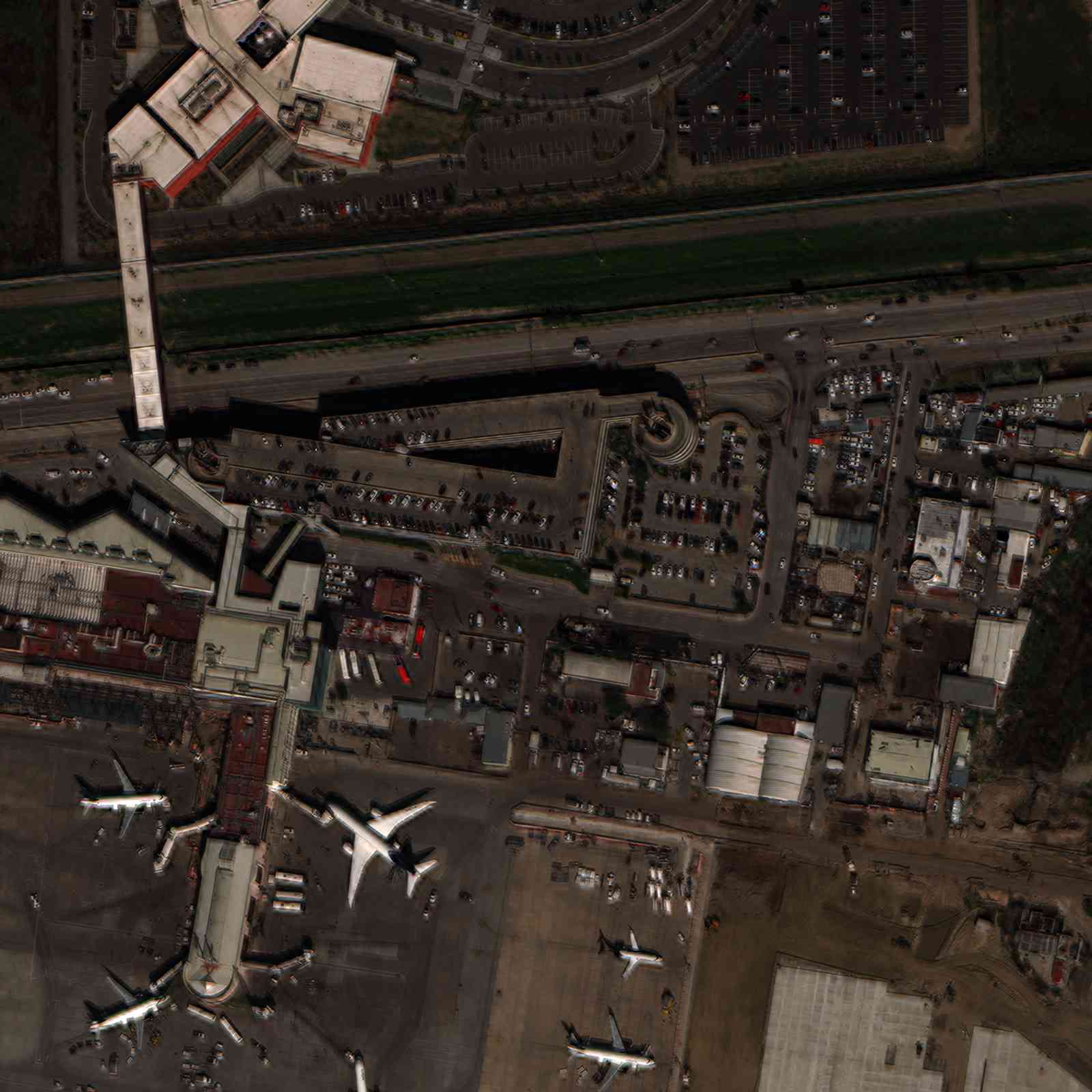}};
					\end{annotatedFigure}\vspace{0.5mm}
					\begin{annotatedFigure}
						{\adjincludegraphics[width=\linewidth,   trim={{.25\width} {.12\width} {.615\width} {.752\width}},clip]	{fig/imgWV3/imgWV3_full_pracs.jpg}};
					\end{annotatedFigure}\vspace{0.5mm}
				\end{minipage}\label{fig:wv3:c}}\hspace{0.001mm}
			\subfloat[BDSD-PC]{\begin{minipage}[t]{0.19\linewidth}
					\begin{annotatedFigure}
						{\adjincludegraphics[width=\linewidth,   trim={{.18\width} {.43\width} {.715\width} {.485\width}},clip] {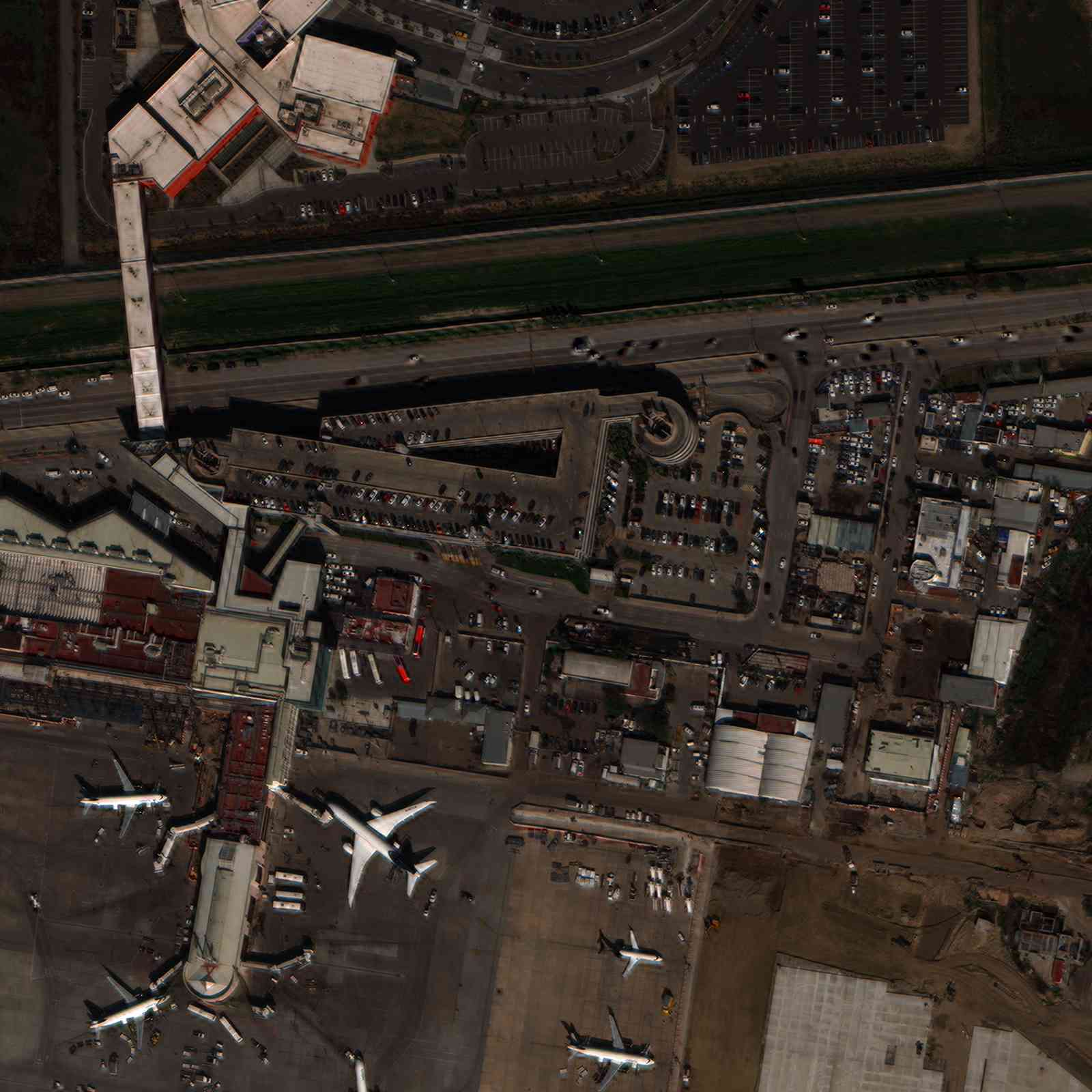}};
					\end{annotatedFigure}\vspace{0.5mm}
					\begin{annotatedFigure}
						{\adjincludegraphics[width=\linewidth,  trim={{.25\width} {.12\width} {.615\width} {.752\width}},clip]	{fig/imgWV3/imgWV3_full_bdsd_pc.jpg}};
					\end{annotatedFigure}\vspace{0.5mm}
				\end{minipage}\label{fig:wv3:d}}\hspace{0.001mm}
			\subfloat[MTF-CBD]{\begin{minipage}[t]{0.19\linewidth}
					\begin{annotatedFigure}
						{\adjincludegraphics[width=\linewidth,   trim={{.18\width} {.43\width} {.715\width} {.485\width}},clip]	{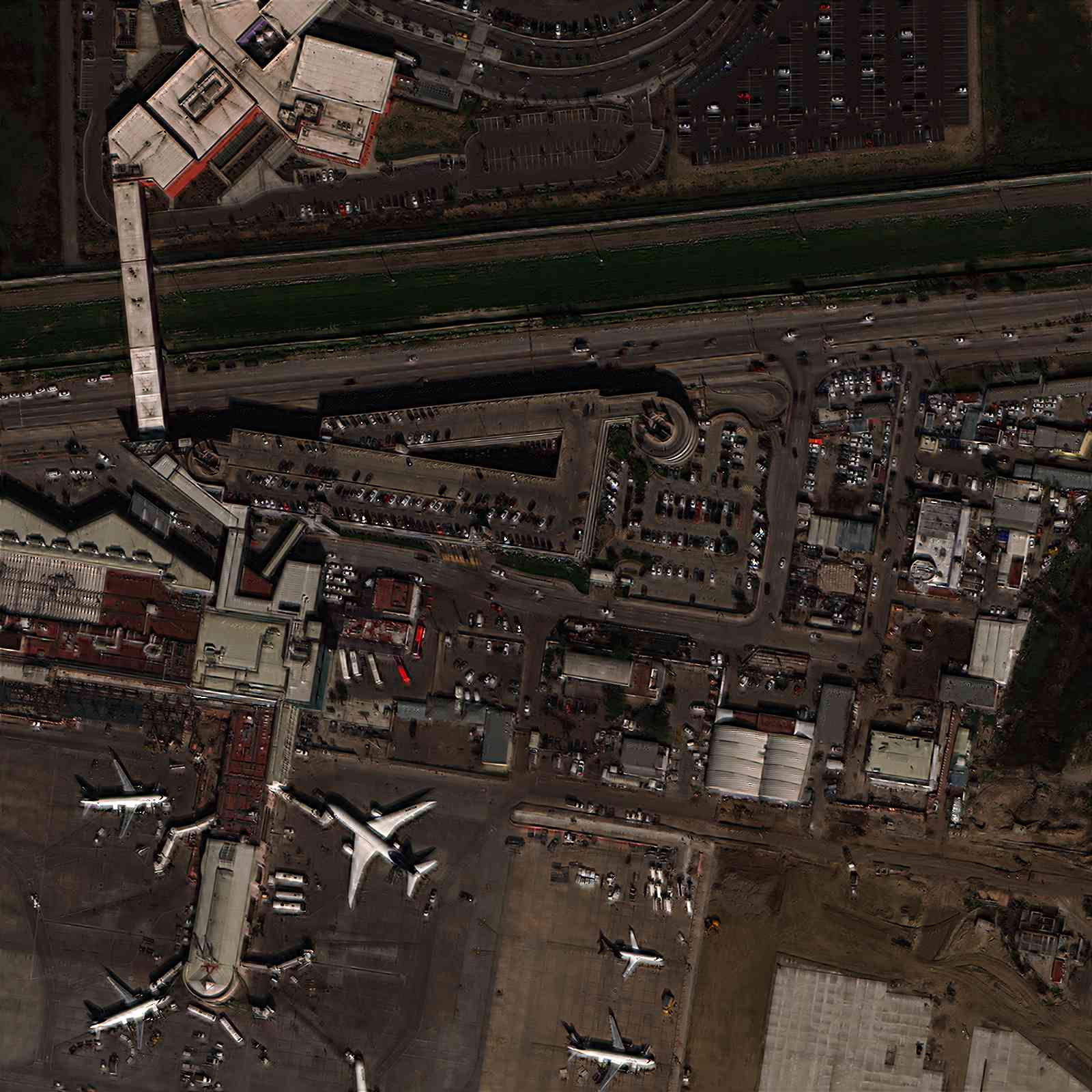}};
					\end{annotatedFigure}\vspace{0.5mm}
					\begin{annotatedFigure}
						{\adjincludegraphics[width=\linewidth,  trim={{.25\width} {.12\width} {.615\width} {.752\width}},clip]	{fig/imgWV3/imgWV3_full_cbd.jpg}};
					\end{annotatedFigure}\vspace{0.5mm}
				\end{minipage}\label{fig:wv3:e}}
		\end{minipage}\\
		\begin{minipage}{1\linewidth}
			\subfloat[GLP-Reg-FS]{\begin{minipage}[t]{0.19\linewidth}
					\begin{annotatedFigure}
						{\adjincludegraphics[width=\linewidth,   trim={{.18\width} {.43\width} {.715\width} {.485\width}},clip]	{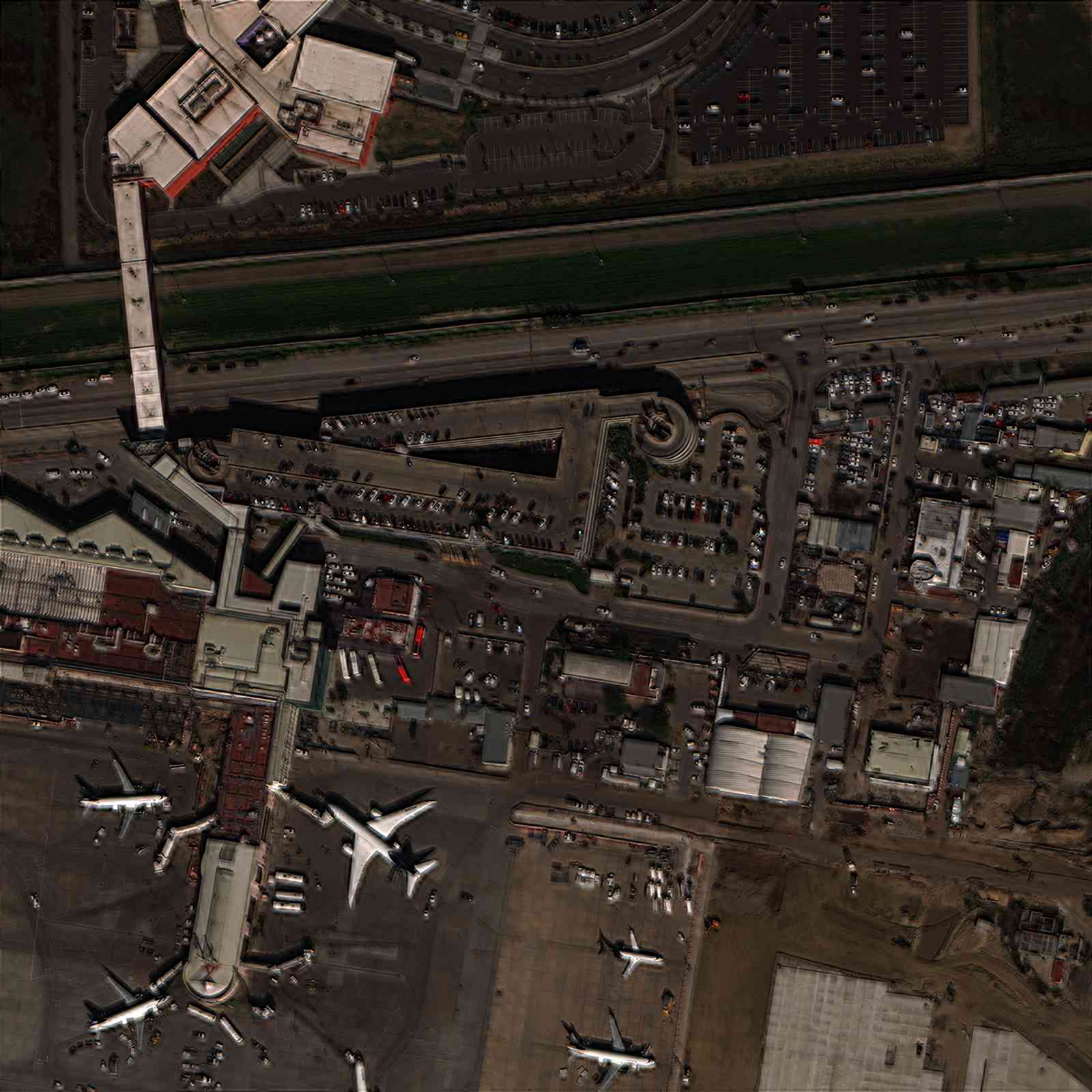}};
					\end{annotatedFigure}\vspace{0.5mm}
					\begin{annotatedFigure}
						{\adjincludegraphics[width=\linewidth,   trim={{.25\width} {.12\width} {.615\width} {.752\width}},clip]	{fig/imgWV3/imgWV3_full_glp_reg_fs.jpg}};
					\end{annotatedFigure}\vspace{0.5mm}
				\end{minipage}\label{fig:wv3:f}}\hspace{0.001mm}
			\subfloat[GSA-Segm]{\begin{minipage}[t]{0.19\linewidth}
					\begin{annotatedFigure}
						{\adjincludegraphics[width=\linewidth,   trim={{.18\width} {.43\width} {.715\width} {.485\width}},clip]	{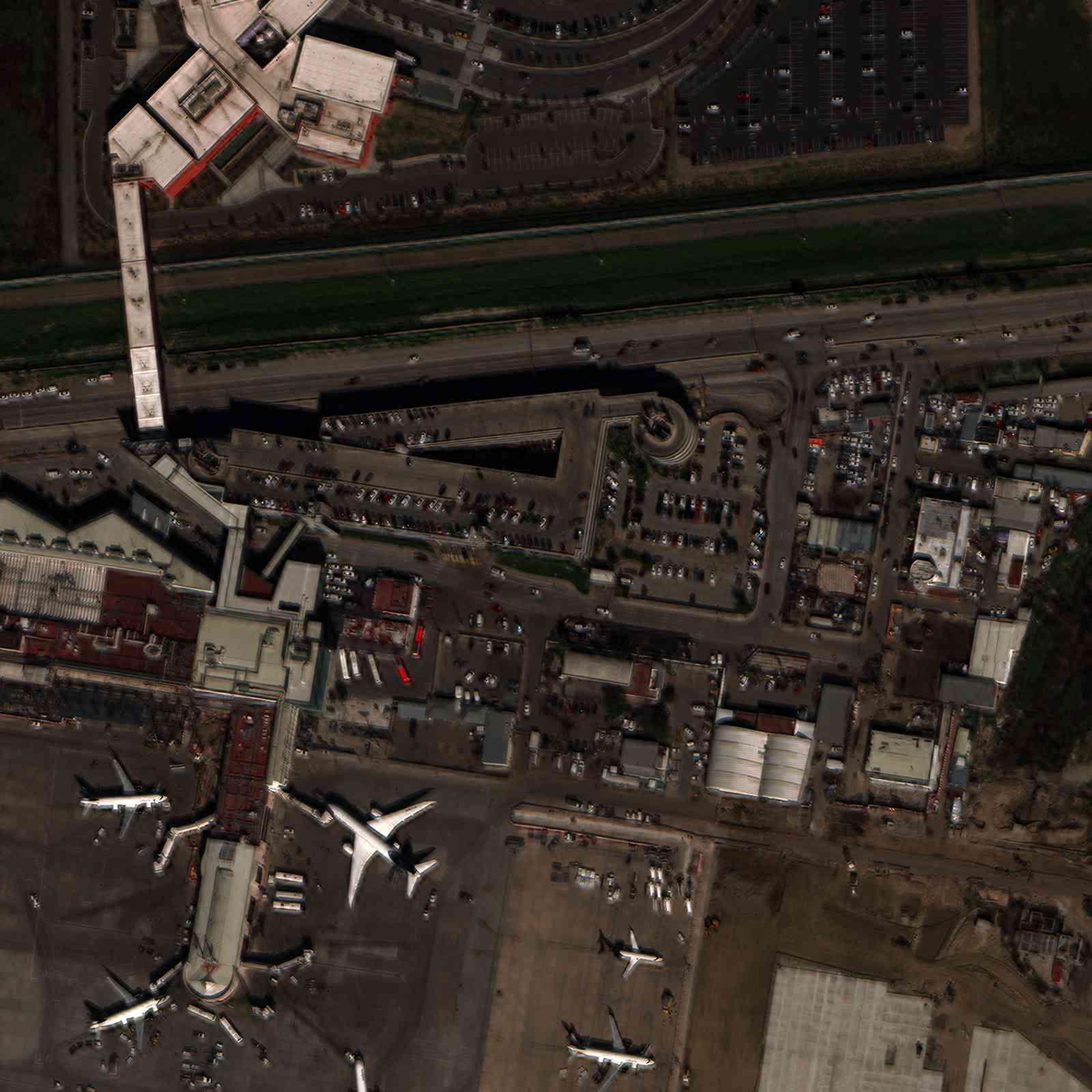}};
					\end{annotatedFigure}\vspace{0.5mm}
					\begin{annotatedFigure}
						{\adjincludegraphics[width=\linewidth,  trim={{.25\width} {.12\width} {.615\width} {.752\width}},clip]	{fig/imgWV3/imgWV3_full_seg12.jpg}};
					\end{annotatedFigure}\vspace{0.5mm}
				\end{minipage}\label{fig:wv3:g}}\hspace{0.001mm}
			\subfloat[GLP-Segm]{\begin{minipage}[t]{0.19\linewidth}
					\begin{annotatedFigure}
						{\adjincludegraphics[width=\linewidth,   trim={{.18\width} {.43\width} {.715\width} {.485\width}},clip]	{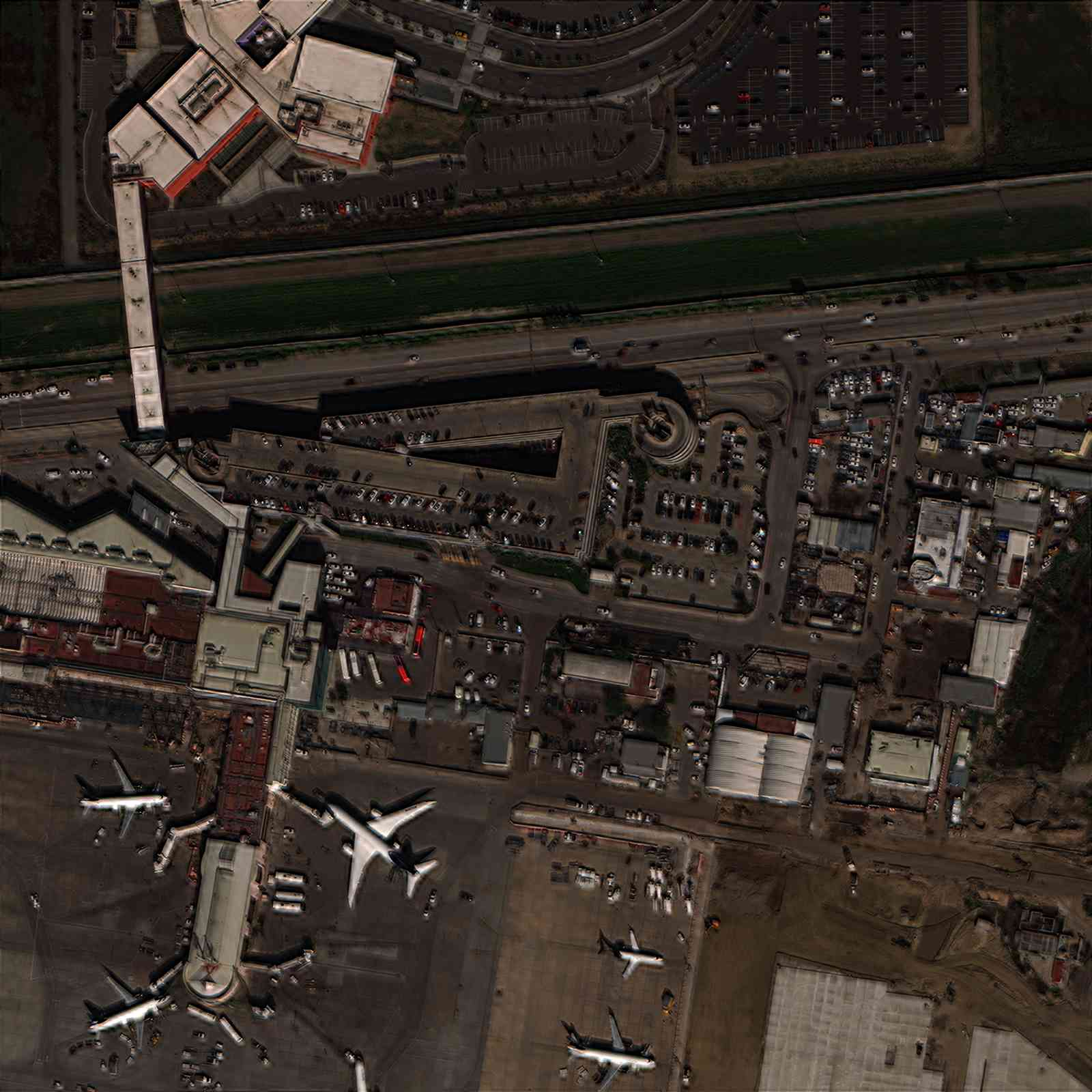}};
					\end{annotatedFigure}\vspace{0.5mm}
					\begin{annotatedFigure}
						{\adjincludegraphics[width=\linewidth,  trim={{.25\width} {.12\width} {.615\width} {.752\width}},clip] {fig/imgWV3/imgWV3_full_seg_glp.jpg}};
					\end{annotatedFigure}\vspace{0.5mm}
				\end{minipage}\label{fig:wv3:h}}\hspace{0.001mm}
			\subfloat[\hspace{-0.5mm}Target-CNN]{\begin{minipage}[t]{0.19\linewidth}
					\begin{annotatedFigure}
						{\adjincludegraphics[width=\linewidth,   trim={{.18\width} {.43\width} {.715\width} {.485\width}},clip]	{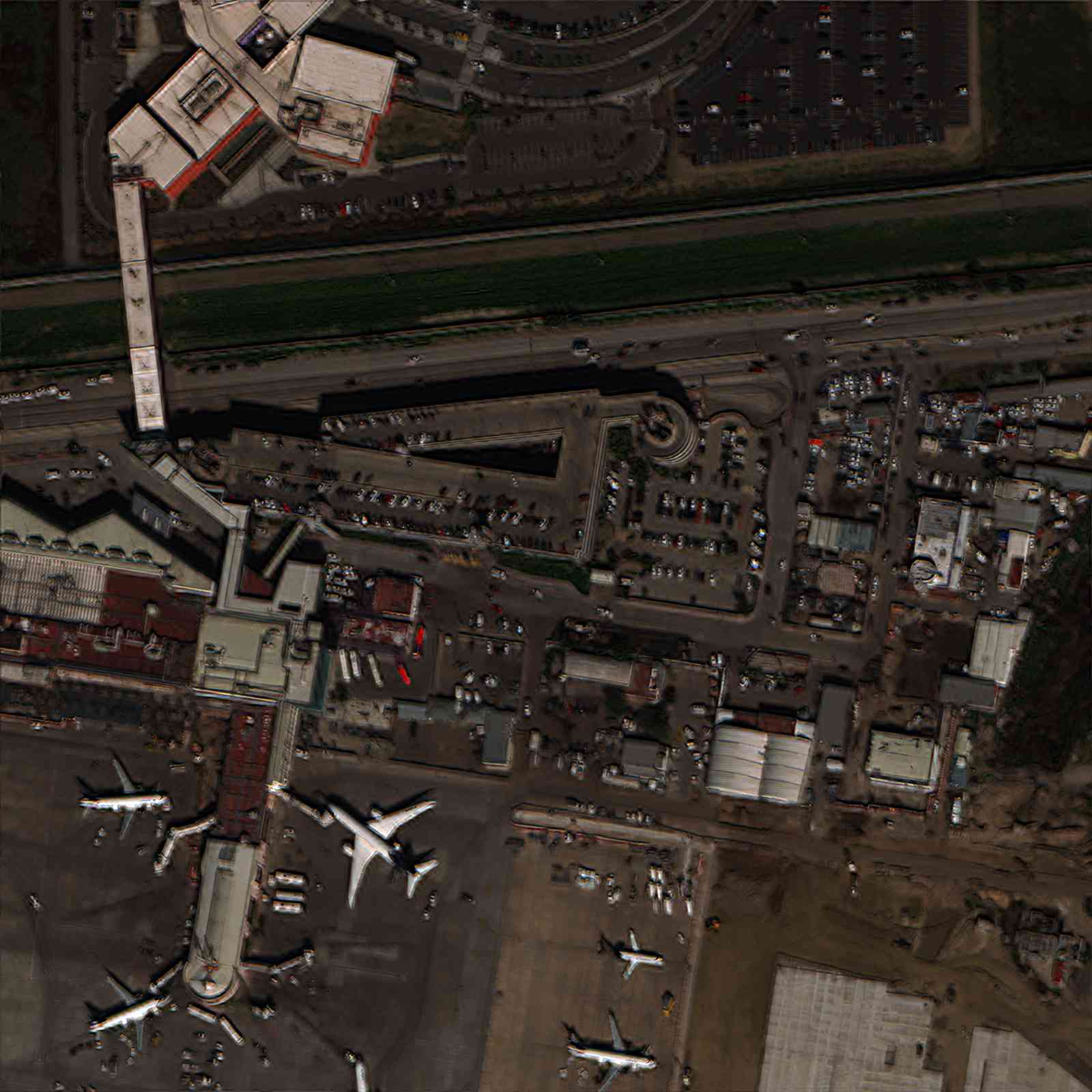}};
					\end{annotatedFigure}\vspace{0.5mm}
					\begin{annotatedFigure}
						{\adjincludegraphics[width=\linewidth,  trim={{.25\width} {.12\width} {.615\width} {.752\width}},clip]	{fig/imgWV3/imgWV3_full_pancnn.jpg}};
					\end{annotatedFigure}\vspace{0.5mm}
				\end{minipage}\label{fig:wv3:i}}\hspace{0.001mm}
			\subfloat[Ours]{\begin{minipage}[t]{0.19\linewidth}
					\begin{annotatedFigure}
						{\adjincludegraphics[width=\linewidth,   trim={{.18\width} {.43\width} {.715\width} {.485\width}},clip]	{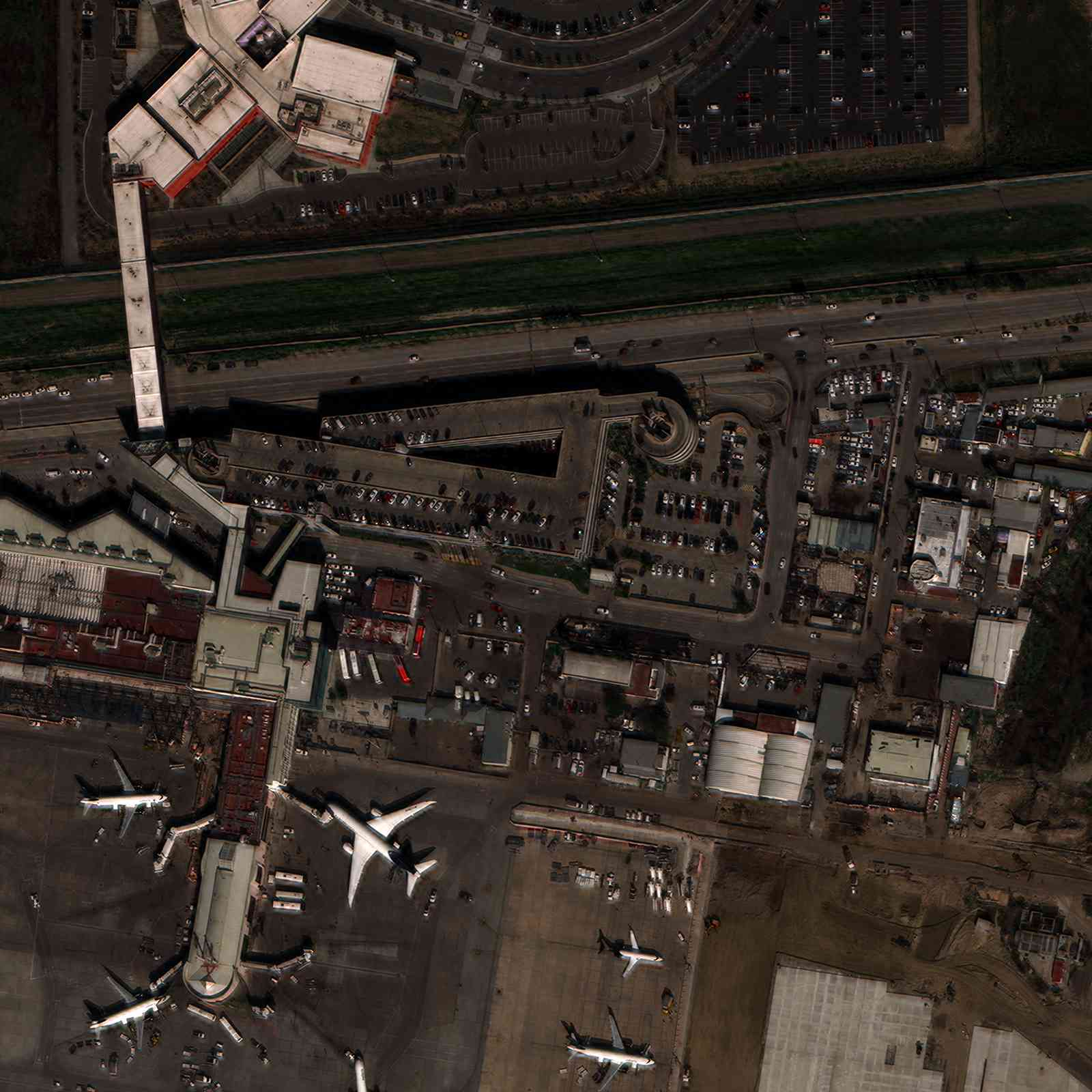}};
					\end{annotatedFigure}\vspace{0.5mm}
					\begin{annotatedFigure}
						{\adjincludegraphics[width=\linewidth,   trim={{.25\width} {.12\width} {.615\width} {.752\width}},clip]	{fig/imgWV3/imgWV3_full_ours.jpg}};
					\end{annotatedFigure}\vspace{0.5mm}
				\end{minipage}\label{fig:wv3:j}}
		\end{minipage}
		\caption{RGB channels of the pansharpened results from different methods on WV3 at full resolution.}
		\label{fig:wv3}
	\end{minipage}
\end{figure*}

\subsection{Comparison at Full Resolution}
\label{sec:full}
The assessment results of the images at full-resolution are shown in~\crefrange{fig:geoeye}{fig:wv3} and Table~\ref{tab:full}. The quantitative values indicating the best performance are marked in bold letters. And the second best results are marked in bold letters with underlines.  As discussed before, since there is no ground truth HR MSI, the non-reference assessment metrics tend to introduce measurement bias. For example, the pansharpened results with blurred effects tend to achieve better values since they are more similar to the LR MSI. Hence, it is important to perform evaluation both quantitatively and visually. 
A good method should be able to reconstruct sharp results with small spectral distortion. It is worth mentioning that from the measurement metrics, the methods with the best performance are PRACS and the supervised deep learning based method, Target-CNN. However, we can observe from the visual inspection that the reconstructed MSI from PRACS looses a lot of high-frequency details as shown in~Figs.~\ref{fig:geoeye:c}, ~\ref{fig:ik:c},~\ref{fig:wv2:c}, and ~\ref{fig:wv3:c}. Thus, although it preserves the spectral information, it could not improve the spatial resolution of HR MSI well. The supervised deep learning based method, Target-CNN, is trained using the first three testing images, i.e., GeoEye-1, IKONOS, and WorldView-2. In addition, its detail extraction and injection mapping functions are learned at reduced resolution and the functions are spatially invariant for pixels with different spectral characteristics. Thus, it introduces artifacts when applied to testing images at full resolution, as shown in Figs.~\ref{fig:geoeye:i},~\ref{fig:ik:i},~\ref{fig:wv2:i} and~\ref{fig:wv3:i}. The proposed unsupervised method could consistently reconstruct sharper HR MSI with less spectral distortion as compared to the other methods regardless of the properties of the objects. 
As shown in Figs.~\ref{fig:geoeye:j}, ~\ref{fig:ik:j},~\ref{fig:wv2:j}, and ~\ref{fig:wv3:j}, the proposed method works well even for small objects in the images because of its self-attention mechanism.
\subsection{Evaluation of the Attention Representations}
\label{sec:attention}

\begin{figure*}[htbp]\centering
	\subfloat[]{\begin{minipage}[t]{0.131\linewidth}
		\begin{annotatedFigure}
			{\adjincludegraphics[width=\linewidth, trim={{.2\width} {.15\width} {.3\width} {.15\width}},clip]	{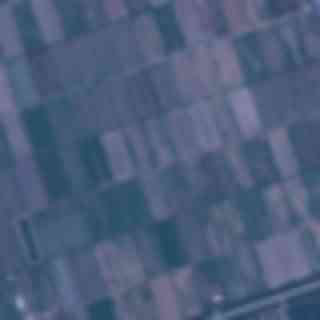}};
			\annotatedFigureBox{red}{0.05,0.35}{0.45,0.65}
		\end{annotatedFigure}\vspace{1mm}
		\begin{annotatedFigure}
			{\adjincludegraphics[width=\linewidth, trim={{.3\width} {.0\width} {.0\width} {.4\width}},clip]	{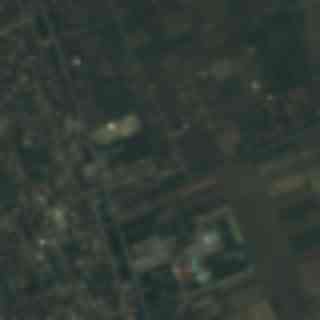}};
			\annotatedFigureBox{red}{0.1,0.2}{0.7,0.7}
		\end{annotatedFigure}\vspace{1mm}
		\begin{annotatedFigure}
			{\adjincludegraphics[width=\linewidth,trim={{.5\width} {.4\width} {.0\width} {.0\width}},clip]	{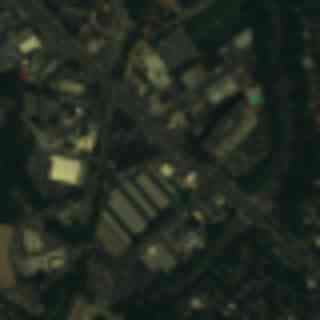}};
			\annotatedFigureBox{yellow}{0.63,0.2}{1,0.95}
		\end{annotatedFigure}\vspace{1mm}
		\begin{annotatedFigure}
			{\adjincludegraphics[width=\linewidth,trim={{.2\width} {.0\width} {.1\width} {.4\width}},clip]	{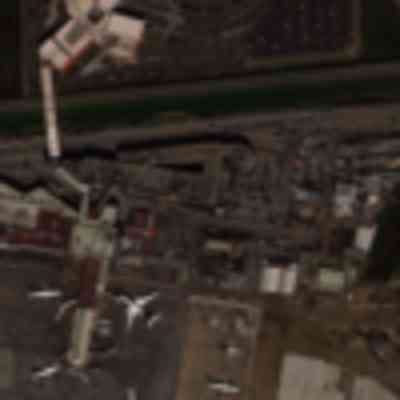}};
			\annotatedFigureBox{red}{0.1,0.1}{0.8,0.6}
		\end{annotatedFigure}
	\end{minipage}\label{fig:all_ev:a}}\hspace{0.01mm}
	\subfloat[]{\begin{minipage}[t]{0.131\linewidth}
		\begin{annotatedFigure}
			{\adjincludegraphics[width=\linewidth, trim={{.2\width} {.15\width} {.3\width} {.15\width}},clip]	{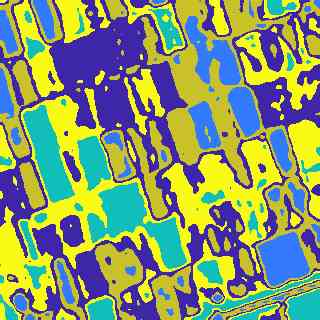}};
			\annotatedFigureBox{red}{0.05,0.35}{0.45,0.65}
		\end{annotatedFigure}\vspace{1mm}
		\begin{annotatedFigure}
			{\adjincludegraphics[width=\linewidth, trim={{.3\width} {.0\width} {.0\width} {.4\width}},clip]	{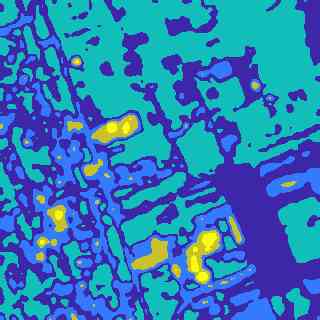}};
			\annotatedFigureBox{red}{0.1,0.2}{0.7,0.7}
		\end{annotatedFigure}\vspace{1mm}
		\begin{annotatedFigure}
			{\adjincludegraphics[width=\linewidth, trim={{.5\width} {.4\width} {.0\width} {.0\width}},clip]	{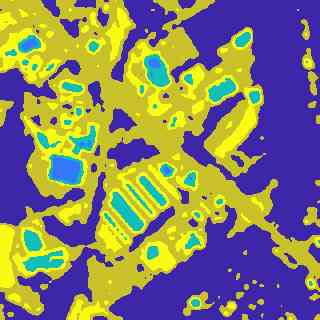}};
			\annotatedFigureBox{yellow}{0.63,0.2}{1,0.95}
		\end{annotatedFigure}\vspace{1mm}
		\begin{annotatedFigure}
			{\adjincludegraphics[width=\linewidth, trim={{.2\width} {.0\width} {.1\width} {.4\width}},clip]	{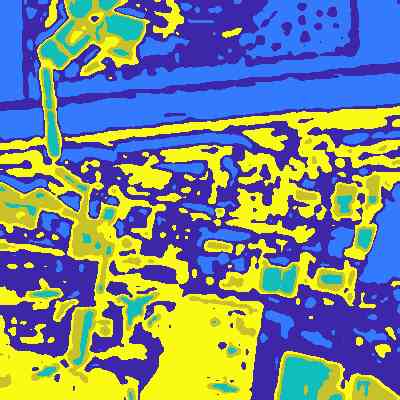}};
			\annotatedFigureBox{red}{0.1,0.1}{0.8,0.6}
		\end{annotatedFigure}
	\end{minipage}\label{fig:all_ev:b}}\hspace{0.01mm}
	\subfloat[]{\begin{minipage}[t]{0.131\linewidth}
		\begin{annotatedFigure}
			{\adjincludegraphics[width=\linewidth, trim={{.2\width} {.15\width} {.3\width} {.15\width}},clip]	{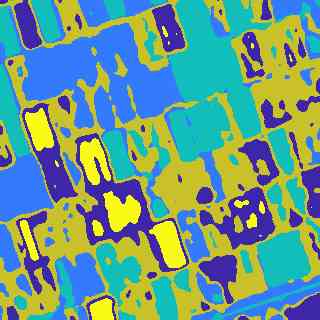}};
			\annotatedFigureBox{red}{0.05,0.35}{0.45,0.65}
		\end{annotatedFigure}\vspace{1mm}
       	\begin{annotatedFigure}
        	{\adjincludegraphics[width=\linewidth, trim={{.3\width} {.0\width} {.0\width} {.4\width}},clip]	{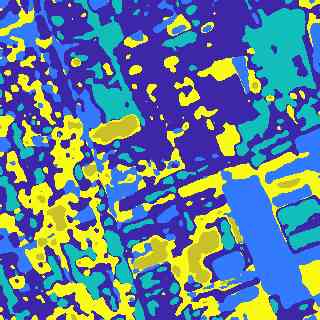}};
        	\annotatedFigureBox{red}{0.1,0.2}{0.7,0.7}
        \end{annotatedFigure}\vspace{1mm}
    	\begin{annotatedFigure}
	    	{\adjincludegraphics[width=\linewidth, trim={{.5\width} {.4\width} {.0\width} {.0\width}},clip]	{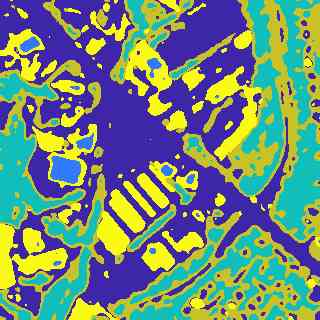}};
	    	\annotatedFigureBox{yellow}{0.63,0.2}{1,0.95}
	    \end{annotatedFigure}\vspace{1mm}
	    \begin{annotatedFigure}
	    	{\adjincludegraphics[width=\linewidth,trim={{.2\width} {.0\width} {.1\width} {.4\width}},clip]	{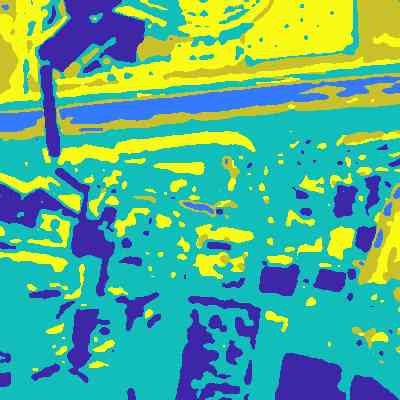}};
	    	\annotatedFigureBox{red}{0.1,0.1}{0.8,0.6}
	    \end{annotatedFigure}
	\end{minipage}\label{fig:all_ev:c}}\hspace{0.01mm}
	\subfloat[]{\begin{minipage}[t]{0.131\linewidth}
		\begin{annotatedFigure}
			{\adjincludegraphics[width=\linewidth, trim={{.2\width} {.15\width} {.3\width} {.15\width}},clip]	{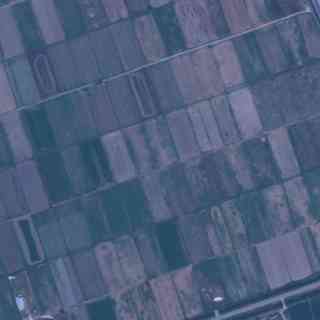}};
			\annotatedFigureBox{red}{0.05,0.35}{0.45,0.65}
		\end{annotatedFigure}\vspace{1mm}
		\begin{annotatedFigure}
			{\adjincludegraphics[width=\linewidth, trim={{.3\width} {.0\width} {.0\width} {.4\width}},clip]	{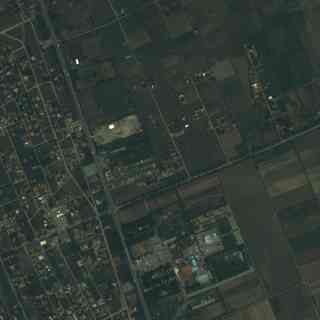}};
			\annotatedFigureBox{red}{0.1,0.2}{0.7,0.7}
		\end{annotatedFigure}\vspace{1mm}
		\begin{annotatedFigure}
			{\adjincludegraphics[width=\linewidth, trim={{.5\width} {.4\width} {.0\width} {.0\width}},clip]	{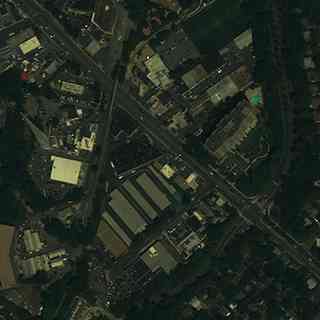}};
			\annotatedFigureBox{yellow}{0.63,0.2}{1,0.95}
		\end{annotatedFigure}\vspace{1mm}
		\begin{annotatedFigure}
			{\adjincludegraphics[width=\linewidth, trim={{.2\width} {.0\width} {.1\width} {.4\width}},clip]	{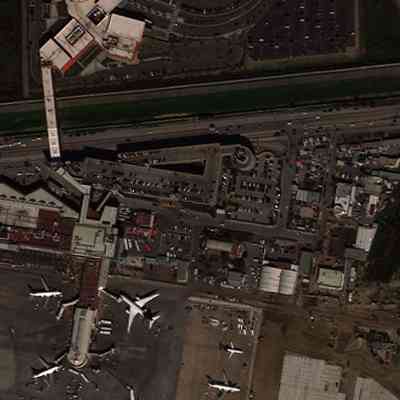}};
			\annotatedFigureBox{red}{0.1,0.1}{0.8,0.6}
		\end{annotatedFigure}
	\end{minipage}\label{fig:all_ev:d}}\hspace{0.01mm}
	\subfloat[]{\begin{minipage}[t]{0.131\linewidth}
		\begin{annotatedFigure}
			{\adjincludegraphics[width=\linewidth, trim={{.2\width} {.15\width} {.3\width} {.15\width}},clip]	{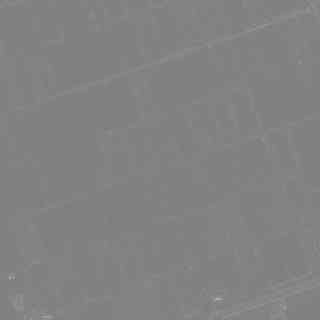}};
			\annotatedFigureBox{red}{0.05,0.35}{0.45,0.65}
		\end{annotatedFigure}\vspace{1mm}
		\begin{annotatedFigure}
			{\adjincludegraphics[width=\linewidth, trim={{.3\width} {.0\width} {.0\width} {.4\width}},clip]	{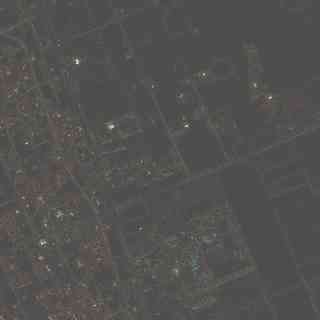}};
			\annotatedFigureBox{red}{0.1,0.2}{0.7,0.7}
		\end{annotatedFigure}\vspace{1mm}
		\begin{annotatedFigure}
			{\adjincludegraphics[width=\linewidth, trim={{.5\width} {.4\width} {.0\width} {.0\width}},clip]	{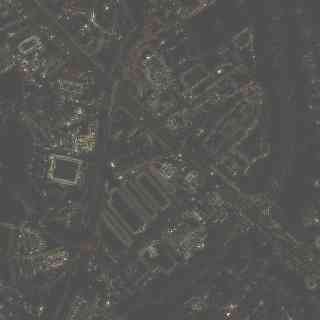}};
			\annotatedFigureBox{yellow}{0.63,0.2}{1,0.95}
		\end{annotatedFigure}\vspace{1mm}
		\begin{annotatedFigure}
			{\adjincludegraphics[width=\linewidth, trim={{.2\width} {.0\width} {.1\width} {.4\width}},clip]	{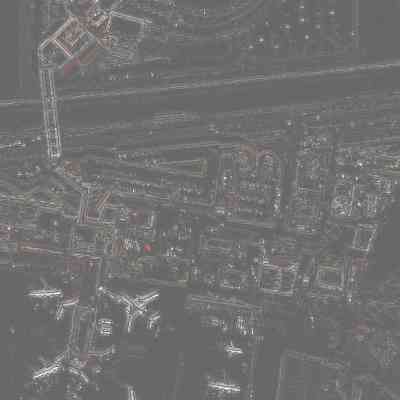}};
			\annotatedFigureBox{red}{0.1,0.1}{0.8,0.6}
		\end{annotatedFigure}
	\end{minipage}\label{fig:all_ev:e}}\hspace{0.01mm}
	\subfloat[]{\begin{minipage}[t]{0.131\linewidth}
		\begin{annotatedFigure}
			{\adjincludegraphics[width=\linewidth, trim={{.2\width} {.15\width} {.3\width} {.15\width}},clip]	{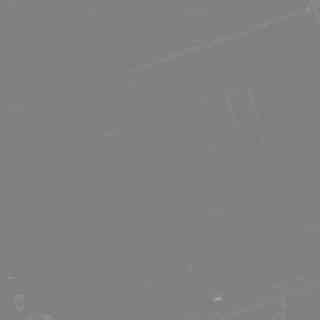}};
			\annotatedFigureBox{red}{0.05,0.35}{0.45,0.65}
		\end{annotatedFigure}\vspace{1mm}
		\begin{annotatedFigure}
			{\adjincludegraphics[width=\linewidth,  trim={{.3\width} {.0\width} {.0\width} {.4\width}},clip]	{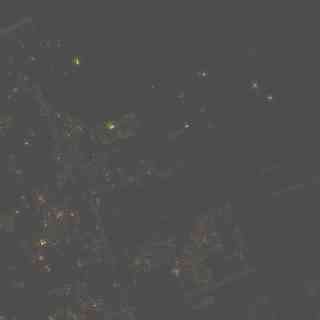}};
			\annotatedFigureBox{red}{0.1,0.2}{0.7,0.7}
		\end{annotatedFigure}\vspace{1mm}
		\begin{annotatedFigure}
			{\adjincludegraphics[width=\linewidth, trim={{.5\width} {.4\width} {.0\width} {.0\width}},clip]	{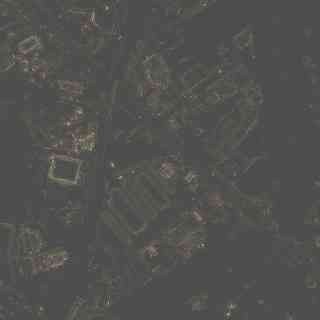}};
			\annotatedFigureBox{yellow}{0.63,0.2}{1,0.95}
		\end{annotatedFigure}\vspace{1mm}
		\begin{annotatedFigure}
			{\adjincludegraphics[width=\linewidth, trim={{.2\width} {.0\width} {.1\width} {.4\width}},clip]	{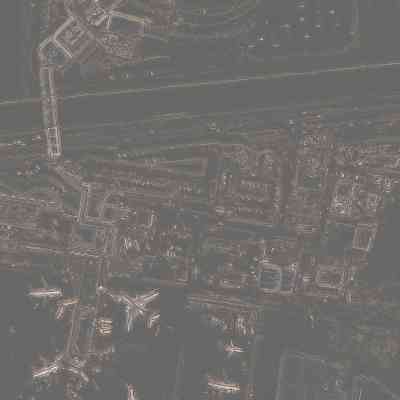}};
			\annotatedFigureBox{red}{0.1,0.1}{0.8,0.6}
		\end{annotatedFigure}
	\end{minipage}\label{fig:all_ev:f}}\hspace{0.01mm}
	\subfloat[]{\begin{minipage}[t]{0.131\linewidth}
		\begin{annotatedFigure}
			{\adjincludegraphics[width=\linewidth, trim={{.2\width} {.15\width} {.3\width} {.15\width}},clip]	{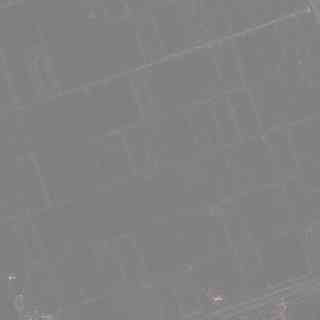}};
			\annotatedFigureBox{red}{0.05,0.35}{0.45,0.65}
		\end{annotatedFigure}\vspace{1mm}
		\begin{annotatedFigure}
			{\adjincludegraphics[width=\linewidth,  trim={{.3\width} {.0\width} {.0\width} {.4\width}},clip]	{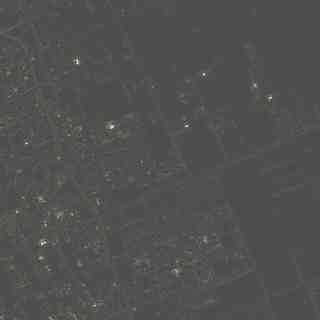}};
			\annotatedFigureBox{red}{0.1,0.2}{0.7,0.7}
		\end{annotatedFigure}\vspace{1mm}
		\begin{annotatedFigure}
			{\adjincludegraphics[width=\linewidth, trim={{.5\width} {.4\width} {.0\width} {.0\width}},clip]	{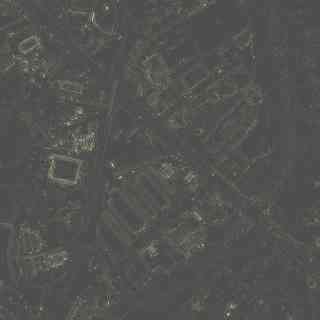}};
			\annotatedFigureBox{yellow}{0.63,0.2}{1,0.95}
		\end{annotatedFigure}\vspace{1mm}
		\begin{annotatedFigure}
			{\adjincludegraphics[width=\linewidth, trim={{.2\width} {.0\width} {.1\width} {.4\width}},clip]	{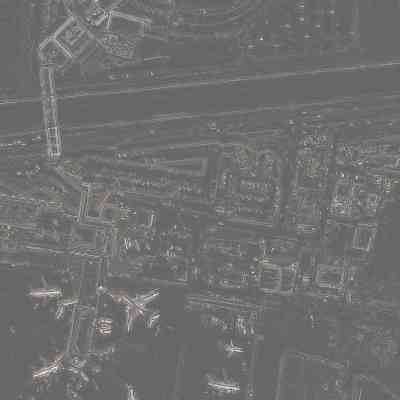}};
			\annotatedFigureBox{red}{0.1,0.1}{0.8,0.6}
		\end{annotatedFigure}
	\end{minipage}\label{fig:all_ev:g}}
	\caption{Evaluation of the stacked self-attention structure on different datasets at reduced resolution,~\ie, GeoEye-1 (first row), IKONOS (second row), WorldView2 (third row) and WorldView3 (fourth row). (a) Up-sampled low-resolution MSI. (b) Spatial varying coefficients estimated by the segmentation of the GSA-Segm. (c) Spatial varying coefficients estimated by the attention representations of the proposed approach. (d) Ground truth HR MSI. (e) The high-frequency details between (d) and (a). (f) The high-frequency details between the reconstructed HR MSI from the GSA-Segm and (a). (g) The high-frequency details between the reconstructed HR MSI from the proposed UP-SAM and (a).}
\label{fig:evaluate}
\end{figure*}

Compared to other approaches, the proposed method is designed with the self-attention mechanism, due to which it  is able to estimate spatial varying detail extraction and injection functions based on the spectral characteristics of MSI. 
In this set of experiments, we further evaluate the effectiveness of the attention representations extracted with the stacked self-attention network, which contributes the most to the performance gain. The evaluation results are shown in Fig.~\ref{fig:evaluate}. Note that, the reason we take GSA-Segm as an example for comparison is that, it defines the spatial varying coefficients and works better than the global-based approaches at both reduced and full resolutions. Besides the structure, the major difference between GSA-Segm and the proposed UP-SAM is that, instead of the segmentation, the proposed method defines the functions based on the spectral characteristics of the MSI, which has sub-pixel accuracy. Figs.~\ref{fig:all_ev:a} and ~\ref{fig:all_ev:d} show the up-sampled LR MSI and HR MSI of the four testing datasets. Figs.~\ref{fig:all_ev:b} show the spatial varying regions defined by the segments of the GSA-Segm method. Each color denotes a region that shares the same injection coefficient for each band. Figs.~\ref{fig:all_ev:c} show the spatial varying regions defined by the attention representations of the proposed method with Eq.~\eqref{equ:max}. Each area shows the pixels whose major spectrum is the same and its injection coefficient is also similar. To examine how the varying injection coefficients influence the final reconstruction results, we calculate the high-frequency injection details in the following figures. Figs.~\ref{fig:all_ev:e} shows the details between the ground truth HR MSI and the up-sampled LR MSI of different images. 
Figs.~\ref{fig:all_ev:f} are the injection details between the reconstructed HR MSI from GSA-Segm and the up-sampled LR MSI. And Figs.~\ref{fig:all_ev:g} are the injection details between the reconstructed HR MSI from the proposed method and the up-sampled LR MSI.

We can observe that, one segment defined by the GSA-Segm method may cover different materials with distinctive spectral characteristics, as shown in the highlighted red rectangles of Figs.~\ref{fig:all_ev:b}. This is because there exists mixed pixels in the LR MSI which tends to lead to such segmentation failure, as shown in Figs.~\ref{fig:all_ev:a}. As a result, the injection details of such region are added with the same coefficient.
This may weaken several details as shown in Figs.~\ref{fig:all_ev:f}. On the other hand, the proposed method is able to distinguish mixed pixels with their attention representations as shown in Figs.~\ref{fig:all_ev:c}. As a result, its injection high-frequency details in Figs.~\ref{fig:all_ev:f} are more close to that of the ground truth shown in Figs.~\ref{fig:all_ev:e}.

\subsection{Evaluation of the Attention Representation based Detail Injection}
\label{sec:injection}
\begin{figure}[h]
		\centering
		\subfloat[GeoEye-1]
		{\begin{minipage}{0.24\linewidth}
			\begin{annotatedFigure}
				{\adjincludegraphics[width=\linewidth, trim={{.2\width} {.15\width} {.3\width} {.15\width}},clip]		{fig/imgGE/imgGE_rd_msi_sc.jpg}};
			\end{annotatedFigure}\vspace{0.5mm}\\
			\begin{annotatedFigure}
				{\adjincludegraphics[width=\linewidth,trim={{.2\width} {.15\width} {.3\width} {.15\width}},clip]		{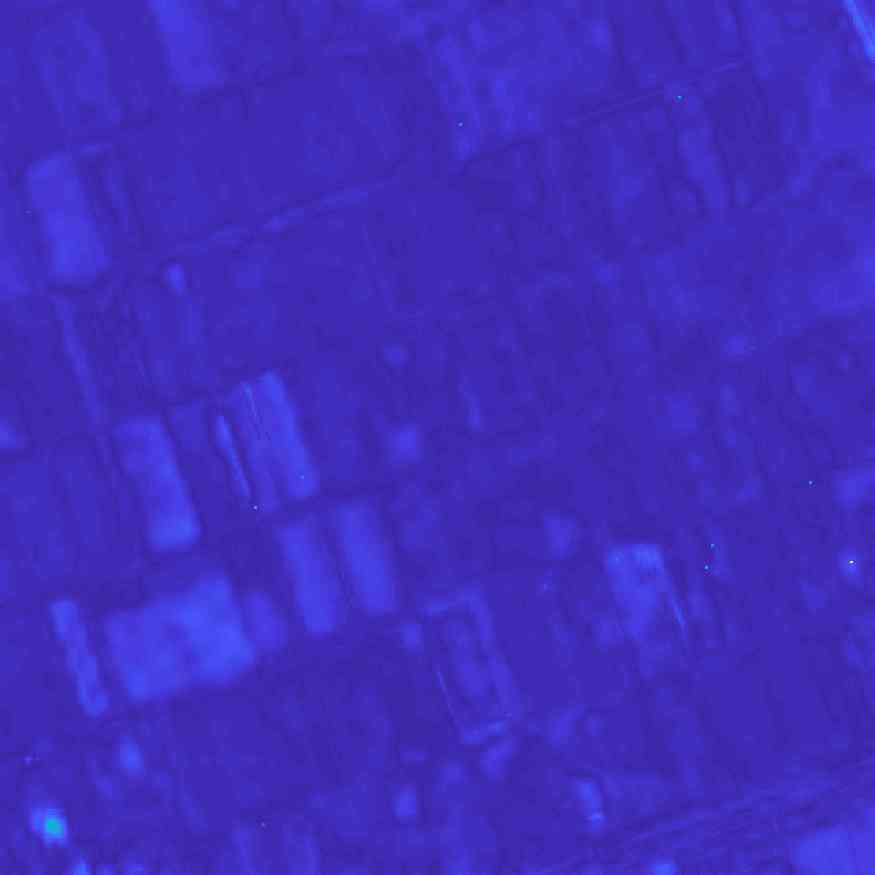}};
			\end{annotatedFigure}\vspace{0.5mm}
		\end{minipage}\label{fig:dorp:a}}\hspace{0.001mm}
		\subfloat[IKONOS]
		{\begin{minipage}{0.24\linewidth}
			\begin{annotatedFigure}
				{\adjincludegraphics[width=\linewidth, trim={{.2\width} {.15\width} {.3\width} {.15\width}},clip]		{fig/imgIK/imgIK_rd_msi.jpg}};
			\end{annotatedFigure}\vspace{0.5mm}\\
			\begin{annotatedFigure}
				{\adjincludegraphics[width=\linewidth,trim={{.2\width} {.15\width} {.3\width} {.15\width}},clip]		{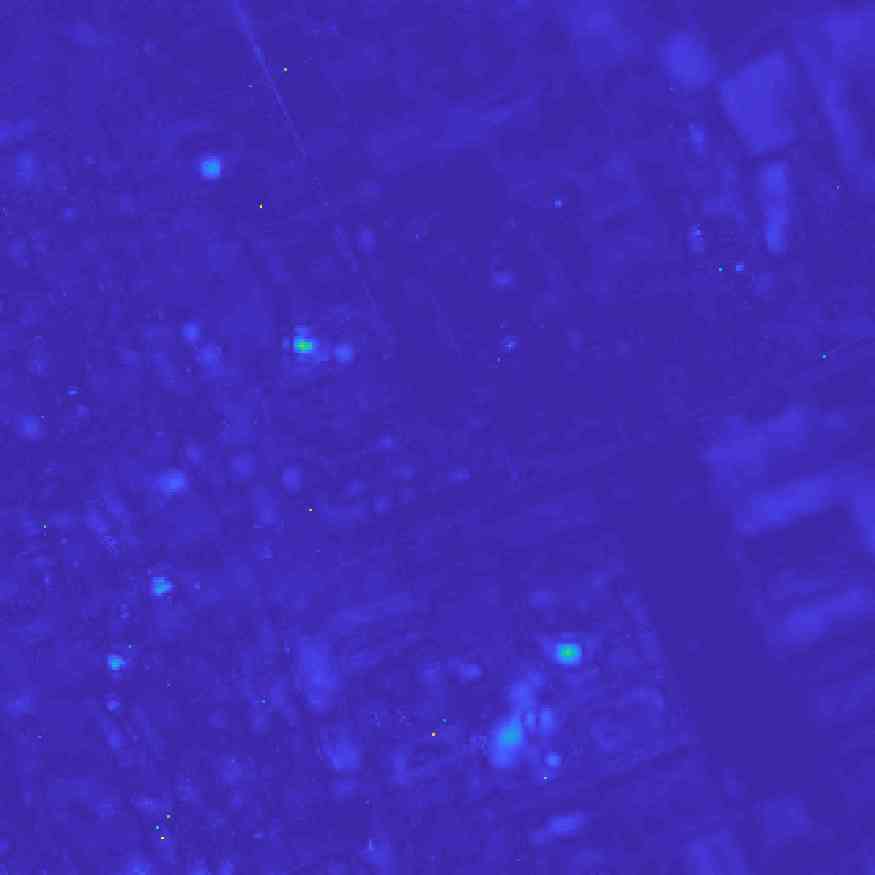}};
			\end{annotatedFigure}\vspace{0.5mm}
		\end{minipage}\label{fig:dorp:b}}\hspace{0.001mm}
		\subfloat[WorldView2]
		{\begin{minipage}{0.24\linewidth}
			\begin{annotatedFigure}
				{\adjincludegraphics[width=\linewidth, trim={{.2\width} {.15\width} {.3\width} {.15\width}},clip]		{fig/imgWV2/imgWV2_rd_msi.jpg}};
			\end{annotatedFigure}\vspace{0.5mm}\\
			\begin{annotatedFigure}
				{\adjincludegraphics[width=\linewidth,trim={{.2\width} {.15\width} {.3\width} {.15\width}},clip]		{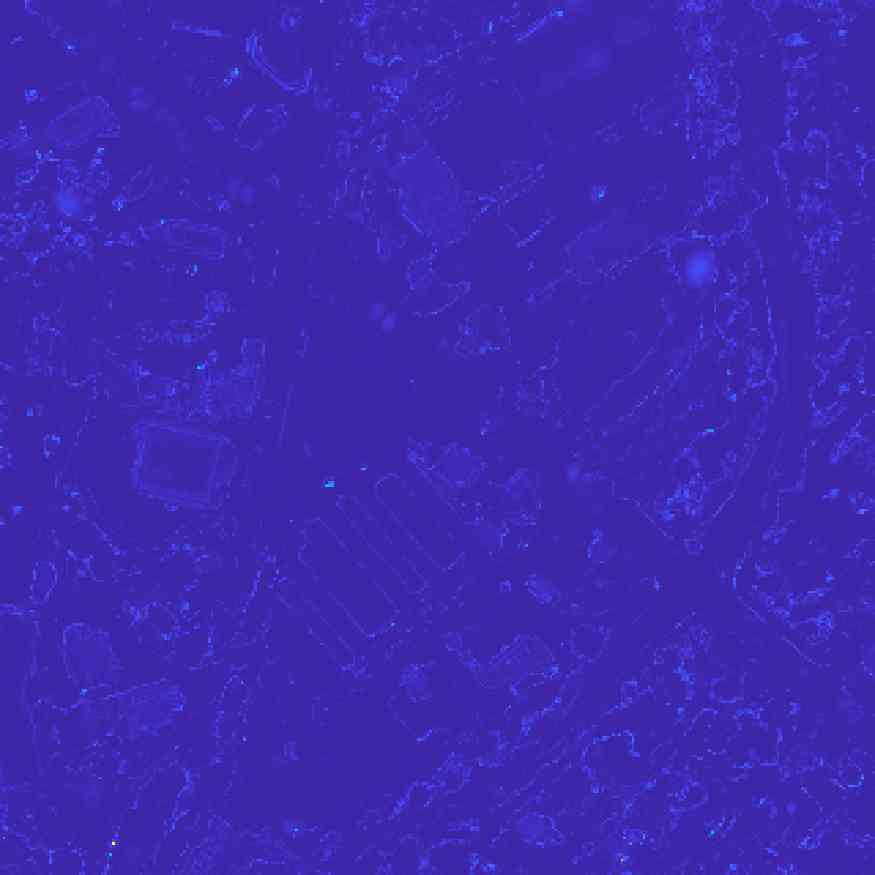}};
			\end{annotatedFigure}\vspace{0.5mm}
		\end{minipage}\label{fig:dorp:c}}\hspace{0.001mm}
		\subfloat[WorldView3]
		{\begin{minipage}{0.24\linewidth}
			\begin{annotatedFigure}
				{\adjincludegraphics[width=\linewidth, trim={{.2\width} {.15\width} {.3\width} {.15\width}},clip]		{fig/imgWV3/imgWV3_rd_msi.jpg}};
			\end{annotatedFigure}\vspace{0.5mm}\\
			\begin{annotatedFigure}
				{\adjincludegraphics[width=\linewidth,trim={{.2\width} {.15\width} {.3\width} {.15\width}},clip]		{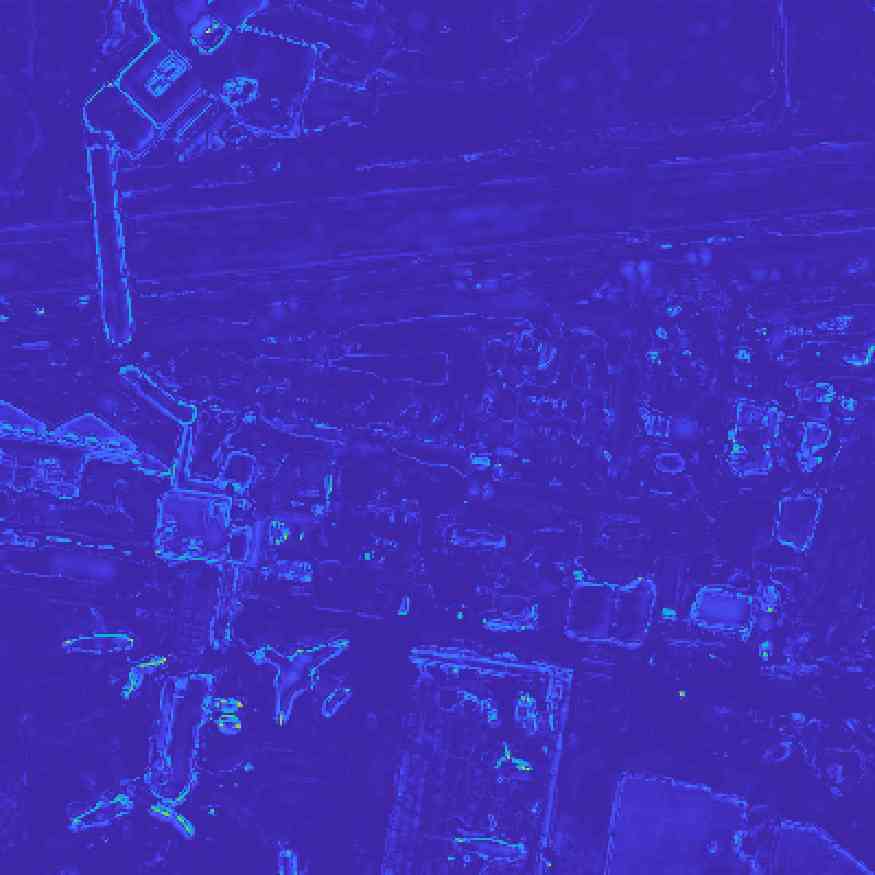}};
			\end{annotatedFigure}\vspace{0.5mm}	
		\end{minipage}\label{fig:dorp:d}}\hspace{0.001mm}
		\caption{Evaluation of the attention representation based detail injection. Top: HR MSI. Bottom: difference image between the reconstructed HR MSI when the details are injected on the MSI and when the details are injected on the representation maps.}
		\label{fig:dorp}
	\end{figure}

Since the proposed UP-SAM provides sub-pixel representation accuracy using the attention mechanism, correspondingly, the detail injection process in UP-SAM is conducted on each representation map instead of the mixed pixels. To evaluate such procedure, we show the difference image between the reconstructed HR MSI when the details are injected on the MSI and when the details are injected on the representation maps in Fig.~\ref{fig:dorp}. We observe that the differences with higher values are mostly shown at the edges of the objects or objects consisting of more than one material, where mixed pixels are more evident. The quantitative comparison in Table~\ref{tab:rp_diff} shows that when we inject the details on the representation maps with sub-pixel accuracy, the performance of the pansharpening is improved. Combining with the observation made in the visual comparison in Fig.~\ref{fig:dorp}, we could infer that the more mixed pixels in the image, the more performance improvements in the pansharpened result.

	\begin{table}[htbp]
		\caption{Performance comparison at reduced resolution when the details are injected on the MSI and when the details are injected on the representation maps (RM), respectively.}
		\label{tab:rp_diff}
		\centering
		\begin{tabular}{l | l | c | c| c | c }
			\hline
			&Injection 	&PSNR	 &SAM	 &ERGAS	  &$Q2^n$\\
			\hline
			\multirow{2}*{GeoEye-1}
			&MSI		&33.8068  &2.2257   &1.3955   &0.91\\
			&RM			&{33.9028}&{2.2108} &{1.3908} &{0.9109}\\
			\hline
			\multirow{2}*{IKONOS}
			&MSI		&34.6055  &3.2293   &2.1638   &0.8784\\
			&RM			&{34.6342}&{3.2122} &{2.1559} &{0.8790}\\
			\hline
			\multirow{2}*{WorldView2}
			&MSI		&30.0499  &7.2992   &4.4758   &0.8900\\
			&RM			&{30.091} &{7.27}   &{4.4527} &{0.8909}\\
			\hline
			\multirow{2}*{WorldView3}
			&MSI		&25.0273  &5.9748   &6.8642   &0.8140\\
			&RM			&{25.0487}&{5.8202} &{6.8566} &{0.8141}\\
			\hline
			&Desired 	&$+\infty$ &0 &0 &1 \\
			\hline  
		\end{tabular}%
	\end{table}	
	
\subsection{Computational Efficiency}
\label{sec:time}
\begin{table*}
\centering
\caption{Run time for tradition unsupervised methods collected on CPUs (top section) and deep learning-based methods (bottom section) collected on GPU (for supervised Target-CNN with training) and CPU (for unsupervised UP-SAM without training).}
	\label{tab:time}
	\begin{tabular}{l | c| c| c |c |c|c|c}
		\hline
		&GSA&PRACS&BDSD-PC&MTF-CBD&GLP-Reg-FS&GSA-Segm&GLP-Segm\\	
		\hline				
		Time (s)&0.19&0.4073&0.8161&0.2329&0.2335&0.3768&0.5804\\
		\hline\hline
		&\multicolumn{3}{c|}{Target-CNN}&\multicolumn{4}{c}{Proposed UP-SAM}\\
		\hline
		Time (s)&\multicolumn{3}{c|}{47.0352 + 2hr training on GPU}&\multicolumn{4}{c}{565.3576 + 0 training on CPU}\\
	\hline
	\end{tabular}
	\end{table*}
	The average processing time of all the approaches on an MSI with resolution $320\times 320\times 8$ are evaluated. The specification of the test computer is Intel Core i7 CPU@3.1 GHz with RAM 8 GB. The efficiency of different approaches are shown in Table~\ref{tab:time}. All the traditional methods (shown in the top section of the table) are non-deep-learning-based unsupervised approaches and are implemented with Matlab on CPU. The proposed UP-SAM and Target-CNN (shown in the bottom section of the table) are deep-learning-based and are implemented with python. UP-SAM is performed on CPU while Target-CNN on GPU. The deep-learning-based methods generally consume more time than non-deep-learning-based methods due to the deep structures adopted. 
	The Target-CNN trains the network to learn a general mapping function between the low-resolution (LR) MSI and the ground truth high-resolution (HR) MSI. Thus the training speed also depends on the number of training samples. Based on the paper~\cite{scarpa2018target}, it requires around 2 hours on a computer with GPU. We expect it requires longer training time on a computer without GPU. We observe that although the proposed UP-SAM method consumes more testing time as compared to that of the Target-CNN, it is unsupervised which means no training is needed. Furthermore, with the self-attention mechanism, the network could consistently reconstruct sharper HR MSI with less spectral distortion as compared to that of the Target-CNN or other non-deep-learning-based methods.


\section{Conclusion} 
\label{sec:conclusion}
We proposed an unsupervised pansharpening method based on the self-attention mechanism. The method estimates the spatial varying detail extraction and injection functions based on the spectral characteristics of the MSI. And the spectral characteristics are learned based on the attention representations extracted with the stacked self-attention model possessing the stick-breaking structure. Both the detail extraction and injection functions were designed according to the attention representations to increase the reconstruction accuracy and reduce the spectral distortion. Extensive experiments on different types of MSI demonstrate the superiority of the proposed approach over state-of-the-art.

\section*{Acknowledgment}
This work was supported in part by NASA NNX12CB05C and NNX16CP38P.The authors would like to thank all the developers of the evaluated methods who kindly offer their codes for comparisons.

\ifCLASSOPTIONcaptionsoff
  \newpage
\fi



%

\bibliographystyle{IEEEtran}
\bibliography{refs}

\begin{thebibliography}{10}
\providecommand{\url}[1]{#1}
\csname url@samestyle\endcsname
\providecommand{\newblock}{\relax}
\providecommand{\bibinfo}[2]{#2}
\providecommand{\BIBentrySTDinterwordspacing}{\spaceskip=0pt\relax}
\providecommand{\BIBentryALTinterwordstretchfactor}{4}
\providecommand{\BIBentryALTinterwordspacing}{\spaceskip=\fontdimen2\font plus
\BIBentryALTinterwordstretchfactor\fontdimen3\font minus
  \fontdimen4\font\relax}
\providecommand{\BIBforeignlanguage}[2]{{%
\expandafter\ifx\csname l@#1\endcsname\relax
\typeout{** WARNING: IEEEtran.bst: No hyphenation pattern has been}%
\typeout{** loaded for the language `#1'. Using the pattern for}%
\typeout{** the default language instead.}%
\else
\language=\csname l@#1\endcsname
\fi
#2}}
\providecommand{\BIBdecl}{\relax}
\BIBdecl

\bibitem{bovolo2009analysis}
F.~Bovolo, L.~Bruzzone, L.~Capobianco, A.~Garzelli, S.~Marchesi, and
  F.~Nencini, ``Analysis of the effects of pansharpening in change detection on
  vhr images,'' \emph{IEEE Geoscience and Remote Sensing Letters}, vol.~7,
  no.~1, pp. 53--57, 2009.

\bibitem{souza2003mapping}
C.~Souza~Jr, L.~Firestone, L.~M. Silva, and D.~Roberts, ``Mapping forest
  degradation in the eastern amazon from spot 4 through spectral mixture
  models,'' \emph{Remote Sensing of Environment}, vol.~87, no.~4, pp. 494--506,
  2003.

\bibitem{gilbertson2017effect}
J.~K. Gilbertson, J.~Kemp, and A.~Van~Niekerk, ``Effect of pan-sharpening
  multi-temporal landsat 8 imagery for crop type differentiation using
  different classification techniques,'' \emph{Computers and Electronics in
  Agriculture}, vol. 134, pp. 151--159, 2017.

\bibitem{song2015adaptive}
Y.~Song, W.~Wu, Z.~Liu, X.~Yang, K.~Liu, and W.~Lu, ``An adaptive pansharpening
  method by using weighted least squares filter,'' \emph{IEEE Geoscience and
  Remote Sensing Letters}, vol.~13, no.~1, pp. 18--22, 2015.

\bibitem{aiazzi2012twenty}
B.~Aiazzi, L.~Alparone, S.~Baronti, A.~Garzelli, and M.~Selva, ``Twenty-five
  years of pansharpening: A critical review and new developments,''
  \emph{Signal and Image Processing for Remote Sensing}, pp. 552--599, 2012.

\bibitem{vivone2014critical}
G.~Vivone, L.~Alparone, J.~Chanussot, M.~Dalla~Mura, A.~Garzelli, G.~A.
  Licciardi, R.~Restaino, and L.~Wald, ``A critical comparison among
  pansharpening algorithms,'' \emph{IEEE Transactions on Geoscience and Remote
  Sensing}, vol.~53, no.~5, pp. 2565--2586, 2014.

\bibitem{restaino2017context}
R.~Restaino, M.~Dalla~Mura, G.~Vivone, and J.~Chanussot, ``Context-adaptive
  pansharpening based on image segmentation,'' \emph{IEEE Transactions on
  Geoscience and Remote Sensing}, vol.~55, no.~2, pp. 753--766, 2017.

\bibitem{8435988}
G.~{Vivone}, P.~{Addesso}, R.~{Restaino}, M.~{Dalla Mura}, and J.~{Chanussot},
  ``Pansharpening based on deconvolution for multiband filter estimation,''
  \emph{IEEE Transactions on Geoscience and Remote Sensing}, vol.~57, no.~1,
  pp. 540--553, Jan 2019.

\bibitem{thomas2008synthesis}
C.~Thomas, T.~Ranchin, L.~Wald, and J.~Chanussot, ``Synthesis of multispectral
  images to high spatial resolution: A critical review of fusion methods based
  on remote sensing physics,'' \emph{IEEE Transactions on Geoscience and Remote
  Sensing}, vol.~46, no.~5, pp. 1301--1312, 2008.

\bibitem{chavez1991comparison}
S.~C. Sides, J.~A. Anderson \emph{et~al.}, ``Comparison of three different
  methods to merge multiresolution and multispectral data- landsat tm and spot
  panchromatic,'' \emph{Photogrammetric Engineering and Remote Sensing},
  vol.~57, no.~3, pp. 295--303, 1991.

\bibitem{aiazzi2007improving}
B.~Aiazzi, S.~Baronti, and M.~Selva, ``Improving component substitution
  pansharpening through multivariate regression of ms $+ $ pan data,''
  \emph{IEEE Transactions on Geoscience and Remote Sensing}, vol.~45, no.~10,
  pp. 3230--3239, 2007.

\bibitem{garzelli2008}
A.~{Garzelli}, F.~{Nencini}, and L.~{Capobianco}, ``Optimal mmse pan sharpening
  of very high resolution multispectral images,'' \emph{IEEE Transactions on
  Geoscience and Remote Sensing}, vol.~46, no.~1, pp. 228--236, Jan 2008.

\bibitem{choi2010new}
J.~Choi, K.~Yu, and Y.~Kim, ``A new adaptive component-substitution-based
  satellite image fusion by using partial replacement,'' \emph{IEEE
  Transactions on Geoscience and Remote Sensing}, vol.~49, no.~1, pp. 295--309,
  2010.

\bibitem{vivone2019robust}
G.~Vivone, ``Robust band-dependent spatial-detail approaches for panchromatic
  sharpening,'' \emph{IEEE Transactions on Geoscience and Remote Sensing},
  vol.~57, no.~9, pp. 6421--6433, 2019.

\bibitem{aiazzi2006mtf}
B.~Aiazzi, L.~Alparone, S.~Baronti, A.~Garzelli, and M.~Selva, ``Mtf-tailored
  multiscale fusion of high-resolution ms and pan imagery,''
  \emph{Photogrammetric Engineering \& Remote Sensing}, vol.~72, no.~5, pp.
  591--596, 2006.

\bibitem{alparone2007comparison}
L.~Alparone, L.~Wald, J.~Chanussot, C.~Thomas, P.~Gamba, and L.~M. Bruce,
  ``Comparison of pansharpening algorithms: Outcome of the 2006 grs-s
  data-fusion contest,'' \emph{IEEE Transactions on Geoscience and Remote
  Sensing}, vol.~45, no.~10, pp. 3012--3021, 2007.

\bibitem{vivone2018full}
G.~Vivone, R.~Restaino, and J.~Chanussot, ``Full scale regression-based
  injection coefficients for panchromatic sharpening,'' \emph{IEEE Transactions
  on Image Processing}, vol.~27, no.~7, pp. 3418--3431, 2018.

\bibitem{wang2013robust}
H.~Wang, W.~Jiang, C.~Lei, S.~Qin, and J.~Wang, ``A robust image fusion method
  based on local spectral and spatial correlation,'' \emph{IEEE Geoscience and
  Remote Sensing letters}, vol.~11, no.~2, pp. 454--458, 2013.

\bibitem{garzelli2014pansharpening}
A.~Garzelli, ``Pansharpening of multispectral images based on nonlocal
  parameter optimization,'' \emph{IEEE Transactions on Geoscience and Remote
  Sensing}, vol.~53, no.~4, pp. 2096--2107, 2014.

\bibitem{bioucas2012hyperspectral}
J.~M. Bioucas-Dias, A.~Plaza, N.~Dobigeon, M.~Parente, Q.~Du, P.~Gader, and
  J.~Chanussot, ``Hyperspectral unmixing overview: Geometrical, statistical,
  and sparse regression-based approaches,'' \emph{IEEE Journal of Selected
  Topics in Applied Earth Observations and Remote Sensing}, vol.~5, no.~2, pp.
  354--379, 2012.

\bibitem{li2016robust}
J.~Li, J.~M. Bioucas-Dias, A.~Plaza, and L.~Liu, ``Robust collaborative
  nonnegative matrix factorization for hyperspectral unmixing,'' \emph{IEEE
  Transactions on Geoscience and Remote Sensing}, vol.~54, no.~10, pp.
  6076--6090, 2016.

\bibitem{qu2018udas}
Y.~Qu and H.~Qi, ``udas: An untied denoising autoencoder with sparsity for
  spectral unmixing,'' \emph{IEEE Transactions on Geoscience and Remote
  Sensing}, vol.~57, no.~3, pp. 1698--1712, 2018.

\bibitem{scarpa2018target}
G.~Scarpa, S.~Vitale, and D.~Cozzolino, ``Target-adaptive cnn-based
  pansharpening,'' \emph{IEEE Transactions on Geoscience and Remote Sensing},
  vol.~56, no.~9, pp. 5443--5457, 2018.

\bibitem{he2019hyperpnn}
L.~He, J.~Zhu, J.~Li, A.~Plaza, J.~Chanussot, and B.~Li, ``Hyperpnn:
  Hyperspectral pansharpening via spectrally predictive convolutional neural
  networks,'' \emph{IEEE Journal of Selected Topics in Applied Earth
  Observations and Remote Sensing}, vol.~12, no.~8, pp. 3092--3100, 2019.

\bibitem{he2019pansharpening}
L.~He, Y.~Rao, J.~Li, J.~Chanussot, A.~Plaza, J.~Zhu, and B.~Li,
  ``Pansharpening via detail injection based convolutional neural networks,''
  \emph{IEEE Journal of Selected Topics in Applied Earth Observations and
  Remote Sensing}, vol.~12, no.~4, pp. 1188--1204, 2019.

\bibitem{huang2015new}
W.~Huang, L.~Xiao, Z.~Wei, H.~Liu, and S.~Tang, ``A new pan-sharpening method
  with deep neural networks,'' \emph{IEEE Geoscience and Remote Sensing
  Letters}, vol.~12, no.~5, pp. 1037--1041, 2015.

\bibitem{masi2016pansharpening}
G.~Masi, D.~Cozzolino, L.~Verdoliva, and G.~Scarpa, ``Pansharpening by
  convolutional neural networks,'' \emph{Remote Sensing}, vol.~8, no.~7, p.
  594, 2016.

\bibitem{wei2017boosting}
Y.~Wei, Q.~Yuan, H.~Shen, and L.~Zhang, ``Boosting the accuracy of
  multi-spectral image pan-sharpening by learning a deep residual network,''
  \emph{arXiv preprint arXiv:1705.07556}, 2017.

\bibitem{he2016deep}
K.~He, X.~Zhang, S.~Ren, and J.~Sun, ``Deep residual learning for image
  recognition,'' \emph{Proceedings of the IEEE conference on Computer Vision
  and Pattern Recognition (CVPR)}, pp. 770--778, 2016.

\bibitem{yang2017pannet}
J.~Yang, X.~Fu, Y.~Hu, Y.~Huang, X.~Ding, and J.~Paisley, ``Pannet: A deep
  network architecture for pan-sharpening,'' \emph{Proceedings of the IEEE
  International Conference on Computer Vision (ICCV)}, pp. 5449--5457, 2017.

\bibitem{pelleg2000x}
D.~Pelleg, A.~W. Moore \emph{et~al.}, ``X-means: Extending k-means with
  efficient estimation of the number of clusters.'' \emph{Proceedings of
  International Conference on Machine Learning (ICML)}, vol.~1, pp. 727--734.,
  2000.

\bibitem{salembier2000binary}
P.~Salembier and L.~Garrido, ``Binary partition tree as an efficient
  representation for image processing, segmentation, and information
  retrieval,'' \emph{IEEE Transactions on Image Processing}, vol.~9, no.~4, pp.
  561--576, 2000.

\bibitem{comaniciu2002mean}
D.~Comaniciu and P.~Meer, ``Mean shift: A robust approach toward feature space
  analysis,'' \emph{IEEE Transactions on Pattern Analysis \& Machine
  Intelligence}, no.~5, pp. 603--619, 2002.

\bibitem{tang2017integrating}
M.~Tang, L.~Gao, A.~Marinoni, P.~Gamba, and B.~Zhang, ``Integrating spatial
  information in the normalized p-linear algorithm for nonlinear hyperspectral
  unmixing,'' \emph{IEEE Journal of Selected Topics in Applied Earth
  Observations and Remote Sensing}, vol.~11, no.~4, pp. 1179--1190, 2017.

\bibitem{luo2018bilinear}
W.~Luo, L.~Gao, R.~Zhang, A.~Marinoni, and B.~Zhang, ``Bilinear normal mixing
  model for spectral unmixing,'' \emph{IET Image Processing}, vol.~13, no.~2,
  pp. 344--354, 2018.

\bibitem{xu2016using}
X.~Xu, X.~Tong, A.~Plaza, Y.~Zhong, H.~Xie, and L.~Zhang, ``Using linear
  spectral unmixing for subpixel mapping of hyperspectral imagery: A
  quantitative assessment,'' \emph{IEEE Journal of Selected Topics in Applied
  Earth Observations and Remote Sensing}, vol.~10, no.~4, pp. 1589--1600, 2016.

\bibitem{qu2018hyperspectral}
Y.~Qu, W.~Wang, R.~Guo, B.~Ayhan, C.~Kwan, S.~Vance, and H.~Qi, ``Hyperspectral
  anomaly detection through spectral unmixing and dictionary-based low-rank
  decomposition,'' \emph{IEEE Transactions on Geoscience and Remote Sensing},
  vol.~56, no.~8, pp. 4391--4405, 2018.

\bibitem{bahdanau2014neural}
D.~Bahdanau, K.~Cho, and Y.~Bengio, ``Neural machine translation by jointly
  learning to align and translate,'' \emph{arXiv preprint arXiv:1409.0473},
  2014.

\bibitem{xu2015show}
K.~Xu, J.~Ba, R.~Kiros, K.~Cho, A.~Courville, R.~Salakhudinov, R.~Zemel, and
  Y.~Bengio, ``Show, attend and tell: Neural image caption generation with
  visual attention,'' \emph{International conference on machine learning}, pp.
  2048--2057, 2015.

\bibitem{gregor2015draw}
K.~Gregor, I.~Danihelka, A.~Graves, D.~J. Rezende, and D.~Wierstra, ``Draw: A
  recurrent neural network for image generation,'' \emph{arXiv preprint
  arXiv:1502.04623}, 2015.

\bibitem{vaswani2017attention}
A.~Vaswani, N.~Shazeer, N.~Parmar, J.~Uszkoreit, L.~Jones, A.~N. Gomez,
  {\L}.~Kaiser, and I.~Polosukhin, ``Attention is all you need,''
  \emph{Advances in Neural Information Processing Systems}, pp. 5998--6008,
  2017.

\bibitem{chen2017pixelsnail}
X.~Chen, N.~Mishra, M.~Rohaninejad, and P.~Abbeel, ``Pixelsnail: An improved
  autoregressive generative model,'' \emph{arXiv preprint arXiv:1712.09763},
  2017.

\bibitem{zhang2018self}
H.~Zhang, I.~Goodfellow, D.~Metaxas, and A.~Odena, ``Self-attention generative
  adversarial networks,'' \emph{arXiv preprint arXiv:1805.08318}, 2018.

\bibitem{sethuraman1994constructive}
J.~Sethuraman, ``A constructive definition of dirichlet priors,''
  \emph{Statistica Sinica}, pp. 639--650, 1994.

\bibitem{nalisnick2017stick}
E.~Nalisnick and P.~Smyth, ``Stick-breaking variational autoencoders,''
  \emph{International Conference on Learning Representations (ICLR)}, 2017.

\bibitem{qu2018unsupervised}
Y.~Qu, H.~Qi, and C.~Kwan, ``Unsupervised sparse dirichlet-net for
  hyperspectral image super-resolution,'' \emph{Proceedings of the IEEE
  Conference on Computer Vision and Pattern Recognition}, pp. 2511--2520, 2018.

\bibitem{huang2018sparse}
S.~Huang and T.~D. Tran, ``Sparse signal recovery via generalized entropy
  functions minimization,'' \emph{IEEE Transactions on Signal Processing},
  vol.~67, no.~5, pp. 1322--1337, 2018.

\bibitem{huang2017densely}
G.~Huang, Z.~Liu, L.~Van Der~Maaten, and K.~Q. Weinberger, ``Densely connected
  convolutional networks,'' \emph{Proceedings of the IEEE Conference on
  Computer Vision and Pattern Recognition}, pp. 4700--4708, 2017.

\bibitem{wald1997fusion}
L.~Wald, T.~Ranchin, and M.~Mangolini, ``Fusion of satellite images of
  different spatial resolutions: Assessing the quality of resulting images,''
  \emph{Photogrammetric Engineering \& Remote Sensing}, vol.~63, no.~6, pp.
  691--699, 1997.

\bibitem{yuhas1992discrimination}
R.~H. Yuhas, A.~F. Goetz, and J.~W. Boardman, ``Discrimination among semi-arid
  landscape endmembers using the spectral angle mapper (sam) algorithm,''
  \emph{Proc. Summaries 3rd Annu. JPL Airborne Geosci. Workshop}, p. 147–149,
  1992.

\bibitem{garzelli2009hypercomplex}
A.~Garzelli and F.~Nencini, ``Hypercomplex quality assessment of
  multi/hyperspectral images,'' \emph{IEEE Geoscience and Remote Sensing
  Letters}, vol.~6, no.~4, pp. 662--665, 2009.

\bibitem{alparone2008multispectral}
L.~Alparone, B.~Aiazzi, S.~Baronti, A.~Garzelli, F.~Nencini, and M.~Selva,
  ``Multispectral and panchromatic data fusion assessment without reference,''
  \emph{Photogrammetric Engineering \& Remote Sensing}, vol.~74, no.~2, pp.
  193--200, 2008.

\end{thebibliography}
\end{document}